\documentclass[12pt]{article}
\usepackage{tikz,amssymb,url,amsmath,amsthm,listings,algorithm,algpseudocode,enumerate, caption, subcaption,xspace, float, verbatim,tabularx,mathtools}
\usepackage[pointedenum]{paralist}
\usepackage{listings}
\usepackage{tikz}
\usepackage[authoryear]{natbib}
\usepackage{epstopdf}
\usepackage[breaklinks=true,pdfstartview=FitH]{hyperref}
\usepackage{times}
\usepackage{graphicx}
\usepackage{color}
\usepackage{multirow}
\usepackage{rotating}
\usepackage{bbm}
\usepackage{array}
\usepackage{latexsym}
\usepackage{xr}
\usepackage[symbol]{footmisc} 
\newcounter{bar}

\def\bx{{\boldsymbol x}}
\def\by{{\boldsymbol y}}
\def\bv{{\boldsymbol v}}
\def\bb{{\boldsymbol b}}
\def\bd{{\boldsymbol d}}

\def\bw{{\boldsymbol w}}
\def\bs{{\boldsymbol s}}

\def\bz{{\boldsymbol z}}

\def\bg{{\boldsymbol g}}
\def\bzero{{\boldsymbol 0}}
\def\btheta{\boldsymbol \theta}

\algblock{ParFor}{EndParFor}
\algnewcommand\algorithmicparfor{\textbf{parfor}}
\algnewcommand\algorithmicpardo{\textbf{do}}
\algnewcommand\algorithmicendparfor{\textbf{end\ parfor}}
\algrenewtext{ParFor}[1]{\algorithmicparfor\ #1\ \algorithmicpardo}
\algrenewtext{EndParFor}{\algorithmicendparfor}

\renewcommand{\baselinestretch}{2}
\renewcommand{\thefootnote}{\normalsize \arabic{footnote}}

\begin{document}
\bigskip

\begin{center} {\LARGE Distributed Newton Methods for Deep Neural Networks}
\end{center}
\begin{center}
{\bf \large Chien-Chih Wang,$^{1}$ Kent Loong Tan,$^{1}$ Chun-Ting Chen,$^{1}$ Yu-Hsiang Lin,$^{2}$ S.~Sathiya Keerthi,$^{3}$ Dhruv Mahajan,$^{4}$ S.~Sundararajan,$^{3}$ Chih-Jen Lin$^{1}$}\\
{$^{1}$Department of Computer Science, National Taiwan University, Taipei 10617, Taiwan}\\
{$^{2}$Department of Physics, National Taiwan University, Taipei 10617, Taiwan}\\
{$^{3}$Microsoft}\\
{$^{4}$Facebook Research}\\
\end{center}
\renewcommand{\thefootnote}{\fnsymbol{footnote}}
\renewcommand{\thefootnote}{\normalsize \arabic{footnote}}

\begin{abstract}
Deep learning involves a difficult non-convex optimization problem with a large number of weights between any two adjacent layers of a deep structure.
To handle large data sets or complicated networks, distributed training is needed, but the calculation of function, gradient, 
and Hessian is expensive.
In particular, the communication and the synchronization cost may become a bottleneck. In this paper, we focus on situations where the model is distributedly stored, and propose
a novel distributed Newton method for training deep neural networks.
By variable and feature-wise data partitions, and some careful designs, we are able to explicitly use the Jacobian matrix for matrix-vector products in the Newton method.
Some techniques are incorporated to reduce the running time as well as the memory consumption.
First, to reduce the communication cost, we propose a diagonalization method such that an approximate Newton direction can be obtained without communication between machines.
Second, we consider subsampled Gauss-Newton matrices for reducing the running time as well as the communication cost.
Third, to reduce the synchronization cost, we terminate the process of finding an approximate Newton direction even though some nodes have not finished their tasks. 
Details of some implementation issues in distributed environments are thoroughly investigated.
Experiments demonstrate that the proposed method is effective for the distributed training of deep neural networks.
In compared with stochastic gradient methods, it is more robust and may give better test accuracy.
\end{abstract}

{\bf Keywords:} Deep Neural Networks, Distributed Newton methods, Large-scale classification, Subsampled Hessian.
\markboth{}{NC instructions}
\ \vspace{-0mm}\\

\section{Introduction}
\label{sec:Introduction}
Recently deep learning has emerged as a useful technique for data classification as well as finding feature representations.
We consider the scenario of multi-class classification. A deep neural network maps each feature vector to one of the class labels by the connection of nodes
in a multi-layer structure. Between two adjacent layers a weight matrix maps the inputs (values in the previous layer) to the outputs (values in the current layer).
Assume the training set includes $(\by^i,\bx^i)$, $i=1,\ldots,l$, where $\bx^i \in \Re^{n_0}$ is the feature vector and $\by^i \in \Re^K$ is the label vector. If $\bx^i$ is associated with label $k$, then
\begin{equation*}
    \by^i = [\underbrace{0,\ldots,0}_{k-1},1,0,\ldots,0]^T \in \Re^K,
\end{equation*}
where $K$ is the number of classes and $\{1,\ldots,K\}$ are possible labels.
After collecting all weights and biases as the model vector $\btheta$ and having a loss function $\xi(\btheta; \bx, \by)$, a neural-network problem can be written as the following optimization problem. 
\begin{equation}
\label{intro-obj}
	\min_{\btheta}\ \ f(\btheta), 
\end{equation}
where
\begin{equation}
\label{intro-obj}
f(\btheta) = \frac{1}{2C} \btheta^T\btheta + \frac{1}{l}\sum_{i=1}^l \xi(\btheta;\bx^i,\by^i).
\end{equation}
The regularization term $\btheta^T\btheta/2$ avoids overfitting the training data, while the parameter $C$ balances the regularization term and the loss term.
The function $f(\btheta)$ is non-convex because of the connection between weights in different layers. This non-convexity and the large number of weights have caused 
tremendous difficulties in training large-scale deep neural networks.
To apply an optimization algorithm for solving \eqref{intro-obj}, the calculation of function, gradient, and Hessian can be expensive. Currently,
stochastic gradient (SG) methods are the most commonly used way to train deep neural networks \citep[e.g.,][]{LB91a,YL98b,LB10a,MZ10a,JD12a,PM15a}. In particular, some expensive operations can be efficiently
conducted in GPU environments \citep[e.g.,][]{DCC10a,AK12b,GEH12a}. Besides stochastic gradient methods, some works such as \cite{JM10a,RK13a,XH16a} have considered a Newton method of 
using Hessian information. Other optimization methods such as ADMM have also been considered \citep{GT16a}.
\par When the model or the data set is large, distributed training is needed. Following the design of the objective function in \eqref{intro-obj}, we note it is easy to achieve data parallelism: if data instances are stored in different computing nodes, then each machine can calculate the local sum of training losses independently.\footnote{Training deep neural networks with data parallelism has been considered in SG, Newton and other optimization methods. For example, \cite{KH15a} implement a parallel Newton method by letting each node store a subset of instances.} However, achieving model parallelism is more difficult because of the complicated structure of deep neural networks. In this work, by considering that the model is distributedly stored we propose a novel distributed Newton method for deep learning.
By variable and feature-wise data partitions, and some careful designs, we are able to explicitly use the Jacobian matrix for matrix-vector products in the Newton method.
Some techniques are incorporated to reduce the running time as well as the memory consumption.
First, to reduce the communication cost, we propose a diagonalization method such that an approximate Newton direction can be obtained without communication between machines.
Second, we consider subsampled Gauss-Newton matrices for reducing the running time as well as the communication cost.
Third, to reduce the synchronization cost, we terminate the process of finding an approximate Newton direction even though some nodes have not finished their tasks.
\par To be focused, among the various types of neural networks, we consider the standard feedforward networks in this work. We do not consider other types
such as the convolution networks that are popular in computer vision.
\par This work is organized as follows. Section \ref{sec:Hessian-free-Deep} introduces existing Hessian-free Newton methods for deep learning.
In Section \ref{sec:Distributed-Deep}, we propose a distributed Newton method for training neural networks. We then develop novel techniques in Section \ref{sec:Reduce-compu-communi}
to reduce running time and memory consumption. In Section \ref{sec:Analyze-algo} we analyze the cost of the proposed algorithm.
Additional implementation techniques are given in Section \ref{sec:other-implement}. 
Then Section \ref{sec:Other-optimization} reviews some existing optimization methods, while experiments in Section \ref{sec:deep-exps} demonstrate the effectiveness of the proposed method.
Programs used for experiments in this paper are available at
\begin{center}\url{http://www.csie.ntu.edu.tw/~cjlin/papers/dnn}.\end{center}
Supplementary materials including a list of symbols and additional experiments can be found at the same web address.

\section{Hessian-free Newton Method for Deep Learning}
\label{sec:Hessian-free-Deep}
In this section, we begin with introducing feedforward neural networks and then review existing Hessian-free Newton methods to solve the optimization problem.
\subsection{Feedforward Networks}
\label{subsec:Feedforward}
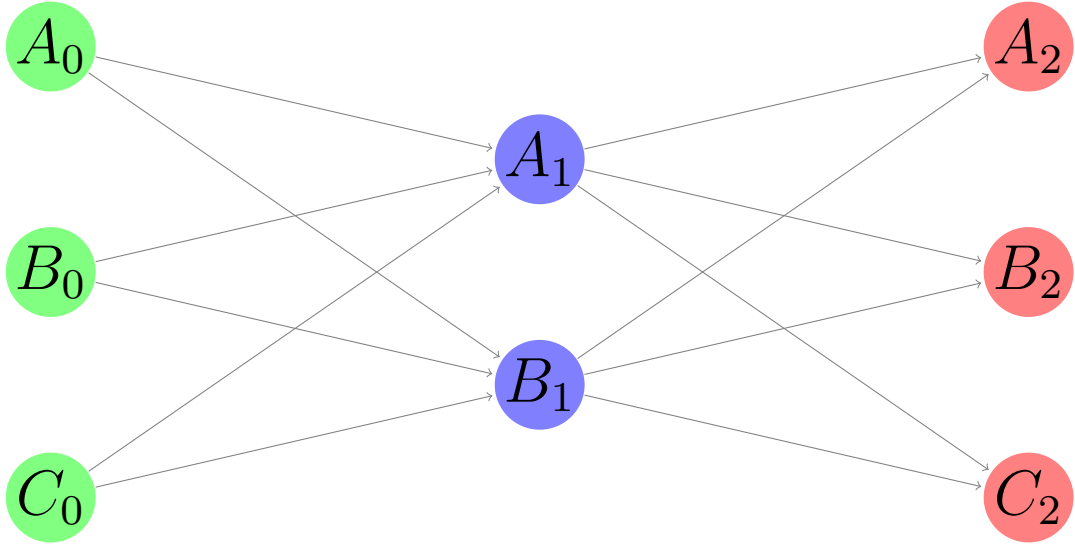
\begin{figure}[t]
\begin{center}
\def\layersep{1cm}

\begin{center}
\begin{tikzpicture}[shorten >=1pt,->,draw=black!50, node distance=\layersep]
    \tikzstyle{every pin edge}=[<-,shorten <=1pt]
    \tikzstyle{neuron}=[circle,fill=black!25,minimum size=17pt,inner sep=0pt]
    \tikzstyle{input neuron}=[neuron, fill=green!50];
    \tikzstyle{output neuron}=[neuron, fill=red!50];
    \tikzstyle{hidden neuron}=[neuron, fill=blue!50];
    \tikzstyle{slave}=[rectangle,fill=green!25,minimum width = 1.5em, minimum height = 1.5em]
    \tikzstyle{annot} = [text width=4em, text centered]


    \node[input neuron,scale=2] (I-1) at (-6cm,-2cm) {$A_0$};
    \node[input neuron,scale=2] (I-2) at (-6cm,-5cm) {$B_0$};
    \node[input neuron,scale=2] (I-3) at (-6cm,-8cm) {$C_0$};

    \node[hidden neuron,scale=2] (H-1) at (0.5cm,-3.5cm) {$A_1$};
    \node[hidden neuron,scale=2] (H-2) at (0.5cm,-6.5cm) {$B_1$};
    
    \node[output neuron,scale=2] (O-1) at (7cm,-2cm) {$A_2$};
    \node[output neuron,scale=2] (O-2) at (7cm,-5cm) {$B_2$};
    \node[output neuron,scale=2] (O-3) at (7cm,-8cm) {$C_2$};

    \path (I-1) edge (H-1);
    \path (I-1) edge (H-2);
    \path (I-2) edge (H-1);
    \path (I-2) edge (H-2);
    \path (I-3) edge (H-1);
    \path (I-3) edge (H-2);

    \path (H-1) edge (O-1);
    \path (H-1) edge (O-2);
    \path (H-1) edge (O-3);
    \path (H-2) edge (O-1);
    \path (H-2) edge (O-2);
    \path (H-2) edge (O-3);

   
\end{tikzpicture}
\end{center}
\caption{An example of feedforward neural networks.\protect\footnotemark}
\label{fig:nnexample}
\end{center}
\end{figure}

A multi-layer neural network maps each feature vector to a class vector via the connection of nodes. There is
a weight vector between two adjacent layers to map the input vector (the previous layer) to the output vector (the current layer).
The network in Figure \ref{fig:nnexample} is an example.
\footnotetext{This figure is modified from the example at \url{http://www.texample.net/tikz/examples/neural-network}.}
Let $n_m$ denote the number of nodes at the $m$th layer. We use $n_0 \text{(input)-}n_1\text{-}\ldots\text{-}n_L(\text{output})$
to represent the structure of the network.\footnote{Note that $n_0$ is the number of features and
$n_L = K$ is the number of classes.\label{n_0}}
The weight matrix $W^m$ and the bias vector $\bb^m$ at the $m$th layer are
\begin{equation*}
W^{m} =
\begin{bmatrix}
w^m_{11}& w^m_{12}& \cdots & w^m_{1n_{m}}\\
w^m_{21}& w^m_{22}& \cdots & w^m_{2n_{m}}\\
\vdots& \vdots& \vdots & \vdots\\
w^m_{n_{m-1}1}& w^m_{n_{m-1}2}& \cdots & w^m_{n_{m-1}n_{m}}
\end{bmatrix}_{n_{m-1} \times n_{m}} \text{and}\quad \  
\bb^m = \begin{bmatrix}
b^m_1\\
b^m_2\\
\vdots\\
b^m_{n_m}
\end{bmatrix}_{n_m \times 1}.
\label{w-and-bias}
\end{equation*}
Let 
\begin{equation*}
\label{input-data}
\bs^{0,i} = \bz^{0,i} = \bx^i
\end{equation*}
be the feature vector for the $i$th instance, and $\bs^{m,i}$ and $\bz^{m,i}$ denote vectors 
of the $i$th instance at the $m$th layer, respectively. We can use 
\begin{align}
       \bs^{m,i} &= (W^m)^T \bz^{m-1,i} + \bb^m,\ m=1,\ldots,L,\ i=1,\ldots,l \nonumber\\
       z_j^{m,i} &= \sigma(s^{m,i}_j),\ j=1,\ldots,n_m,\ m = 1,\ldots,L,\ i=1,\ldots,l\label{xtoz}
\end{align}
to derive the value of the next layer, where $\sigma(\cdot)$ is the activation function.

If $W^m$'s columns are concatenated to the following vector
\begin{equation*}
\bw^m = \begin{bmatrix} w^m_{11} & \ldots & w^m_{n_{m-1}1} & w^m_{12} & \ldots & w^m_{n_{m-1}2} & \ldots & w^m_{1n_{m}} & \ldots & w^m_{n_{m-1}n_{m}} \end{bmatrix}^T,
\end{equation*}
then we can define
\begin{equation*}
	\btheta = \begin{bmatrix} \bw^1 \\ \bb^1 \\ \vdots \\ \bw^L \\ \bb^L\end{bmatrix}
\end{equation*}
as the weight vector of a whole deep neural network. The total number of parameters is 
\begin{equation*}
	n = \sum_{m=1}^L \left(n_{m-1} \times n_m + n_m\right).
\end{equation*}
Because $\bz^{L,i}$ is the output vector of the $i$th data, by a loss function to compare it with
the label vector $\by^i$, a neural network solves the following regularized optimization problem
\begin{equation*}
\label{square-loss}
\min_{\btheta}\ f(\btheta),
\end{equation*}
where
\begin{equation}
\label{obj-function}
f(\btheta) = \frac{1}{2C} \btheta^T\btheta + \frac{1}{l} \sum_{i=1}^{l} \xi(\bz^{L,i};\by^i),
\end{equation}
$C > 0$ is a regularization parameter, and $\xi(\bz^{L,i};\by^i)$ is a convex function of $\bz^{L,i}$. 
Note that we rewrite the loss function $\xi(\btheta;\bx^i,\by^i)$ in \eqref{intro-obj} as $\xi(\bz^{L,i};\by^i)$ because $\bz^{L,i}$
is decided by $\btheta$ and $\bx^i$.
In this work, we consider the following loss function
\begin{equation}
\label{thiswork-loss}
    \xi(\bz^{L,i};\by^i) = || \bz^{L,i} - \by^i ||^2.
\end{equation}
\par The gradient of $f(\btheta)$ is
\begin{equation}
\label{whole-gradient}
	\nabla f(\btheta) =  \frac{1}{C}{\btheta} + \frac{1}{l} \sum_{i=1}^l (J^i)^T \nabla_{\bz^{L,i}} \xi(\bz^{L,i};\by^i),
\end{equation}
where
\begin{equation}
\label{jacobian}
J^i =
\begin{bmatrix}
\frac{\partial z_1^{L,i}}{\partial \theta_1} & \cdots & \frac{\partial z_1^{L,i}}{\partial \theta_n} \\
\vdots & \vdots & \vdots\\
\frac{\partial z_{n_L}^{L,i}}{\partial \theta_1} & \cdots & \frac{\partial z_{n_L}^{L,i}}{\partial \theta_n}
\end{bmatrix}_{n_{L} \times n },\ i=1,\ldots,l, 
\end{equation}
is the Jacobian of $\bz^{L,i}$, which is a function of $\theta$.
The Hessian matrix of $f(\btheta)$ is
\begin{align}
\label{hessiantogauss}
\nabla^2 f(\btheta) =&  \frac{1}{C} \mathcal I + \frac{1}{l} \sum_{i=1}^l 
(J^i)^T B^{i}J^i
\nonumber\\
&+\frac{1}{l} \sum_{i=1}^l\sum_{j=1}^{n_L} \frac{\partial \xi(\bz^{L,i};\by^i)}{\partial z_j^{L,i}}
\begin{bmatrix}
\frac{\partial^2 z_j^{L,i}}{\partial \theta_1 \partial \theta_1} & \cdots & \frac{\partial^2 z_j^{L,i}}{\partial \theta_1 \partial \theta_n}\\
\vdots & \ddots & \vdots\\
\frac{\partial^2 z_j^{L,i}}{\partial \theta_n \partial \theta_1} & \cdots & \frac{\partial^2 z_j^{L,i}}{\partial \theta_n \partial \theta_n}
\end{bmatrix},
\end{align}
where $\mathcal I$ is the identity matrix and
\begin{equation}
	B^{i}_{ts} = \frac{\partial^2 \xi(\bz^{L,i};\by^i)}{\partial z_t^{L,i} \partial z_s^{L,i}},\ t=1,\ldots,n_L,\ s=1,\ldots,n_L.
\label{Bts}
\end{equation}
From now on for simplicity we let
\begin{equation*}
\xi_i \equiv \xi_i(\bz^{L,i};\by^i).
\end{equation*}
\subsection{Hessian-free Newton Method}
\label{subsec:Hessian-free}
For the standard Newton methods, at the $k$th iteration, we find a direction $\bd^k$ 
minimizing the following second-order approximation of the function value:
\begin{equation}
\label{min-hessian}
\min_{\bd}\quad \frac{1}{2} \bd^T H^k \bd + \nabla f(\btheta^k)^T\bd,
\end{equation}
where $H^k = \nabla^2 f(\btheta^k)$ is the Hessian matrix of $f(\btheta^k)$.
To solve \eqref{min-hessian}, first we calculate the gradient vector by a backward process based on \eqref{xtoz} through the following equations:
\begin{align}
&\frac{\partial \xi_i}{\partial s^{m,i}_j} = \frac{\partial \xi_i}{\partial z^{m,i}_j} \sigma' (s^{m,i}_j),\ i=1,\ldots,l,\ m=1,\ldots,L,\ j=1,\ldots,n_m\label{ztox-g}\\
&\frac{\partial \xi_i}{\partial z^{m-1,i}_t} = \sum_{j=1}^{n_m} \frac{\partial \xi_i}{\partial s^{m,i}_j}w^m_{tj},\ i=1,\ldots,l,\ m=1,\ldots,L,\ t=1,\ldots,n_{m-1}\label{communi-g}\\
&\frac{\partial f}{\partial w^m_{tj}}= \frac{1}{C} w^m_{tj} + \frac{1}{l} \sum_{i=1}^l \frac{\partial \xi_i}{\partial s^{m,i}_j} z^{m-1,i}_t,\ m=1,\ldots,L,\ j=1,\ldots,n_m,\ t=1,\ldots,n_{m-1} \label{xtow-g}\\
&\frac{\partial f}{\partial b^m_{j}} = \frac{1}{C} b^m_{j} + \frac{1}{l} \sum_{i=1}^l \frac{\partial \xi_i}{\partial s^{m,i}_j},\ m=1,\ldots,L,\ j=1,\ldots,n_m. \label{xtow-g-bias}
\end{align}
Note that formally the summation in \eqref{xtow-g} should be
\begin{equation*}
\sum_{i=1}^l \sum_{i'=1}^l \frac{\partial \xi_i}{\partial s_j^{m,i'}} z_t^{m-1,i'},
\end{equation*}
but it is simplified because $\xi_i$ is associated with only $s_j^{m,i}$.
\par If $H^k$ is positive definite, then \eqref{min-hessian} is equivalent to solving the following linear system:
\begin{equation}
\label{full-newton-noLM}
H^k  \bd = - \nabla f(\btheta^k).
\end{equation}
Unfortunately, for the optimization problem \eqref{min-hessian}, it is well known that the objective function may be non-convex and therefore
$H^k$ is not guaranteed to be positive definite.
Following \cite{NNS02a}, we can use the Gauss-Newton matrix as 
an approximation of the Hessian. That is, we remove the last term in \eqref{hessiantogauss} and obtain the following positive-definite matrix.
\begin{equation}
\label{def-gauss-newton}
G = \frac{1}{C} \mathcal I + \frac{1}{l}\sum_{i=1}^l (J^i)^T B^i J^i. 
\end{equation}
Note that from \eqref{Bts}, each $B^i$, $i=1,\ldots,l$ is positive semi-definite if we require that $\xi(\bz^{L,i};\by^i)$ is a convex function of $\bz^{L,i}$.
Therefore, instead of using \eqref{full-newton-noLM}, we solve the following linear system to find a $\bd^k$ for deep neural networks.
\begin{equation}
\label{full-gaussnewton}
(G^k  + \lambda_k \mathcal I) \bd = - \nabla f(\btheta^k),
\end{equation}
where $G^k$ is the Gauss-Newton matrix at the $k$th iteration and we add a term $\lambda_k \mathcal{I}$ because of considering 
the Levenberg-Marquardt method (see details in Section \ref{subsec:summary-procedure}).
\par For deep neural networks, because the total number of weights may be very large, it is hard to store the Gauss-Newton matrix. 
Therefore, Hessian-free algorithms have been applied to solve \eqref{full-gaussnewton}. 
Examples include \cite{JM10a, JN11a}. Specifically, conjugate gradient (CG) methods 
are often used so that a sequence of Gauss-Newton matrix vector products are conducted. 
\cite{JM10a, CCW15a} use $\mathcal R$-operator \citep{BAP94a} to implement the product without storing the Gauss-Newton matrix.

Because the use of $\mathcal{R}$ operators for the Newton method is not
the focus of this work, we leave some detailed discussion 
in Sections II--III in supplementary materials. 

\section{Distributed Training by Variable Partition} 
\label{sec:Distributed-Deep}
The main computational bottleneck in a Hessian-free Newton method is the sequence of matrix-vector
products in the CG procedure. To reduce the running time, parallel matrix-vector multiplications should be conducted.
However, the $\mathcal R$ operator discussed in Section \ref{sec:Hessian-free-Deep} and Section II in supplementary materials is inherently sequential.
In a forward process results in the current layer must be finished before the next. 
In this section, we propose an effective distributed algorithm for training deep neural networks.

\subsection{Variable Partition}
\label{subsec:Variable}
\begin{figure}[t]
\begin{center}
\def\layersep{1cm}

\begin{center}
\begin{tikzpicture}[shorten >=1pt,->,draw=black!50, node distance=\layersep]
    \tikzstyle{every pin edge}=[<-,shorten <=1pt]
    \tikzstyle{neuron}=[circle,fill=black!25,minimum size=17pt,inner sep=0pt]
    \tikzstyle{input neuron}=[neuron, fill=green!50];
    \tikzstyle{output neuron}=[neuron, fill=red!50];
    \tikzstyle{hidden neuron}=[neuron, fill=blue!50];
    \tikzstyle{slave}=[rectangle,fill=green!25,minimum width = 1.5em, minimum height = 1.5em]
    \tikzstyle{annot} = [text width=3em, text centered]


    \node[input neuron,scale=2.5] (I-1) at (-6cm,-2cm) {\small $A_0$};
    \node[input neuron,scale=2.5] (I-2) at (-6cm,-5cm) {\small $B_0$};
    \node[input neuron,scale=2.5] (I-3) at (-6cm,-8cm) {\small $C_0$};

    \node[slave,scale=1.5] (S-1) at (-3.25cm, 0cm)   {\small $A_0$,\small $A_1$};
    \node[slave,scale=1.5] (S-2) at (-3.25cm, -2cm)  {\small $A_0$,\small $B_1$};
    \node[slave,scale=1.5] (S-3) at (-3.25cm, -4cm)  {\small $B_0$,\small $A_1$};
    \node[slave,scale=1.5] (S-4) at (-3.25cm, -6cm)  {\small $B_0$,\small $B_1$};
    \node[slave,scale=1.5] (S-5) at (-3.25cm, -8cm)  {\small $C_0$,\small $A_1$};
    \node[slave,scale=1.5] (S-6) at (-3.25cm, -10cm) {\small $C_0$,\small $B_1$};

    \node[hidden neuron,scale=2.5] (H-1) at (0.5cm,-3.5cm) {\small $A_1$};
    \node[hidden neuron,scale=2.5] (H-2) at (0.5cm,-6.5cm) {\small $B_1$};
    
    \node[slave,scale=1.5] (S2-1) at (4.25cm, 0cm)   {\small $A_1$,\small $A_2$};
    \node[slave,scale=1.5] (S2-2) at (4.25cm, -2cm)  {\small $A_1$,\small $B_2$};
    \node[slave,scale=1.5] (S2-3) at (4.25cm, -4cm)  {\small $A_1$,\small $C_2$};
    \node[slave,scale=1.5] (S2-4) at (4.25cm, -6cm)  {\small $B_1$,\small $A_2$};
    \node[slave,scale=1.5] (S2-5) at (4.25cm, -8cm)  {\small $B_1$,\small $B_2$};
    \node[slave,scale=1.5] (S2-6) at (4.25cm, -10cm) {\small $B_1$,\small $C_2$};

    \node[output neuron,scale=2.5] (O-1) at (7cm,-2cm) {\small $A_2$};
    \node[output neuron,scale=2.5] (O-2) at (7cm,-5cm) {\small $B_2$};
    \node[output neuron,scale=2.5] (O-3) at (7cm,-8cm) {\small $C_2$};

    \path (I-1) edge (S-1);
    \path (I-1) edge (S-2);
    \path (I-2) edge (S-3);
    \path (I-2) edge (S-4);
    \path (I-3) edge (S-5);
    \path (I-3) edge (S-6);

    \path (S-1) edge (H-1);
    \path (S-3) edge (H-1);
    \path (S-5) edge (H-1);
    \path (S-2) edge (H-2);
    \path (S-4) edge (H-2);
    \path (S-6) edge (H-2);

    \path (H-1) edge (S2-1);
    \path (H-1) edge (S2-2);
    \path (H-1) edge (S2-3);
    \path (H-2) edge (S2-4);
    \path (H-2) edge (S2-5);
    \path (H-2) edge (S2-6);

    \path (S2-1) edge (O-1);
    \path (S2-4) edge (O-1);
    \path (S2-2) edge (O-2);
    \path (S2-5) edge (O-2);
    \path (S2-3) edge (O-3);
    \path (S2-6) edge (O-3);

   
\end{tikzpicture}
\end{center}
\caption{An example of splitting variables in Figure \ref{fig:nnexample} to $12$ partitions by a split structure of $3$-$2$-$3$. 
Each circle corresponds to a neuron sub-group in a layer, while each square is a partition corresponding to weights connecting one neuron sub-group
in a layer to one neuron sub-group in the next layer.}
\label{fig:neuron}
\end{center}
\end{figure}
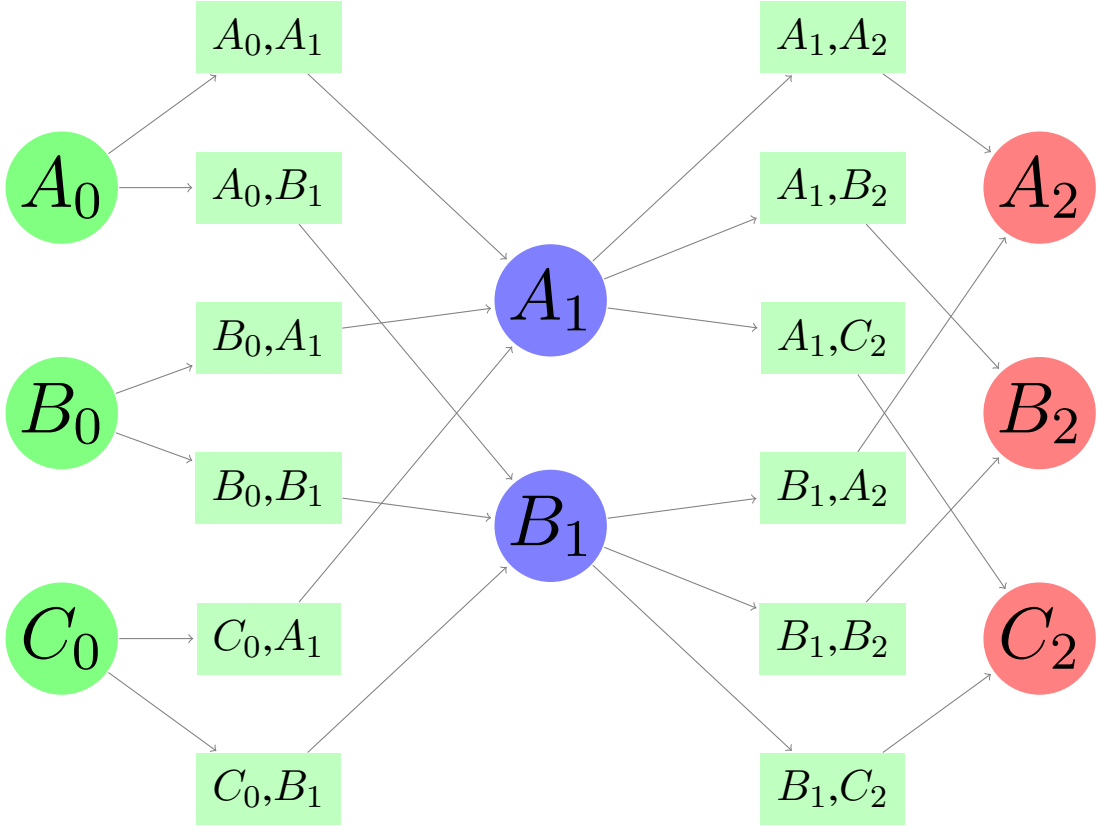

Instead of using the $\mathcal R$ operator to calculate the matrix-vector product, we consider the whole Jacobian matrix and
directly use the Gauss-Newton matrix in \eqref{def-gauss-newton} for the matrix-vector products in the CG procedure. 
This setting is possible because of the following reasons.
\begin{enumerate}[1.]
\item A distributed environment is used.
\item With some techniques we do not need to explicitly store every element of the Jacobian matrix.
\end{enumerate}
Details will be described in the rest of this paper.
To begin we split each $J^i$ to $P$ partitions
\begin{equation*}
	J^i = \begin{bmatrix} J^i_1 & \cdots & J^i_P\end{bmatrix}.
\end{equation*}
Because the number of columns in $J^i$ is the same as the number of variables in the optimization problem,
essentially we partition the variables to $P$ subsets. Specifically, we split neurons in each layer to several groups.
Then weights connecting one group of the current layer to one group of the next layer form a subset of our variable partition.
For example, assume we have a $150$-$200$-$30$ neural network in Figure \ref{fig:neuron}. By splitting the three layers to $3$, $2$, $3$ groups,
we have a total number of partitions $P = 12$. The partition $(A_0,A_1)$ in Figure \ref{fig:neuron} is responsible for a $50\times100$ sub-matrix of $W^1$. 
In addition, we distribute the variable $\bb^m$ to partitions corresponding to the first neuron sub-group of the $m$th layer.
For example, the $200$ variables of $\bb^1$ is split to $100$ in the partition $(A_0,A_1)$ and $100$ in the partition $(A_0,B_1)$.
\par By the variable partition, we achieve model parallelism. Further, because $\bz^{0,i} = \bx^{i}$ from \eqref{input-data}, our data points are split in a feature-wise way to nodes corresponding to partitions between layers 0 and 1. Therefore, we have data parallelism.

\par With the variable partition, the second term in the Gauss-Newton matrix \eqref{def-gauss-newton} for the $i$th instance can be represented as
\begin{equation*}
    (J^i)^T B^{i} J^i = 
\begin{bmatrix}
    (J^i_{1})^T B^{i} J^i_{1} & \cdots & (J^i_{1})^T B^{i} J^i_{P} \\
        & \ddots & \\
    (J^i_{P})^T B^{i} J^i_{1} & \cdots & (J^i_{P})^T B^{i} J^i_{P}
\end{bmatrix}.
\end{equation*}
In the CG procedure to solve \eqref{full-gaussnewton}, the product between the Gauss-Newton matrix and a vector $\bv$ is

\begin{equation}
\label{real-jaco}
G\bv = \begin{bmatrix} \frac{1}{l}\sum_{i=1}^l (J^i_{1})^T B^{i} (\sum_{p=1}^P J^i_{p}\bv_p) + \frac{1}{C}\bv_1\\ \vdots 
        \\ \frac{1}{l}\sum_{i=1}^l (J^i_{P})^T B^{i} (\sum_{p=1}^P J^i_{p}\bv_p) + \frac{1}{C}\bv_P \end{bmatrix}
        , \text{where}\ \bv = \begin{bmatrix} \bv_{1} \\ \vdots \\ \bv_{P} \end{bmatrix}
\end{equation}
is partitioned according to our variable split.
From \eqref{Bts} and the loss function defined in \eqref{thiswork-loss},
\begin{equation*}
B^i_{ts} 
= \frac{\partial^2 \left(\sum_{j=1}^{n_L} (z^{L,i}_j - y^i_j)^2 \right)}{\partial z^{L,i}_t \partial z^{L,i}_s} 
= \frac{\partial \left(2(z^{L,i}_t - y^i_t) \right)}{\partial z^{L,i}_s} 
= \begin{cases}
    2 & \text{if } t = s,\\
    0 & \text{otherwise}.
\end{cases}
\end{equation*}

\par However, after the variable partition, each $J^i$ may still be a huge matrix. The total space for storing $J^i_p,\ \forall i$ is roughly
\begin{equation*}
n_L \times \frac{n}{P} \times l.
\end{equation*}
If $l$, the number of data instances, is so large such that 
\begin{equation*}
l \times \frac{n_L}{P} > n,
\end{equation*}
than storing $J^i_p,\ \forall i$ requires more space than the $n \times n$ Gauss-Newton matrix. 
To reduce the memory consumption, we will propose effective techniques in Sections \ref{subsec:Distributed-Jacobian}, \ref{subsec:Subsampled}, and \ref{subsec:pipeline}. 

\par With the variable partition, function, gradient, and Jacobian calculations become complicated. 
We discuss details in Sections \ref{subsec:Distributed-Function} and \ref{subsec:Distributed-Jacobian}.

\subsection{Distributed Function Evaluation}
\label{subsec:Distributed-Function}
\par From \eqref{xtoz} we know how to evaluate the function value in a single machine, but the implementation in a distributed environment is not trivial.
Here we check the details from the perspective of an individual partition. 
Consider a partition that involves neurons in sets $T_{m-1}$ and $T_m$ from layers $m-1$ and $m$, respectively. Thus
\begin{equation*}
T_{m-1} \subset \{1,\ldots,n_{m-1}\}\ \text{and}\ T_{m} \subset \{1,\ldots,n_{m}\}.
\end{equation*}
Because \eqref{xtoz} is a forward process, we assume that
\begin{equation*}
s^{m-1,i}_t,\ i=1,\ldots,l,\ \forall t \in T_{m-1}
\end{equation*}
are available at the current partition. The goal is to generate
\begin{equation*}
s^{m,i}_j,\ i=1,\ldots,l,\ \forall j \in T_m
\end{equation*}
and pass them to partitions between layers $m$ and $m+1$. 
To begin, we calculate
\begin{equation} 
\label{z-alg}
	z^{m-1,i}_t = \sigma(s^{m-1,i}_t),\ i = 1, \ldots, l \text{ and } t \in T_{m-1}.
\end{equation}
Then, from \eqref{xtoz}, the following local values can be calculated for $i=1,\ldots,l,\ j \in T_m$
\begin{equation}
\label{local-weight-sum}
\begin{cases}
\sum_{t \in T_{m-1}} w_{tj}^m z^{m-1,i}_t + b^m_j & \text{if $T_{m-1}$ is the first neuron sub-group of layer $m-1$},\\
\sum_{t \in T_{m-1}} w_{tj}^m z^{m-1,i}_t & \text{otherwise}.\\
\end{cases}
\end{equation}
After the local sum in \eqref{local-weight-sum} is obtained, we must sum up values in partitions between layers $m-1$ and $m$.
\begin{equation}
    s^{m,i}_j = \sum_{T_{m-1} \in P_{m-1}} \Big(\text{ local sum in \eqref{local-weight-sum} }\Big),
\label{dist_eval}
\end{equation}
where $i=1,\ldots,l$, $j \in T_m$, and
\begin{equation*}
	P_{m-1} = \{ T_{m-1} \ |\  T_{m-1} \text{ is any sub-group of neurons at layer $m-1$}\}.
\end{equation*}
The resulting $s^{m,i}_j$ values should be broadcasted to partitions between layers $m$ and $m+1$ that correspond to the neuron subset $T_m$.
We explain details of \eqref{dist_eval} and the {\it broadcast} operation in Section \ref{subsubsec:fun-allreduce}.
\subsubsection{Allreduce and Broadcast Operations}
\label{subsubsec:fun-allreduce}
The goal of \eqref{dist_eval} is to generate and broadcast $s^{m,i}_j$ values to some partitions between layers $m$ and $m+1$, so a {\it reduce} operation seems to be sufficient.
However, we will explain in Section \ref{subsec:Distributed-Jacobian} that for the Jacobian evaluation and then the product
between Gauss-Newton matrix and a vector, the partitions between layers $m-1$ and $m$ corresponding to $T_m$ also need $s^{m,i}_j$ for calculating
\begin{equation}
\label{dist-next}
z^{m,i}_j = \sigma(s^{m,i}_j),\ i=1,\ldots,l,\ j \in T_m.
\end{equation}
To this end, we consider an {\it allreduce} operation so that 
not only are values reduced from some partitions between layers $m-1$ and $m$, but also the result is broadcasted to them. 
After this is done, we make the same result $s^{m,i}_j$ available in partitions between layers $m$ and $m+1$ by choosing
the partition corresponding to the first neuron sub-group of layer $m-1$ to conduct a {\it broadcast} operation.
Note that for partitions between layers $L-1$ and $L$ (i.e., the last layer), a broadcast operation is not needed.
\par Consider the example in Figure \ref{fig:neuron}. For partitions $(A_1,A_2)$, $(A_1,B_2)$, and $(A_1,C_2)$, all of them must get $s^{1,i}_j, j \in A_1$ calculated via \eqref{dist_eval}:
\begin{equation}
s^{1,i}_j = \underbrace{\sum_{t\in A_0} w^1_{tj} z^{0,i}_t + b^1_j}_{(A_0,A_1)}\ +\ 
\underbrace{\sum_{t\in B_0} w^1_{tj} z^{0,i}_t}_{(B_0,A_1)}\ +\ 
\underbrace{\sum_{t\in C_0} w^1_{tj} z^{0,i}_t}_{(C_0,A_1)}.
\label{reduce-sum-eval}
\end{equation}
The three local sums are available at partitions $(A_0,A_1)$, $(B_0,A_1)$ and $(C_0,A_1)$ respectively.
We first conduct an {\it allreduce} operation so that $s^{1,i}_j,\ j \in A_1$ are available at partitions $(A_0,A_1)$, $(B_0,A_1)$, and $(C_0,A_1)$. 
Then we choose $(A_0,A_1)$ to broadcast values to $(A_1,A_2)$, $(A_1,B_2)$, and $(A_1,C_2)$. 

\par Depending on the system configurations, suitable ways
can be considered for implementing the {\it allreduce} and 
the {\it broadcast} operations \citep{RT05a}.
In Section IV of supplementary materials we give details of our implementation.

\par To derive the loss value, we need one final {\it reduce} operation. For the example in Figure \ref{fig:neuron}, 
in the end we have $z^{2,i}_j,\ j \in A_2,\ B_2,\ C_2$ respectively available in partitions 
\begin{equation*}
(A_1,A_2),\ (A_1,B_2),\ \text{and } (A_1,C_2).
\end{equation*}
We then need the following {\it reduce} operation
\begin{equation}
\label{distri-loss-sum}
||\bz^{2,i} - \by^{i}||^2 = \sum_{j \in A_2} (z^{2,i}_j - y^i_j)^2 + \sum_{j \in B_2} (z^{2,i}_j - y^i_j)^2 + \sum_{j \in C_2} (z^{2,i}_j - y^i_j)^2
\end{equation}
and let $(A_1,A_2)$ have the loss term in the objective value.
\par We have discussed the calculation of the loss term in the objective value, but we also need to obtain the regularization term $\btheta^T \btheta/2$. 
One possible setting is that before the loss-term calculation we run a {\it reduce} operation to sum up all local regularization terms.
For example, in one partition corresponding to neuron subgroups $T_{m-1}$ at layer $m-1$ and $T_m$ at layer $m$, the local value is
\begin{equation}
\label{local-reg}
\sum_{t \in T_{m-1}} \sum_{j \in T_m} (w^{m}_{tj})^2.
\end{equation}
On the other hand, we can embed the calculation into the forward process for obtaining the loss term. The idea is that we append the local regularization term in \eqref{local-reg}
to the vector in \eqref{local-weight-sum} for an {\it allreduce} operation in \eqref{dist_eval}. The cost is negligible because we only increase the length
of each vector by one. After the {\it allreduce} operation, we broadcast the resulting vector to partitions between layers $m$ and $m+1$ that corresponding to the neuron subgroup $T_m$.
We cannot let each partition collect the broadcasted value for subsequent {\it allreduce} operations because regularization terms in previous layers would be calculated several times.
To this end, we allow only the partition corresponding to $T_{m}$ in layer $m$ and the first neuron subgroup in layer $m+1$ to collect the value and include it with the local regularization term
for the subsequent {\it allreduce} operation. By continuing the forward process, in the end we get the whole regularization term.
\par We use Figure \ref{fig:neuron} to give an illustration. The {\it allreduce} operation in \eqref{reduce-sum-eval} now also calculates 
\begin{equation}
\underbrace{\sum_{t\in A_0} \sum_{j \in A_1} (w^1_{tj})^2 + \sum_{j \in A_1} (b^1_{j})^2}_{(A_0,A_1)}+
\underbrace{\sum_{t\in B_0} \sum_{j \in A_1} (w^1_{tj})^2}_{(B_0,A_1)}+
\underbrace{\sum_{t\in C_0} \sum_{j \in A_1} (w^1_{tj})^2}_{(C_0,A_1)}.
\label{reduce-sum-reg-eval}
\end{equation}
The resulting value is broadcasted to
\begin{equation*}
(A_1,A_2),\ (A_1,B_2),\ \text{and } (A_1,C_2).
\end{equation*}
Then only $(A_1,A_2)$ collects the value and generate the following local sum:
\begin{equation*}
\eqref{reduce-sum-reg-eval}
+\sum_{t \in A_1} \sum_{j \in A_2} (w^2_{tj})^2 + \sum_{j \in A_2} (b^2_{j})^2.
\end{equation*}
In the end we have
\begin{enumerate}[1.]
\item
$(A_1,A_2)$ contains regularization terms from
\begin{equation*}
(A_0,A_1),\ (B_0,A_1),\ (C_0,A_1),\ (A_1,A_2),\ (A_0,B_1),\ (B_0,B_1),\ (C_0,B_1),\ (B_1,A_2). 
\end{equation*}
\item
$(A_1,B_2)$ contains regularization terms from
\begin{equation*}
(A_1,B_2),\ (B_1,B_2).
\end{equation*}
\item
$(A_1,C_2)$ contains regularization terms from
\begin{equation*}
(A_1,C_2),\ (B_1,C_2).
\end{equation*}
\end{enumerate}
We can then extend the reduce operation in \eqref{distri-loss-sum} to generate the final value of the regularization term.
\renewcommand{\baselinestretch}{1.3}
\begin{algorithm}[t]
  \caption{Function evaluation in a distributed system}
  \begin{algorithmic}[1]
		\State Let $T_{m-1}$ and $T_m$ be the subsets of neurons at the $(m-1)$th and $m$th layers corresponding to the current partition.
        \If{$m = 1$}
        \State Read $s^{m-1,i}_t$ from input, where $i=1,\ldots,l$, and $t \in T_{m-1}$.
        \Else
        \State Wait for $s^{m-1,i}_t$, $i=1,\ldots,l,\ t \in T_{m-1}$. 
		\State Calculate $z^{m-1,i}_t$ by \eqref{z-alg}. 
        \EndIf
		\State After calculating \eqref{local-weight-sum}, run an {\it allreduce} operation to have
			\begin{equation}
			\label{s-alg} 
				s^{m,i}_j,\ i = 1, \ldots, l \text{ and } j \in T_m,
			\end{equation}
	    \Statex available in all partitions between layers $m-1$ and $m$ corresponding to $T_m$.
		\If{$T_{m-1}$ is the first neuron sub-group of layer $m-1$}
        	\If{$m < L$}
				\State We broadcast values in \eqref{s-alg} to partitions between layers $m$ and $m+1$
                \Statex \indent \indent \hspace{2ex} corresponding to the neuron subgroup $T_{m}$; see the description after \eqref{reduce-sum-eval}
			\Else
				\State Calculate
				\begin{equation*}
					\sum_{i=1}^l \sum_{j \in T_L} \xi(z^{L,i}_j;y^i_j) + \text{accumulated regularization terms}
				\end{equation*}
				\State If $T_L$ is the first neuron sub-group of layer $L$, run a {\it reduce} operation 
				\Statex \indent \indent \hspace{2ex} to get the final $f$; see \eqref{distri-loss-sum}.
			\EndIf
        \EndIf
  \end{algorithmic}
  \label{alg:eval}
\end{algorithm}
\renewcommand{\baselinestretch}{2}
\subsection{Distributed Jacobian Calculation}
\label{subsec:Distributed-Jacobian}
From \eqref{jacobian} and similar to the way of calculating the gradient in \eqref{ztox-g}-\eqref{xtow-g-bias}, the Jacobian matrix
satisfies the following properties.
\begin{align}
\frac{\partial z^{L,i}_u}{\partial w^m_{tj}} &= \frac{\partial z^{L,i}_u}{\partial s^{m,i}_j} \frac{\partial s^{m,i}_j}{\partial w^m_{tj}}, \label{xtow-jaco-1}\\
\frac{\partial z^{L,i}_u}{\partial b^m_{j}}  &= \frac{\partial z^{L,i}_u}{\partial s^{m,i}_j} \frac{\partial s^{m,i}_j}{\partial b^m_j}, \label{xtow-jaco-bias-1}
\end{align}
where $i=1,\ldots,l,\ u=1,\ldots,n_L,\ m=1,\ldots,L,\ j=1,\ldots,n_m$, and $t=1,\ldots,n_{m-1}$.
However, these formulations do not reveal how they are calculated in a distributed setting. Similar to Section \ref{subsec:Distributed-Function}, we check details from the 
perspective of any variable partition.
Assume the current partition involves neurons in sets $T_{m-1}$ and $T_{m}$ from layers $m-1$ and $m$, respectively. 
Then we aim to obtain the following Jacobian components.
\begin{equation*}
\frac{\partial z^{L,i}_u}{\partial w^m_{tj}}\ \text{and}\ \frac{\partial z^{L,i}_u}{\partial b^m_j},\ \forall t \in T_{m-1},\ 
\forall j \in T_m,\ u=1,\ldots,n_L,\ i=1,\ldots,l.
\end{equation*}
Before showing how to calculate them, we first get from \eqref{xtoz} that
\begin{align}
 \frac{\partial z^{L,i}_u}{\partial s^{m,i}_j} &= \frac{\partial z^{L,i}_u}{\partial z^{m,i}_j} \frac{\partial z^{m,i}_j}{\partial s^{m,i}_j} =\frac{\partial z^{L,i}_u}{\partial z^{m,i}_j}\sigma' (s^{m,i}_j), \label{ztox-jaco} \\
\frac{\partial s^{m,i}_j}{\partial w^m_{tj}} &= z^{m-1,i}_t \text{ and } \frac{\partial s^{m,i}_j}{\partial b^m_j} = 1 \label{stow-jaco}, \\
\frac{\partial z^{L,i}_u}{\partial z^{L,i}_j} &=
\begin{cases}
1 & \text{if } j = u, \\
0 & \text{otherwise.}
\end{cases}
\label{zLindependent}
\end{align}
From \eqref{xtow-jaco-1}-\eqref{zLindependent}, the elements for the local Jacobian matrix can be derived by
\begin{align}
\frac{\partial z^{L,i}_u}{\partial w^m_{tj}} &= \frac{\partial z^{L,i}_u}{\partial z^{m,i}_j} \frac{\partial z^{m,i}_j}{\partial s^{m,i}_j} \frac{\partial s^{m,i}_j}{\partial w^m_{tj}} = 
\begin{cases} 
\frac{\partial z^{L,i}_u}{\partial z^{m,i}_j} \sigma'(s^{m,i}_j) z^{m-1,i}_t &\text{if}\ m < L,\\
\sigma'(s^{L,i}_u) z^{L-1,i}_t &\text{if}\ m=L,\ j = u,\\
0 &\text{if}\ m=L,\ j \neq u,
\end{cases}\label{xtow-jaco}\\
\intertext{and} 
\frac{\partial z^{L,i}_u}{\partial b^m_{j}} &= \frac{\partial z^{L,i}_u}{\partial z^{m,i}_j} \frac{\partial z^{m,i}_j}{\partial s^{m,i}_j} \frac{\partial s^{m,i}_j}{\partial b^m_{j}} = 
\begin{cases}
\frac{\partial z^{L,i}_u}{\partial z^{m,i}_j} \sigma'(s^{m,i}_j) &\text{if}\ m < L,\\
\sigma'(s^{L,i}_u) &\text{if}\ m=L,\ j = u,\\
0 &\text{if}\ m=L,\ j \neq u,
\end{cases}
\label{xtow-jaco-bias}
\end{align}
where $u = 1,\ldots,n_L$, $i=1,\ldots,l$, $t \in T_{m-1}$, and $j \in T_m$. 
 
\par We discuss how to have values in the right-hand side of \eqref{xtow-jaco} and \eqref{xtow-jaco-bias} available at the current computing node. 
From \eqref{z-alg}, we have
\begin{equation*} 
z^{m-1,i}_t,\ \forall i=1,\ldots,l,\ \forall t \in T_{m-1}
\end{equation*}
available in the forward process of calculating the function value. Further, in \eqref{dist_eval}-\eqref{dist-next} to obtain $z^{m,i}_j$
for layers $m$ and $m+1$, we use an {\it allreduce} operation rather than a {\it reduce} operation so that 
\begin{equation*}
s^{m,i}_j,\ \forall i = 1,\ldots,l,\ \forall j \in T_m
\end{equation*}
are available at the current partition between layers $m-1$ and $m$. Therefore, $\sigma'(s^{m,i}_j)$ in \eqref{xtow-jaco}-\eqref{xtow-jaco-bias}
can be obtained. The remaining issue is to generate $\partial z^{L,i}_u/\partial z^{m,i}_j$.
We will show that they can be obtained by a backward process.
Because the discussion assumes that currently we are at a partition between layers $m-1$ and $m$, we show details of generating $\partial z^{L,i}_u/\partial z^{m-1,i}_t$
and dispatching them to partitions between $m-2$ and $m-1$. 
From \eqref{xtoz} and \eqref{ztox-jaco}, $\partial z^{L,i}_u/z^{m-1,i}_t$ can be calculated by
\begin{equation}
\frac{\partial z^{L,i}_u}{\partial z^{m-1,i}_t} = \sum_{j=1}^{n_m} \frac{\partial z^{L,i}_u}{\partial s^{m,i}_j} \frac{\partial s^{m,i}_j}{\partial z^{m-1,i}_t} 
= \sum_{j=1}^{n_m} \frac{\partial z^{L,i}_u}{\partial z^{m,i}_j} \sigma'(s^{m,i}_j) w^m_{tj}.
\label{communi-jaco}
\end{equation}
Therefore, we consider a backward process of using $\partial z^{L,i}_u/\partial z^{m,i}_j$ to generate $\partial z^{L,i}_u/\partial z^{m-1,i}_t$.
In a distributed system, from \eqref{zLindependent} and \eqref{communi-jaco},
\begin{equation}
\frac{\partial z^{L,i}_u}{\partial z^{m-1,i}_t} =
\begin{cases}
\sum_{T_m \in P_m} \sum_{j \in T_m} \frac{\partial z^{L,i}_u}{\partial z^{m,i}_j} \sigma'(s^{m,i}_j) w^m_{tj} &\text{if}\ m < L,\\
\sum_{T_m \in P_m} \sigma'(s^{L,i}_u) w^L_{tu}&\text{if}\ m=L,
\end{cases}
\label{jaco-sum}
\end{equation}
where $i=1,\ldots,l,\ u=1,\ldots,n_L,\ t \in T_{m-1}$, and
\begin{equation}
\label{pmdefine}
	P_m = \{ T_m \ |\  T_m \text{ is any sub-group of neurons at layer $m$}\}.
\end{equation}
Clearly, each partition calculates the local sum over $j \in T_m$. Then a {\it reduce} operation is needed to sum up values in all corresponding partitions between layers $m-1$ and $m$.
Subsequently, we discuss details of how to transfer data to partitions between layers $m-2$ and $m-1$.

Consider the example in Figure \ref{fig:neuron}. The partition $(A_0,A_1)$ must get
\begin{equation*}
	\frac{\partial z^{L,i}_u}{\partial z^{1,i}_t},\ t \in A_1,\ u=1,\ldots,n_L,\ i=1,\ldots,l.
\end{equation*}
From \eqref{jaco-sum},
\begin{equation}
    \frac{\partial z^{L,i}_u}{\partial z^{1,i}_t}\ =\ \underbrace{\sum_{j \in A_2} \frac{\partial z^{L,i}_u}{\partial z^{2,i}_j} \sigma'(s^{2,i}_j) w^2_{tj} }_{(A_1,A_2)}
                                                     +\underbrace{\sum_{j \in B_2} \frac{\partial z^{L,i}_u}{\partial z^{2,i}_j} \sigma'(s^{2,i}_j) w^2_{tj} }_{(A_1,B_2)}
    												 +\underbrace{\sum_{j \in C_2} \frac{\partial z^{L,i}_u}{\partial z^{2,i}_j} \sigma'(s^{2,i}_j) w^2_{tj} }_{(A_1,C_2)}.
\label{reduce-sum-jaco}
\end{equation}
Note that these three sums are available at partitions $(A_1,A_2)$, $(A_1,B_2)$, and $(A_1,C_2)$, respectively. Therefore, \eqref{reduce-sum-jaco} is a {\it reduce} operation.
Further, values obtained in \eqref{reduce-sum-jaco} are needed in partitions not only $(A_0,A_1)$ but also $(B_0,A_1)$ and $(C_0,A_1)$. Therefore, we need a {\it broadcast}
operation so values can be available in the corresponding partitions.

For details of implementing {\it reduce} and {\it broadcast} operations, see Section IV of supplementary materials. Algorithm \ref{alg:jaco} summarizes the backward process to calculate $\partial z^{L,i}_u/\partial z^{m,i}_j$. 
\subsubsection{Memory Requirement}
\label{subsubsec:Jmemory}
We have mentioned in Section \ref{subsec:Variable} that storing all elements in the Jacobian matrix may not be viable.
In the distributing setting, if we store all Jacobian elements corresponding to the current partition, then 
\begin{equation}
\label{J-memeory}
|T_{m-1}| \times |T_m| \times n_L \times l
\end{equation} 
space is needed. We propose a technique to save space by noting that \eqref{xtow-jaco-1} can be written as the product of two terms. 
From \eqref{ztox-jaco}-\eqref{stow-jaco}, the first term is related to only $T_m$, while the second is
related to only $T_{m-1}$:
\begin{equation}
\label{Jelement-split}
\frac{\partial z^{L,i}_u}{\partial w^m_{tj}} 
=[\frac{\partial z^{L,i}_u}{\partial s^{m,i}_j}][\frac{\partial s^{m,i}_j}{\partial w^m_{tj}}]
=[\frac{\partial z^{L,i}_u}{\partial z^{m,i}_j} \sigma'(s^{m,i}_j)][z^{m-1,i}_t].
\end{equation}
They are available in our earlier calculation. 
Specifically, we allocate space to receive $\partial z^{L,i}_u/\partial z^{m,i}_j$ from previous layers. After obtaining the values, we
replace them with 
\begin{equation}
\label{Jelement-store}
\frac{\partial z^{L,i}_u}{\partial z^{m,i}_j} \sigma'(s^{m,i}_j)
\end{equation}
for the future use. Therefore, the Jacobian matrix is not explicitly stored. Instead, we use the two terms in \eqref{Jelement-split} for the product between the Gauss-Newton matrix and a vector in the CG procedure. See details in Section \ref{subsec:Jacobian-Vector}.
Note that we also need to calculate and store the local sum 
before the reduce operation in \eqref{jaco-sum} for getting $\partial z^{L,i}_u/\partial z^{m-1,i}_t,\ \forall t \in T_{m-1},\ \forall u,\ \forall i$. 
Therefore, the memory consumption is proportional to
\begin{equation*}
	l \times n_L \times ( |T_{m-1}| + |T_m|).
\end{equation*}
This setting significantly reduces the memory consumption of directly storing the Jacobian matrix in \eqref{J-memeory}.
\subsubsection{Sigmoid Activation Function}
\label{subsubsec:sigmoid}
In the discussion so far, we consider a general differentiable activation function $\sigma(s^{m,i}_j)$. In the implementation in this paper,
we consider the sigmoid function except the output layer:
\begin{equation}
\label{this-paper-activation}
z^{m,i}_j = \sigma(s^{m,i}_j) = \begin{cases}
\frac{1}{1+e^{-s^{m,i}_j}} & \text{ if } m < L,\\
s^{m,i}_j & \text{ if } m = L.
\end{cases}
\end{equation}
Then,
\begin{equation*}
\sigma'(s^{m,i}_j) = \begin{cases}
\frac{e^{-s^{m,i}_j}}{\left(1+e^{-s^{m,i}_j}\right)^2} = z^{m,i}_j (1-z^{m,i}_j) & \text{ if } m < L,\\
1 & \text{ if } m = L.
\end{cases}
\end{equation*}
and \eqref{xtow-jaco}-\eqref{xtow-jaco-bias} become

\begin{equation*}
\frac{\partial z^{L,i}_u}{\partial w^m_{tj}} = 
\begin{cases} 
\frac{\partial z^{L,i}_u}{\partial z^{m,i}_j} z^{m,i}_j (1-z^{m,i}_j) z^{m-1,i}_t,\\
z^{L-1,i}_t,\\
0 ,
\end{cases}\!,
\frac{\partial z^{L,i}_u}{\partial b^m_{j}} = 
\begin{cases}
\frac{\partial z^{L,i}_u}{\partial z^{m,i}_j} z^{m,i}_j (1 - z^{m,i}_j) &\text{if}\ m < L,\\
1 &\text{if}\ m=L,\ j = u,\\
0 &\text{if}\ m=L,\ j \neq u,
\end{cases}
\end{equation*}
where $u = 1,\ldots,n_L$, $i=1,\ldots,l$, $t \in T_{m-1}$, and $j \in T_m$. 

\subsection{Distributed Gradient Calculation}
\label{subsec:distri-gradient}
For the gradient calculation, from \eqref{obj-function}, 
\begin{equation}
\label{distri-grad}
\frac{\partial f}{\partial w^m_{tj}}
=\frac{1}{C} w^m_{tj} + \frac{1}{l} \sum_{i=1}^l \frac{\partial \xi_i}{\partial w^m_{tj}}
=\frac{1}{C} w^m_{tj} + \frac{1}{l} \sum_{i=1}^l \sum_{u=1}^{n_L} \frac{\partial \xi_i}{\partial z^{L,i}_u} \frac{\partial z^{L,i}_u}{\partial w^m_{tj}},
\end{equation}
where $\partial z^{L,i}_u/\partial w^m_{tj},\ \forall t,\ \forall j$ are components of the Jacobian matrix; see also the matrix form in \eqref{whole-gradient}.
From \eqref{xtow-jaco}, we have known how to calculate $\partial z^{L,i}_u/\partial w^m_{tj}$. Therefore, if $\partial \xi_i / \partial z^{L,i}_u$ is passed
to the current partition, we can easily obtain the gradient vector via \eqref{distri-grad}. This can be finished in the same backward process of calculating the Jacobian matrix.
\par On the other hand, in the technique that will be introduced in Section \ref{subsec:Subsampled}, we only consider a subset of instances to construct the Jacobian matrix as well as
the Gauss-Newton matrix. That is, by selecting a subset $S \subset \{1, \ldots,l\}$, then only $J^i, \forall i \in S$ are considered. 
Thus we do not have all the needed $\partial z^{L,i}_u/\partial w^m_{tj}$ for \eqref{distri-grad}. 
In this situation, we can separately consider a backward process to calculate the gradient vector.
From a derivation similar to \eqref{xtow-jaco},
\begin{equation}
\label{nabla_w-cost}
\frac{\partial \xi_i}{\partial w^m_{tj}}=
\frac{\partial \xi_i}{\partial z^{m,i}_j} \sigma'(s^{m,i}_j) z^{m-1,i}_t,\ m = 1, \ldots,  L.
\end{equation}
By considering $\partial \xi_i/\partial z^{m,i}_j$ to be like $\partial z^{L,i}_u/\partial z^{m,i}_j$ in \eqref{jaco-sum}, we
can apply the same backward process so that each partition between layers $m-2$ and $m-1$ must wait for $\partial \xi_i/\partial z^{m-1,i}_j$ from partitions between layers $m-1$ and $m$:
\begin{equation}
\label{nabla_z-cost}
\frac{\partial \xi_i}{\partial z^{m-1,i}_t} =
\sum_{T_m \in P_m} \sum_{j \in T_m} \frac{\partial \xi_i}{\partial z^{m,i}_j} \sigma'(s^{m,i}_j) w^m_{tj},
\end{equation}
where $i=1,\ldots,l$, $t \in T_{m-1}$, and $P_m$ is defined in \eqref{pmdefine}. For the initial $\partial \xi_i/\partial z^{L,i}_j$
in the backward process, from the loss function defined in \eqref{thiswork-loss},
\begin{equation*}
\frac{\partial \xi_i}{\partial z^{L,i}_j} = 2 \times \left( z^{L,i}_j - y^i_j \right).
\end{equation*}
\par From \eqref{distri-grad}, a difference from the Jacobian calculation is that here we obtain a sum over all instances $i$.
Earlier we separately maintain terms related to $T_{m-1}$ and $T_m$ to avoid storing all Jacobian elements. With the summation over $i$,
we can afford to store $\partial f/\partial w^m_{tj}$ and $\partial f/\partial b^m_j$, $\forall t \in T_{m-1},\ \forall j \in T_m$.
\renewcommand{\baselinestretch}{1.3}
\begin{algorithm}[t]
  \caption{Calculation of $\partial z^{L,i}_u/\partial s^{m,i}_j,\ u=1,\ldots,n_L,\ j=1,\ldots,|T_m|$ in a distributed system.}
  \begin{algorithmic}[1]
		\State Let $T_{m-1}$ and $T_m$ be the subsets of neurons at the $(m-1)$th and $m$th layers corresponding to the current partition.
		\If{$m = L$}
			\State Calculate
			\begin{equation*}
			\frac{\partial z^{L,i}_u}{\partial z^{m,i}_j} =
			\begin{cases}
				2(z^{L,i}_u-y^i_u) &\text{if}\ j = u,\\
				0 &\text{if}\ j \neq u,
			\end{cases} 
			,\ u=1,\ldots,n_L,\ i=1,\ldots,l,\ \text{and } j \in T_m.
			\end{equation*}
		\Else
			\State Wait for $\partial z^{L,i}_u/\partial z^{m,i}_j$, $u=1,\ldots,n_L$, $i=1,\ldots,l$, and $j \in T_m$.
		\EndIf
		\State Calculate
		\begin{equation}
    		\label{pzps-alg}
			\frac{\partial z^{L,i}_u}{\partial s^{m,i}_j} = \frac{\partial z^{L,i}_u}{\partial z^{m,i}_j} \sigma' (s_j^{m,i}),\ 
			u=1,\ldots,n_L,\ i=1,\ldots,l,\ \text{and } j \in T_m.
		\end{equation}
		\If{$m > 1$}
			\State Calculate the local sum
        \begin{equation}
          \label{jaco-local-sum-alg}
          \sum_{j \in T_m} \frac{\partial z^{L,i}_u}{\partial s^{m,i}_j} w^m_{tj},\ t \in T_{m-1}
        \end{equation}
        \Statex \indent \indent \hspace{-1.2ex} and do the {\it reduce} operation to obtain
			\begin{equation}
			\label{partialz-alg} 
				\frac{\partial z^{L,i}_u}{\partial z^{m-1,i}_t},\ u=1,\ldots,n_L,\ i=1,\ldots,l, \text{ and } t \in T_{m-1}.
			\end{equation}
			\If{$T_m$ is the first neuron sub-group of layer $m$}
				\State Broadcast values in \eqref{partialz-alg} to partitions between layers $m-2$ and $m-1$ 
                \Statex \indent \indent \hspace{2ex} corresponding to the neuron sub-group $T_{m-1}$ at layer $m-1$; 
                \Statex \indent \indent \hspace{2ex} see the description after \eqref{reduce-sum-jaco}.
			\EndIf
		\EndIf
  \end{algorithmic}
  \label{alg:jaco}
\end{algorithm}
\renewcommand{\baselinestretch}{2}

\section{Techniques to Reduce Computational, Communication, and Synchronization Cost}
\label{sec:Reduce-compu-communi}
In this section we propose some novel techniques to make the distributed Newton method a practical approach for deep neural networks.
\subsection{Diagonal Gauss-Newton Matrix Approximation}
\label{subsec:diagonalize}
\par In \eqref{real-jaco} for the Gauss-Newton matrix-vector products in the CG procedure, we notice that the communication occurs for reducing $P$ vectors
\begin{equation*}
    J^i_1 \bv_1,\ldots,J^i_P \bv_P,
\end{equation*}
each with size $\mathcal{O}(n_L)$, and then broadcasting the sum to all nodes. To avoid the high communication cost in some distributed systems, 
we may consider the diagonal blocks of the Gauss-Newton matrix as its approximation:
\begin{equation}
\label{gausshat}
\hat{G} = \frac{1}{C} \mathcal I + 
\begin{bmatrix}
    \frac{1}{l}\sum_{i=1}^l (J^i_1)^T B^i J^i_1 & & \\
        & \ddots & \\
     & & \frac{1}{l} \sum_{i=1}^l (J^i_P)^T B^i J^i_P
\end{bmatrix}.
\end{equation}
Then \eqref{full-gaussnewton} becomes $P$ independent linear systems
\begin{align}
    (\frac{1}{l}\sum_{i=1}^l (J^i_1)^T B^i J^i_1+\frac{1}{C} \mathcal I + \lambda_k \mathcal I)\bd^k_1 &= -\bg^k_1,\nonumber\\
    \vdots \label{p-linear}\\
    (\frac{1}{l}\sum_{i=1}^l (J^i_P)^T B^i J^i_P+\frac{1}{C} \mathcal I + \lambda_k \mathcal I)\bd^k_P &= -\bg^k_P,\nonumber
\end{align}
where $\bg_1^k, \ldots, \bg_P^k$ are local components of the gradient:
\begin{equation*}
\nabla f(\btheta^k) =
\begin{bmatrix}
\bg_1^k \\ \vdots \\ \bg_P^k
\end{bmatrix}.
\end{equation*}
The matrix-vector product becomes
\begin{equation}
G\bv \approx \hat{G}\bv = 
\begin{bmatrix} \frac{1}{l}\sum_{i=1}^l (J^i_{1})^T B^i J^i_{1}\bv_1 + \frac{1}{C}\bv_1\\ \vdots \\ \frac{1}{l}\sum_{i=1}^l (J^i_{P})^T  B^{i} J^i_{P}\bv_P + \frac{1}{C}\bv_P 
\end{bmatrix}, 
\label{approx-jaco}
\end{equation}
in which each $(G\bv)_p$ can be calculated using only local information because we have independent linear systems. 
For the CG procedure at any partition, it is terminated if the following relative stopping condition holds
\begin{equation}
    \label{CG-stopping-cond}
    ||\frac{1}{l}\sum_{i=1}^l (J^i_p)^T B^i J^i_p \bv_p + (\frac{1}{C} + \lambda_k )\bv_p + \bg_p^k|| \leq \sigma ||\bg_p^k||
\end{equation}
or the number of CG iterations reaches a pre-specified limit. Here $\sigma$ is a pre-specified tolerance. Unfortunately, partitions may finish their CG procedures at different time,
a situation that results in significant waiting time. To address this synchronization cost, we propose some novel techniques in Section \ref{subsec:Synchronization}.
\par Some past works have considered using diagonal blocks as the approximation of the Hessian. For logistic regression, \cite{YB13a} consider diagonal elements of the Hessian
to solve several one-variable sub-problems in parallel. \cite{DM14a} study a more general setting in which using diagonal blocks is a special case.

\subsection{Product Between Gauss-Newton Matrix and a Vector}
\label{subsec:Jacobian-Vector}
In the CG procedure the main computational task is the matrix-vector product. We present techniques for the efficient calculation.
From \eqref{approx-jaco}, for the $p$th partition, the product between the local diagonal block of the Gauss-Newton matrix and a vector $\bv_p$ takes the following form.
\begin{equation*}
(J_p^i)^T B^i J_p^i \bv_p.
\end{equation*} 
Assume the $p$th partition involves neuron sub-groups $T_{m-1}$ and $T_m$ respectively in layers $m-1$ and $m$, and this partition is not responsible to handle the bias term $\bb^m_j,\ \forall j \in T_m$.
Then
\begin{equation*}
J^i_p \in \mathcal R^{n_L \times (|T_{m-1}| \times |T_{m}|)} \text{ and } \bv_p \in \mathcal R^{(|T_{m-1}| \times |T_{m}|) \times 1}.
\end{equation*}
Let $\text{mat}(\bv_p) \in \mathcal R^{|T_{m-1}|\times|T_m|}$ be the matrix representation of $\bv_p$. 
From \eqref{Jelement-split}, the $u$th component of $(J^i_p\bv_p)_u$ is
\begin{equation}
\label{direct-JV-cost}
\sum_{t \in T_{m-1}} \sum_{j \in T_m} \frac{\partial z^{L,i}_u}{\partial w^{m}_{tj}} (\text{mat}(\bv_p))_{tj}
= \sum_{t \in T_{m-1}} \sum_{j \in T_m} \frac{\partial z^{L,i}_u}{\partial s^{m,i}_j} z^{m-1,i}_t (\text{mat}(\bv_p))_{tj}.
\end{equation}
A direct calculation of the above value requires $\mathcal{O}(|T_{m-1}|\times|T_m|)$ operations. 
Thus to get all $u=1,\ldots,n_L$ components, the total computational cost is proportional to
\begin{equation*}
n_{L} \times |T_{m-1}| \times |T_{m}|.
\end{equation*}
We discuss a technique to reduce the cost by rewriting \eqref{direct-JV-cost} as
\begin{equation*}
 \sum_{j \in T_m} \frac{\partial z^{L,i}_u}{\partial s^{m,i}_j}\left(\sum_{t \in T_{m-1}} z^{m-1,i}_t (\text{mat}(\bv_p))_{tj}\right).
\end{equation*}
While calculating
\begin{equation*}
\sum_{t \in T_{m-1}} z^{m-1,i}_t (v_p)_{tj},\ \forall j \in T_m
\end{equation*}
still needs $\mathcal{O}(|T_{m-1}|\times|T_m|)$ cost, we notice that these values are independent of $u$. That is, they
can be stored and reused in calculating $(J^i_p\bv_p)_u,\ \forall u$. Therefore, the total computational cost is significantly reduced to
\begin{equation}
\label{jv-comp-cost}
|T_{m-1}| \times |T_{m}| + n_L \times |T_m|.
\end{equation}

\par The procedure of deriving $(J^i_p)^T (B^i J_p^i \bv_p)$ is similar. Assume 
\begin{equation*}
\bar{\bv} = B^i J_p^i \bv_p \in \mathcal R^{n_L \times 1}. 
\end{equation*}
From \eqref{Jelement-split},
\begin{align}
\text{mat}\left((J^i_p)^T \bar{\bv}\right)_{tj} &= \sum_{u=1}^{n_L} \frac{\partial z^{L,i}_u}{\partial w^m_{tj}} \bar{v}_{u} \nonumber\\
							 &= \sum_{u=1}^{n_L} \frac{\partial z^{L,i}_u}{\partial s^{m,i}_{j}} z^{m-1,i}_t \bar{v}_{u} \nonumber\\ 
							 &= z^{m-1,i}_t \left( \sum_{u=1}^{n_L} \frac{\partial z^{L,i}_u}{\partial s^{m,i}_{j}} \bar{v}_{u}\right). \label{JTV-independent} 
\end{align}
Because 
\begin{equation}
\label{JTv}
\sum_{u=1}^{n_L} \frac{\partial z^{L,i}_u}{\partial s^{m,i}_{j}} \bar{v}_{u},\ \forall j \in T_m
\end{equation}
are independent of $t$, we can calculate and store them for the computation in \eqref{JTV-independent}.
Therefore, the total computational cost is proportional to
\begin{equation}
\label{JTv-comp-cost}
|T_{m-1}| \times |T_{m}| + n_L \times |T_m|,
\end{equation}
which is the same as that for $(J^i_p \bv_p)$.
\par In the above discussion, we assume that diagonal blocks of the Gauss-Newton matrix are used. If instead the whole Gauss-Newton matrix
is considered, then we calculate
\begin{equation*}
(J^i_{p_1})^T(B^i(J^i_{p_2} \bv_{p_2})),
\end{equation*}
for any two partitions $p_1$ and $p_2$. The same techniques introduced 
in this section can be applied because \eqref{direct-JV-cost} and \eqref{JTV-independent} are two independent operations.

\subsection{Subsampled Hessian Newton Method}
\label{subsec:Subsampled}
From \eqref{def-gauss-newton} we see that the computational cost 
between the Gauss-Newton matrix and a vector is 
proportional to the number of data. To reduce the cost, 
subsampled Hessian Newton method \citep{RHB11a,JM10a,CCW15a} 
have been proposed for selecting a subset of data at each iteration to form an approximate Hessian. Instead of $\nabla^2 f(\btheta)$ in \eqref{full-newton-noLM} we use a subset $S$ to have
\begin{equation*}
	\frac{\mathcal I}{C} + \frac{1}{|S|}\sum_{i \in S} \nabla^2_{\btheta\btheta} \xi(\bz^{L,i};\by^i).
\end{equation*}
Note that $\bz^{L,i}$ is a function of $\btheta$. The idea behind this subsampled Hessian is that when a large set of points are under the same distribution,
\begin{equation*}
	\frac{1}{|S|}\sum_{i\in S} \xi(\bz^{L,i};\by^i).
\end{equation*}
is a good approximation of the average training losses. For neural networks we consider the Gauss-Newton matrix, so \eqref{def-gauss-newton} becomes the following subsampled Gauss-Newton matrix.
\begin{equation}
\label{sampled-Gauss}
	G^S = \frac{\mathcal I}{C} + \frac{1}{|S|} \sum_{i \in S} (J^i)^T B^i J^i.
\end{equation}
Now denote the subset at the $k$th iteration as $S_k$. The linear system \eqref{full-gaussnewton} is changed to
\begin{equation}
\label{sample-gaussnewton}
(G^{S_k}  + \lambda_k \mathcal I) \bd^k = - \nabla f(\btheta^k).
\end{equation}
After variable partitions, the independent linear systems are
\begin{align}
  \left( \lambda_k\mathcal I + \frac{1}{C} \mathcal I + \frac{1}{|S_k|} \sum_{i \in S_k} (J^i_1)^T B^i J^i_1 \right) \bd^k_1 &= -\bg_1^k, \nonumber\\
	\vdots  \label{block-gaussnewton}\\
  \left( \lambda_k\mathcal I + \frac{1}{C} \mathcal I + \frac{1}{|S_k|} \sum_{i \in S_k} (J^i_P)^T B^i J^i_P \right) \bd^k_P &= -\bg_P^k. \nonumber
\end{align}
\par While using diagonal blocks of the Gauss-Newton matrix avoids the communication between partitions, the resulting direction may not be as good as
that of using the whole Gauss-Newton matrix. Here we extend an approach by \cite{CCW15a} to pay some extra cost for improving the direction.
Their idea is that after the CG procedure of using a sub-sampled Hessian, they consider the full Hessian to adjust the direction. Now in the CG procedure we use a block diagonal approximation of the sub-sampled matrix $G^{S_k}$, so after that we consider the whole $G^{S_k}$ for adjusting the direction. Specifically, if $\bd^k$ is obtained from the CG procedure, we solve the following two-variable optimization problem that involves $G^{S_k}$.
\begin{equation}
\label{linear-comb-now}
\min_{\beta_1,\beta_2}\; \frac{1}{2} (\beta_1 \bd^k + \beta_2 \bar{\bd}^k)^T G^{S_k} (\beta_1 \bd^k + \beta_2 \bar{\bd}^k) + \nabla f(\btheta^k)^T (\beta_1 \bd^k + \beta_2 \bar{\bd}^k),
\end{equation}
where $\bar{\bd}^k$ is a chosen vector. Then the new direction is
\begin{equation*}
\bd^k \leftarrow \beta_1 \bd^k + \beta_2 \bar{\bd}^k.
\end{equation*}
Here we follow \cite{CCW15a} to choose
\begin{equation*}
\bar{\bd}^k = \bd^{k-1}.
\end{equation*}
Notice that we choose $\bar{\bd}^0$ to be the zero vector. A possible advantage of considering $\bd^{k-1}$ is that it is from the previous iteration of using
a different data subset $S_{k-1}$ for the subsampled Gauss-Newton matrix. Thus it provides information from instances not in the current $S_k$.
\par To solve \eqref{linear-comb-now}, because $G^{S_k}$ is positive definite, it is equivalent to solving the following two-variable linear system.
\begin{equation}
\label{sol-comp}
\left(
\begin{array}{cccc}
(\bd^k)^T G^{S_k} \bd^k & (\bar{\bd}^k)^T G^{S_k} \bd^k\\
(\bar{\bd}^k)^T G^{S_k} \bd^k & (\bar{\bd}^k)^T G^{S_k} \bar{\bd}^k
\end{array} \right) \left(\begin{array}{cccc}\beta_1 \\ \beta_2 \end{array} \right)
= \left(\begin{array}{cccc} -\nabla f(\btheta^k)^T \bd^k \\ -\nabla f(\btheta^k)^T \bar{\bd}^k \end{array} \right).
\end{equation}
Note that the construction of \eqref{sol-comp} involves 
the communication between partitions; see detailed
discussion in Section V of 
supplementary materials. The effectiveness of using
\eqref{linear-comb-now} is investigated in Section
VII.

\par In some situations, the linear system \eqref{sol-comp} may be ill-conditioned. We set $\beta_1 = 1$ and $\beta_2 = 0$ if
\begin{equation}
\label{determinant}
	\begin{vmatrix} (\bd^k)^T G^{S_k} \bd^k & (\bar{\bd}^k)^T G^{S_k}\bd^k \\ (\bar{\bd}^k)^T G^{S_k} \bd^k & (\bar{\bd}^k)^T G^{S_k}\bar{\bd}^k \end{vmatrix} \leq \varepsilon,
\end{equation}
where $\varepsilon$ is a small number.

\subsection{Synchronization Between Partitions}
\label{subsec:Synchronization}
While the setting in \eqref{approx-jaco} has made each node conduct its own CG procedure without communication, we must
wait until all nodes complete their tasks before getting into the next Newton iteration.
This synchronization cost can be significant. We note that the running time at each partition may vary because of the following reasons.
\begin{enumerate}[1.]
\item
Because we select a subset of weights between two layers as a partition, the number of variables in each partition may be different. For
example, assume the network structure is 
\begin{equation*}
	50\text{-}100\text{-}2.
\end{equation*}
The last layer has only two neurons because of the small number of classes. 
For the weight matrix $W^m$, a partition between the last two layers can have at most $200$ variables. 
In contrast, a partition between the first two layers may have more variables. 
Therefore, in the split of variables we should make partitions as balanced as possible.
A example will be given later when we introduce the experiment settings in Section \ref{subsec:Analysis-newton}.

\item 
Each node can start its first CG iteration after the needed information is available. 
From \eqref{ztox-jaco}-\eqref{xtow-jaco-bias}, the calculation of the information needed for matrix-vector products 
involves a backward process, so partitions corresponding to neurons in the last layers start the CG procedure earlier than those of the first layers.
\end{enumerate}

\par To reduce the synchronization cost, a possible solution is to terminate the CG procedure for all partitions if one of them reaches its CG stopping condition:
\begin{equation}
\label{cg-stopping-sample}
||(\lambda_k + \frac{1}{C}) \bv_p + \frac{1}{|S_k|}\sum_{i \in S_k} (J^i_p)^T B^i J^i_p\bv_p + \bg_p|| \leq \sigma ||\bg_p||.
\end{equation}
However, under this setting the CG procedure may terminate too early because some partitions have not conducted enough CG steps yet. To strike for a balance, in our implementation we terminate the CG procedure for all partitions when the following conditions are satisfied:
\begin{enumerate}[1.]
	\item Every partition has reached a pre-specified minimum number of CG iterations, $\text{CG}_{\text{min}}$.
	\item A certain percentage of partitions have reached their stopping conditions, \eqref{cg-stopping-sample}.
\end{enumerate}
In Section \ref{subsec:Analysis-newton}, we conduct experiments with different percentage values to check the effectiveness of this setting.

\subsection{Summary of the Procedure}
\label{subsec:summary-procedure}
We summarize in Algorithm \ref{alg:distri} the proposed distributed subsampled Hessian Newton algorithm. Besides materials described earlier in this section, here we explain other steps in the algorithm.

First, in most optimization algorithms, after a direction $\bd^k$ is obtained, a suitable step size $\alpha_k$ must be decided to ensure the sufficient decrease of $f(\btheta^k + \alpha_k \bd^k)$. Here we consider a backtracking line search by selecting the largest $\alpha_k \in \{1, \frac{1}{2}, \frac{1}{4}, \ldots\}$ such that the following sufficient decrease condition on the function value holds.
\begin{equation}
\label{linesearch-condition}
f(\btheta^{k} + \alpha_k \bd^k) \leq f(\btheta^k) + \eta \alpha_k \nabla f(\btheta^k)^T \bd^k,
\end{equation}
where $\eta \in (0,1)$ is a pre-defined constant.

Secondly, we follow \cite{JM10a,JM12a,CCW15a} to apply the Levenberg-Marquardt method by introducing a term $\lambda_k \mathcal{I}$
in the linear system \eqref{full-gaussnewton}. Define
\begin{equation*}
  \rho_k = \frac{f(\btheta^k + \alpha_k \bd^k) - f(\btheta^k)}{\alpha_k \nabla f(\btheta^k)^T \bd^k + \frac{1}{2} (\alpha_k)^2 (\bd^k)^T G^{S_k} \bd_k}
\end{equation*} 
as the ratio between the actual function reduction and the predicted reduction. Based on $\rho_k$, the following rule
derives the next $\lambda_{k+1}$. 
\begin{equation}
\label{LM-rules}
\lambda_{k+1} =
\begin{cases}
\lambda_k \times \text{drop}& \rho_k > 0.75, \\
\lambda_k & 0.25 \leq \rho_k \leq 0.75, \\
\lambda_k \times \text{boost}& \text{otherwise,}
\end{cases}
\end{equation}
where (drop,boost) are given constants. 
Therefore, if the predicted reduction is close to the true function reduction, we reduce $\lambda_k$ such that
a direction closer to the Newton direction is considered. In contrast, if $\rho_k$ is small, we enlarge $\lambda_k$
so that a conservative direction close to the negative gradient is considered.
\par Note that line search already serves as a way to adjust the direction according to the function-value
reduction, so in optimization literature line search and Levenberg-Marquardt method are seldom applied concurrently.
Interestingly, in recent studies of Newton methods for neural networks, both techniques are considered. Our preliminary investigation in Section VI of supplementary materials shows that using Levenberg-Marquardt method together with line search is very helpful, but more detailed studies can be a future research issue.
\par In Algorithm \ref{alg:distri} we show a master-master implementation, so
the same program is used at each partition. Some
careful designs are needed to ensure that all partitions
get consistent information. For example, we can use the
same random seed to ensure that at each iteration all 
partitions select the same set $S_k$ in constructing the subsampled Gauss-Newton matrix.
\renewcommand{\baselinestretch}{1.3}
\begin{algorithm}[t]
  \caption{A distributed subsampled Hessian Newton method with variable partition.}
  \begin{algorithmic}[1]
  \State Given $\epsilon \in (0,1)$, $\lambda_{1}$, $\sigma \in (0,1)$, $\eta \in (0,1)$, $\text{CG}_{\max}$, $\text{CG}_{\min}$, and $r \in (0,100]$.
  \State Let $p$ be the index of the current partition and generate the initial local model vector $\btheta^1_p$.
  \State Compute $f(\btheta^{1})$.
  \For{$k = 1,\ldots,$}
    \State Choose a set $S_k \subset \{1,\ldots,l\}$.
  	\State Compute $\bg_p^k$ and $J^i_p, \forall i \in S_k$.
	\State Approximately solve the linear system in \eqref{block-gaussnewton} by CG to obtain a direction $\bd_p^k$
	\Statex \hspace{3ex} after 
	\begin{equation*}
        || (\lambda_k \mathcal I + \frac{1}{C} \mathcal I + \frac{1}{|S_k|}\sum_{i=1}^{|S_k|}(J^i_p)^TB^iJ^i_p)\bd_p^k + \bg_p^k|| \leq \sigma ||\bg_p^k||
	\end{equation*}
  \Statex \indent \hspace{3ex} is satisfied or $\# \text{CG}_p^k \geq \text{CG}_{\max}$ or 
	\begin{equation*}
	 \text{\{\# partitions finished} \geq r\% \times P \text{ and }  \#\text{CG}^k_p \geq \text{CG}_{\min}\}, 
	\end{equation*}
	\Statex \indent \hspace{3ex} where $\#\text{CG}_p^k$ is the number of CG iterations that have been run so far.
	\State Derive $\bd^k_p = \beta_1\bd^{k}_p + \beta_2 \bd^{k-1}_p$ by solving \eqref{linear-comb-now}.
	\State $\alpha^k = 1$.
	\While{true}
		\State Update $\btheta^{k+1}_p = \btheta^{k}_p + \alpha^k \bd^{k}_p$ and then compute $f(\btheta^{k+1})$.
		\If{$T_m$ and $T_{m-1}$ are the first neuron subgroups at layers $L$ and $L-1$, respectively,}
			\If{\eqref{linesearch-condition} is satisfied}
				\State Notify all partitions to stop. 
			\EndIf
		\Else
			\State Wait for the notification to stop.
		\EndIf
		\If{the stop notification has been received}
			\State break;
		\EndIf
		\State $\alpha^k = \alpha^k/2$.
	\EndWhile
	\State Update $\lambda_{k+1}$ based on \eqref{LM-rules}.
  \EndFor
  \end{algorithmic}
  \label{alg:distri}
\end{algorithm}
\renewcommand{\baselinestretch}{2}

\section{Analysis of the Proposed Algorithm}
\label{sec:Analyze-algo}
In this section, we analyze Algorithm \ref{alg:distri} on the memory requirement, the computational cost, and the communication cost. 
We assume that the full training set is used. If the 
subsampled Hessian method in Section \ref{subsec:Subsampled} is applied, 
then in the Jacobian calculation and the Gauss-Newton matrix vector product the ``$l$'' term in our analysis should be replaced by 
the subset size $|S|$.
\subsection{Memory Requirement at Each Partition}
\label{subsec:Memory}
Assume the partition corresponds to the neuron sub-groups $T_{m-1}$ at layer $m-1$ and $T_m$ at layer $m$.
We then separately consider the following situations.
\begin{enumerate}[1.]
\item Local weight matrix:
    Each partition must store the local weight matrix.
    \begin{equation*}
        w^m_{tj},\ \forall t \in T_{m-1}, \text{and } \forall j \in T_m.
    \end{equation*}
    If $T_{m-1}$ is the first neuron sub-group of layer $m-1$, it also needs to store
    \begin{equation*}
        b^m_{j},\ \forall j \in T_m.
    \end{equation*}
    Therefore, the memory usage at each partition for the local weight matrix is proportional to
    \begin{equation*}
        |T_{m-1}| \times |T_m| + |T_m|.
    \end{equation*}
\item Function evaluation: 
	From Section \ref{subsec:Distributed-Function}, we must store part of $\bz^{m-1,i}$ 
    and $\bz^{m,i}$ vectors.\footnote{Note that the same vector is used to store the $\bs$ vector before it is transformed to $\bz$ by the activation function.} The memory usage at each partition is
    \begin{equation}
	\label{memory-fun}
        l \times ( |T_{m-1}| + |T_m|).
    \end{equation}
\item Gradient evaluation:
    First, we must store 
    \begin{equation*}
			\frac{\partial f}{\partial w^m_{tj}} \text{ and } \frac{\partial f}{\partial b^m_j}, t \in T_{m-1}, j \in T_m
	\end{equation*}
	after the gradient evaluation.
    Second, for the backward process, from \eqref{nabla_z-cost}, we must store
	\begin{equation*}
	\frac{\partial \xi_i}{\partial z^{m-1,i}_t},\ \forall t \in T_{m-1},\ \forall i\ \text{ and }\ \frac{\partial \xi_i}{\partial z^{m,i}_j},\ \forall j \in T_m,\ \forall i. 
	\end{equation*}
    Therefore, the memory usage in each partition is proportional to
    \begin{equation}
	\label{memory-grad}
        (|T_{m-1}| \times |T_m| + |T_m|) + l \times ( |T_{m-1}| + |T_m|).
    \end{equation}
\item Jacobian evaluation:
    From the discussion in Section \ref{subsubsec:Jmemory}, the memory consumption is proportional to
    \begin{equation}
        \label{Jmemory}
        l \times n_L \times ( |T_{m-1}| + |T_m|).
    \end{equation}
\end{enumerate}
In summary, the memory bottleneck is on terms that are related to the number of instances.  
To reduce the memory use, we have considered a technique in Section \ref{subsec:Subsampled} to replace the term $l$ in \eqref{Jmemory} with a smaller subset size $|S^k|$. We will further discuss a technique to reduce the memory consumption in Section \ref{subsec:pipeline}.
\subsection{Computational Cost}
\label{subsec:Computation-cost}
We analyze the computational cost at each partition.
For the sake of simplicity, we make the following assumptions.
\begin{itemize}
\item At the $m$th layer neurons are evenly split to several sub-groups, each of which has $|T_m|$ elements.
\item Calculating the activation function $\sigma(s)$ needs $1$ operation.
\end{itemize}
The following analysis is for a partition between layers $m-1$ and $m$.
\begin{enumerate}[1.]
\item Function evaluation: From Algorithm \ref{alg:eval}, after $s^{m-1,i}_t,\ i=1,\ldots,l,\ t \in T_{m-1}$ are available, we must calculate \eqref{z-alg} and \eqref{local-weight-sum}. 
The dominant one is \eqref{local-weight-sum}, so the computational cost of function evaluation is
\begin{equation}
\label{f-compu-cost}
\mathcal {O} (l \times |T_m| \times |T_{m-1}|). 
\end{equation}
\item Gradient evaluation: 
Assume that the current partition has received $\partial \xi_i/\partial z^{m,i}_j,\ i=1,\ldots,l,\ j \in T_m$.
From \eqref{nabla_w-cost}, we calculate
\begin{align*}
\frac{\partial f}{\partial w^m_{tj}} 
&= \frac{1}{C}w^m_{tj} + \frac{1}{l} \sum_{i=1}^l \frac{\partial \xi_i}{\partial w^m_{tj}} \\
&= \frac{1}{C}w^m_{tj} + \frac{1}{l} \sum_{i=1}^l \frac{\partial \xi_i}{\partial z^{m,i}_j} \sigma'(s^{m,i}_j) z^{m-1,i}_t,\ \forall t \in T_{m-1},\ \forall j \in T_{m}, 
\end{align*}
which costs
\begin{equation*}
\mathcal {O} ( l \times |T_m| \times |T_{m-1}|). 
\end{equation*}
Then for the {\it reduce} operation in \eqref{nabla_z-cost}, calculating the local sum
\begin{equation*}
\sum_{j \in T_m} \frac{\partial \xi_i}{\partial z^{m,i}_j} \sigma'(s^{m,i}_j) w^m_{tj} ,\ i=1,\ldots,l,\ t \in T_{m-1}.
\end{equation*}
has a similar cost. Thus the computational cost of gradient evaluation is
\begin{equation}
\label{g-compu-cost}
\mathcal {O} ( l \times |T_m| \times |T_{m-1}|). 
\end{equation}
\item Jacobian evaluation:
From \eqref{pzps-alg} and \eqref{jaco-local-sum-alg} in Algorithm \ref{alg:jaco},
the computational cost is
\begin{equation}
\label{J-compu-cost}
\mathcal {O} (n_L \times l \times |T_m| \times |T_{m-1}|). 
\end{equation}
\item Gauss-Newton matrix-vector products:
Following \eqref{JTv-comp-cost} in Section \ref{subsec:Jacobian-Vector}, ,
the computational cost for Gauss-Newton matrix vector products is
\begin{equation}
\label{GV-compu-cost}
\#\, \mbox{CG iterations} \times \left( l \times \left(|T_{m-1}| \times |T_{m}| + n_L \times |T_m|\right) \right).
\end{equation}
\end{enumerate}
\par From \eqref{f-compu-cost}-\eqref{GV-compu-cost}, we can derive the following conclusions.
\begin{enumerate}[1.]
\item The computational cost is proportional to the number of training data, the number of classes, and the number of variables in a partition.
\item In general, \eqref{J-compu-cost} and \eqref{GV-compu-cost} dominate the computational cost. Especially, when the number of CG iterations is large,
\eqref{GV-compu-cost} becomes the bottleneck. 
\item If the subsampling techniques in Section \ref{subsec:Subsampled} is used, 
then $l$ in \eqref{J-compu-cost}-\eqref{GV-compu-cost} is replaced with the size of the subset. 
Therefore, the computational cost at each partition in a Newton iteration can be effectively reduced. However, the number of iterations may be increased.
\item The computational cost can be reduced by splitting neurons at each layer to as many sub-groups as possible. 
However, because each partition corresponds to a computing node, more partitions imply a higher synchronization cost. 
Further, the total number of neurons at each layer is different, so the size of each partition may significant vary, a situation that further worsens the synchronization issue.
\end{enumerate}

\subsection{Communication Cost}
\label{subsec:Communi-cost}
We have shown in Section \ref{subsec:Variable} that by using diagonal blocks of the Gauss-Newton matrix, each partition conducts a CG procedure without
communicating with others. However, communication cost still occurs for function, gradient, and Jacobian evaluation. We discuss details for the Jacobian
evaluation because the situation for others is similar.

To simplify the discussion we make the following assumptions.
\begin{enumerate}[1.]
\item At the $m$th layer neurons are evenly split to several sub-groups, each of which has $|T_m|$ elements.
Thus the number of neuron sub-groups at layer $m$ is $n_m/|T_m|$. 
\item Each partition sends or receives one message at a time.
\item Following \cite{MB94a}, the time to send or receive a vector $\bv$ is
\begin{equation*}
\alpha + \beta \times |\bv|,
\end{equation*}
where $|\bv|$ is the length of $\bv$, $\alpha$ is the start-up cost of a transfer and $\beta$ is the transfer rate of the network.
\item The time to add a vector $\bv$ and another vector of the same size is
\begin{equation*}
\gamma \times |\bv|.
\end{equation*}
\item
Operations (including communications) of independent groups of nodes can be conducted in parallel.
For example, the two trees in Figure IV.3 of supplementary materials involve two independent sets of partitions. 
We assume that the two {\it reduce} operations can be conducted in parallel.
\end{enumerate}
From \eqref{jaco-sum}, for partitions between layers $m-1$ and $m$ that correspond to the same neuron sub-group $T_{m-1}$
at layer $m-1$, the {\it reduce} operation on $\partial z^{L,i}_u/\partial z^{m-1,i}_t,\ u=1,\ldots,n_L,\ t \in T_{m-1},\ i=1,\ldots,l$ sums up
\begin{equation*}
\frac{n_m}{|T_m|} \text{ vectors of } l \times n_L \times |T_{m-1}| \text{ size.}
\end{equation*} 
For example, the layer $2$ in Figure \ref{fig:neuron} is split to three groups $A_2$, $B_2$ and $C_2$, so for the sub-group $A_1$ in layer $1$, three vectors from ($A_1$, $A_2$), ($A_1$, $B_2$) and ($A_1$, $C_2$) are reduced. 
Following the analysis in \cite{JP07b}, the communication cost for the {\it reduce} operation is
\begin{equation}
\label{comm_red}
\mathcal {O}(\lceil (\log_2 ( \frac{n_m}{|T_m|}) \rceil \times \left(\alpha + (\beta + \gamma)\times( l \times n_L \times |T_{m-1}|)\right).
\end{equation}
Note that between layers $m-1$ and $m$
\begin{equation*}
    \frac{n_{m-1}}{|T_{m-1}|}\ \text{{\it reduce} operations}
\end{equation*}
are conducted and each takes the communication cost shown in \eqref{comm_red}. However, by our assumption they can be fully parallelized.

The reduced vector of size $l \times n_L \times |T_{m-1}|$ is then broadcasted to $n_{m-2}/|T_{m-2}|$ partitions. 
Similar to \eqref{comm_red}, the communication cost is
\begin{equation}
  \label{comm_bcast}
  \mathcal {O}(\lceil (\log_2 ( \frac{n_{m-2}}{|T_{m-2}|}) \rceil \times ( \alpha + \beta \times ( l \times n_L \times |T_{m-1}|) ) ).
\end{equation}
The $\gamma$ factor in \eqref{comm_red} does not appear here because we do not need to sum up vectors. 
\par Therefore, the total communication cost of the Jacobian evaluation is the sum of 
\eqref{comm_red} and \eqref{comm_bcast}. We can make the following conclusions.
\begin{enumerate}
    \item[1.] The communication cost is proportional to the number of training instances as well as the number of classes.
    \item[2.] From \eqref{comm_red} and \eqref{comm_bcast}, a smaller $|T_{m-1}|$ reduces the communication cost. However, we can not split neurons at each layer to too many groups because of the following reasons. First, we assumed earlier that for independent sets of partitions, their operations including communication within each set can be fully parallelized. In practice, the more independent sets the higher synchronization cost. Second, when there are too many partitions the block diagonal matrix in \eqref{gausshat} may not be a good approximation of the Gauss-Newton matrix.
\end{enumerate}

\section{Other Implementation Techniques}
\label{sec:other-implement}
In this section, we discuss additional techniques implemented in the proposed algorithm.
\subsection{Pipeline Techniques for Function and Gradient Evaluation}
\label{subsec:pipeline}
The discussion in Section \ref{sec:Analyze-algo} indicates that in our proposed method the memory requirement, the computational cost and the communication cost all linearly
increase with the number of data. For the product between the Gauss-Newton matrix and a vector, we have considered using subsampled Gauss-Newton matrices in Section \ref{subsec:Subsampled}
to effectively reduce the cost. To avoid that function and gradient evaluations become the bottleneck, here we discuss a pipeline technique.
\par The idea follows from the fact that in \eqref{obj-function} 
\begin{equation*}
    \xi_i, \forall i
\end{equation*}
are independent from each other. The situation is the same for 
\begin{equation*}
(J^i)^T\nabla_{\bz^{L,i}}\xi(\bz^{L,i};\by^i), \forall i
\end{equation*}
in \eqref{whole-gradient}. Therefore, in the forward (or the backward) process, once results related to an instance $\bx^i$ are ready,
they can be passed immediately to partitions in the next (or previous) layers. Here we consider a mini-batch implementation. Take the function evaluation as an example.
Assume $\{1,\ldots,l\}$ is split to $R$ equal-sized subsets $S_1,\ldots,S_R$.
At a variable partition between layers $m-1$ and $m$, we showed earlier that local values in \eqref{local-weight-sum} are obtained for all instances $i=1,\ldots,l$. 
Now instead we calculate
\begin{equation*}
\sum_{t \in T_{m-1}} w_{tj}^m z^{m-1,i}_t + b^m_j,\ j \in T_m,\ i \in S_r.
\end{equation*}
The values are used to calculate
\begin{equation*}
s^{m,i}_j,\ \forall i \in S_r.
\end{equation*}
\par By this setting we achieve better parallelism. Further, because we split $\{1,\ldots,l\}$ to subsets with the same size, the memory space allocated
for a subset can be reused by another. Therefore, the memory usage is reduced by $R$ folds.

\subsection{Sparse Initialization}
\label{subsec:Sparse-Init}
A well-known problem in training neural networks is the easy overfitting because of an enormous number of weights. 
Following the approach in Section $5$ of \cite{JM10a}, we implement the sparse initialization for the weights to train deep neural networks. 
For each neuron in the $m$th layer, 
among the $n_{m-1}$ weights connected to it, we randomly assign several weights to have values from the $\mathcal{N}(0,1)$ distribution. Other weights are kept zero.
\par We will examine the effectiveness of this initialization in Sections \ref{subsec:multiclass} and \ref{subsec:compare-sgd}.
\section{Existing Optimization Methods for Training Neural Networks}
\label{sec:Other-optimization}
Besides Newton methods considered in this work, many other optimization methods have been applied to train neural networks. We briefly discuss
the most commonly used one in this section.
\subsection{Stochastic Gradient Methods}
\label{subsec:SGD}
For deep neural networks, it is time-consuming to calculate the gradient vector because from \eqref{whole-gradient}, 
we must go through the whole training data set.
Instead of using all data instances, stochastic gradient (SG) methods randomly choose an example $(\by^{i_k},\bx^{i_k})$ to derive the following sub-gradient vector to
update the weight matrix.
\begin{equation*}
    \nabla f^{i_k}(\btheta^k) =  \frac{\btheta^k}{C} + (J^{i_k})^T \nabla_{\bz^{L,i_k}} \xi(\bz^{L,i_k};\by^{i_k}).
\end{equation*}
Algorithm \ref{alg:sgd} gives the standard setting of SG methods. 
\renewcommand{\baselinestretch}{1.3}
\begin{algorithm}[t]
  \caption{Standard stochastic gradient methods}
  \begin{algorithmic}[1]
  \State Given a learning rate $\eta$.
  \For{$k = 0,\ldots$}
  	\State Choose $i_k \in \{1,\ldots,l\}$.
	\State $\btheta^{k+1} = \btheta^k - \eta \nabla f^{i_k}(\btheta^k)$.
  \EndFor
  \end{algorithmic}
  \label{alg:sgd}
\end{algorithm}
\renewcommand{\baselinestretch}{2}
\par Assume that one epoch means the SG procedure goes through the whole training data set once.
Based on the frequent updates of the weight matrix, SG methods can get a reasonable solution in a few epochs.
Another advantage of SG methods is that Algorithm \ref{alg:sgd} is easy to implement. 
However, if the variance of the gradient vector for each instance is large, SG methods may have slow convergence. 
To address this issue, mini-batch SG method have been proposed to accelerate the convergence speed \citep[e.g.,][]{LB91a,JD12a,JN11a,PB14a}. 
Assume $S_k \subset \{1,\ldots,l\}$ is a subset of the training data. The sub-gradient vector can be as follows:
\begin{equation*}
    \nabla f^{S_k}(\btheta^k) =  \frac{\btheta^k}{C} + \frac{1}{|S_k|} \sum_{i \in S_k} (J^{i})^T \nabla_{\bz^{L,i}} \xi(\bz^{L,i};\by^{i}).
\end{equation*}
However, when SG methods meet ravines which cause the particular dimension apparent to other dimensions, they are easier to drop to local optima. 
\cite{BTP64} proposes using the previous direction with momentum as part of the current direction. This setting may decrease the impact of a particular dimension.
Algorithm \ref{alg:sgd-batch} gives details of a mini-batch SG method with momentum implemented in {\sl Theano/Pylearn2} \citep{IJG13a}.
\par Many other variants of SG methods have been proposed, but it has been shown \citep[e.g.,][]{IS13a} that the mini-batch SG with momentum is a strong baseline.
Thus in this work we do not include other types of SG algorithms for comparison.
\renewcommand{\baselinestretch}{1.3}
\begin{algorithm}[t]
  \caption{Mini-batch stochastic gradient methods in {\sl Theano/Pylearn2} \citep{IJG13a}.}
  \begin{algorithmic}[1]
    \State Given epoch $ = 0$, min\_epochs $ = 200$, a learning rate $\eta$, a minimum learning rate $\eta_{\min}=10^{-6}$, $\alpha=0$, $r=0$, $X=10^{-5}$, $N=10$, a batch size $b = |S_k| = 100$, an initial momentum $m_0=0.9$, a
  final momentum $m_f=0.99$, an exponentially decay factor $\gamma=1.0000002$, and an updating vector $\bv \leftarrow \bzero$.
  \State counter $\leftarrow N$.
  \State lowest\_value $\leftarrow \infty$.
  \While{epoch $<$ min\_epochs or counter $> 0$}
  \State Split the whole training data into $K$ disjoint subsets, $S_k,\ k=1,\ldots,K$.
  \State $\alpha \leftarrow \min(\text{epoch}/\text{min\_epochs}, 1.0)$.
  \State $m \leftarrow (1-\alpha)m_0 + \alpha m_f$.
  \For{$k=1,\ldots,K$}
  \State $\bv \leftarrow m\bv - \max(\eta/\gamma^{r},\ \eta_{\min})\nabla f^{S_k}(\btheta)$. 
	\State $\btheta \leftarrow \btheta + \bv$.
    \State $r \leftarrow r+1$.
  \EndFor
  \State epoch $\leftarrow$ epoch $ +\ 1$.
  \State Calculate the function value $h$ of the validation set.
  \If{($h < (1-X)\times$ lowest\_value)}
    \State counter $\leftarrow N$.
  \Else
    \State counter $\leftarrow$ counter $-\ 1$.
  \EndIf
  \State lowest\_value $\leftarrow \min(\text{lowest\_value},\ h)$.
  \EndWhile
  \end{algorithmic}
  \label{alg:sgd-batch}
\end{algorithm}
\renewcommand{\baselinestretch}{2}
\par Unfortunately, both SG and mini-batch SG methods have a well known issue in choosing a suitable learning rate and a momentum coefficient for different problems. We will conduct some experiments in Section \ref{sec:deep-exps}.

\section{Experiments}
\label{sec:deep-exps}
We consider the following data sets for experiments. All except {\sf Sensorless} come with training and test sets. We split {\sf Sensorless} as described below.
\begin{itemize}
\item
{\sf HIGGS}: This binary classification data set is from high energy physics applications. It is selected for our experiments because feedforward networks have been successfully applied \citep{PB14a}. 
Note that a scalar output $y$ is enough to represent two classes in a binary classification problem. Based on this idea, we set $n_L = 1$, and have each $y^i \in \{-1, 1\}$. 
The predicted outcome is the first class if $y \geq 0$ and is the second class if $y < 0$. This data set is mainly used in Section \ref{subsec:compare-sgd} for a comparison with results in \cite{PB14a}.
\item
{\sf Letter}: This set is from the Statlog collection \citep{DM94a} and we scale values of each feature to be in $[-1,1]$. 
\item
{\sf MNIST}: This data set for hand-written digit recognition \citep{YL98a} is widely used to benchmark classification algorithms.
We consider a scaled version, where every feature value is divided by $255$.
\item
{\sf Pendigits}: This data set is originally from \cite{FA96a}.
\item
{\sf Poker}: This data set is from UCI machine learning repository \citep{ML13a}. It has been studied by, for example, \cite{PL10c}.
\item
{\sf Satimage}: This set is from the Statlog collection \citep{DM94a} and we scale values of each feature to be in $[-1,1]$. 
\item
{\sf SensIT Vehicle}: This data set, from \cite{MD04a}, includes signals from acoustic and seismic sensors in order to classify the different vehicles.
We use the original version without scaling.
\item
{\sf Sensorless}: This data set is from \cite{FP13a}. We scale values of each feature to be in $[0,1]$, and then 
conduct stratified random sampling to select $10,000$ instances to be the test set and the rest of the data to be the training set.
\item
{\sf SVHN}: This data, originally from Google Street View images, consists of colored images of house numbers \citep{YN11a}. 
We scale the data set to $[0,1]$ by considering the largest and the smallest feature values of the entire data set.
\begin{equation*}
M \equiv \max_i \max_p(\bx_i)_p \text{ and } m \equiv \min_i \min_p(\bx_i)_p.
\end{equation*}
Then the $p$th element of $\bx_i$ is changed to
\begin{equation*}
(\bx_i)_p \leftarrow \frac{(\bx_i)_p - m} {M - m}.
\end{equation*}
\item
{\sf USPS}: This data set, from \cite{JJH94a}, is used on recognizing handwritten ZIP codes and we scale values of each feature to be in $[-1,1]$. 
\end{itemize}
All data sets, with statistics in Table \ref{table:deep-dataset}, are publicly available.\footnote{All data sets used can be found at \url{https://www.csie.ntu.edu.tw/~cjlin/libsvmtools/datasets/}.} Detailed settings for each data such as the network structure are given in Table \ref{table:exp1-data-setting}.
How to decide a suitable network structure is beyond the scope of this work, but if possible, we follow the setting in earlier works.
For example, we consider the structure in \cite{LW13a} for {\sf MNIST} and \cite{BN15b} for {\sf SVHN}.
From Table \ref{table:exp1-data-setting}, the model used for {\sf SVHN} is the largest. If the number of neurons in each layer is further increased,
then the model must be stored in different machines.
\begin{table}[t]
\caption{Summary of the data sets: $n_0$ is the number of features, $l$ is the number of training instances, $l_t$ is the number of testing instances, 
and $K$ is the number of classes.}
\label{table:deep-dataset}
\begin{center}
\begin{tabular}{l r r r r}
Data set & $n_0$ & $l$ & $l_t$ & $K$ \\
\hline
{\sf Letter} & 16 & 15,000 & 5,000 & 26 \\
\hline
{\sf MNIST} & 784 & 60,000 & 10,000 & 10 \\
\hline
{\sf Pendigits} & 16 & 7,494 & 3,498 & 10 \\
\hline
{\sf Poker} & 10 & 25,010 & 1,000,000 & 10 \\
\hline
{\sf Satimage} & 36 & 4,435 & 2,000 & 6 \\
\hline
{\sf SensIT Vehicle} & 100 & 78,823 & 19,705 & 3 \\
\hline
{\sf Sensorless} & 48 & 48,509 & 10,000 & 11 \\
\hline
{\sf SVHN} & 3,072 & 73,257 & 26,032 & 10 \\
\hline
{\sf USPS} & 256 & 7,291 & 2,007 & 10 \\
\hline
{\sf HIGGS} & 28 & 10,500,000 & 500,000 & 2
\end{tabular}
\end{center}
\end{table}

\begin{table}[t]
\begin{center}%
\caption{Details of the distributed network for each data. Sampling rate is the percentage of training data used to calculate the subsampled Gauss-Newton matrix.}%
\label{table:exp1-data-setting}
\begin{tabular}{l c c c c}%
Data set & Sampling rate & Network structure & Split structure & \# partitions\\
\hline
{\sf Letter} & $20\%$ & $16$-$300$-$300$-$300$-$300$-$26$ & $1$-$2$-$1$-$1$-$1$-$1$ & $7$\\
\hline
{\sf MNIST} & $20\%$ & $784$-$800$-$800$-$10$ & $1$-$1$-$3$-$1$ & $7$\\
\hline
{\sf Pendigits} & $20\%$ & $16$-$300$-$300$-$10$ & $1$-$2$-$2$-$1$ & $8$\\
\hline
{\sf Poker} & $20\%$ & $10$-$200$-$200$-$200$-$10$ & $1$-$1$-$1$-$1$-$1$ & $4$\\
\hline
{\sf SensIT Vehicle} & $20\%$ & $100$-$300$-$300$-$3$ & $1$-$2$-$2$-$1$ & $8$\\
\hline
{\sf Sensorless} & $20\%$ & $48$-$300$-$300$-$300$-$11$ & $1$-$2$-$1$-$2$-$1$ & $8$\\
\hline
{\sf Satimage} & $20\%$ & $36$-$1000$-$500$-$6$ & $1$-$2$-$2$-$1$ & $8$\\
\hline
{\sf SVHN} & $10\%$ & $3072$-$4000$-$4000$-$10$ & $3$-$2$-$2$-$1$ & $12$\\
\hline
{\sf USPS} & $20\%$ & $256$-$300$-$300$-$10$ & $1$-$2$-$2$-$1$ & $8$
\end{tabular}
\end{center}
\end{table}

\par We give parameters used in our algorithm. For the sparse initialization discussed in Section \ref{subsec:Sparse-Init}, among $n_{m-1}$ weights connected to a neuron in layer $m$, $\lceil \sqrt{n_{m-1}} \rceil$ are selected to have non-zero values. For the CG stopping condition \eqref{CG-stopping-cond}, we set $\sigma = 0.001$ and CG$_{\max} = 250$. Further, the minimal number of CG steps run at each partition, CG$_{\min}$, is set to be $3$. For the implementation of the Levenberg-Marquardt method, we set the initial $\lambda_1 = 1$. The (drop, boost) constants in \eqref{LM-rules} are ($2/3$, $3/2$).
For solving \eqref{linear-comb-now} to get the update direction after the CG procedure, we set $\varepsilon = 10^{-5}$ in \eqref{determinant}.

\subsection{Analysis of Distributed Newton Methods}
\label{subsec:Analysis-newton}
We have proposed several techniques to improve upon the basic implementation of the Newton method in a distributed environment. 
Here we investigate their effectiveness by considering the following methods.
Note that because of the high memory consumption of some larger sets, we always implement
the subsampled Hessian Newton method discussed in Section \ref{subsec:Subsampled}.
\begin{enumerate}[1.]
  \item {\sl subsampled-GN}: we use the whole subsampled Gauss-Newton matrix defined in \eqref{sampled-Gauss} to conduct the matrix-vector product in the CG procedure 
and then solve \eqref{linear-comb-now} to get the update direction after the CG procedure \citep{CCW15a}.
  \item {\sl diag}: it is the same as {\sl subsampled-GN} except that only diagonal blocks of the subsampled Gauss-Newton matrix are used; see \eqref{block-gaussnewton}. 
  \item {\sl diag $+$ sync $50\%$}: it is the same as {\sl diag} except that we consider the technique in Section \ref{subsec:Synchronization} to reduce the synchronization time. 
We terminate the CG procedure when $50\%$ of partitions have reached their local stopping conditions \eqref{cg-stopping-sample}.
  \item {\sl diag $+$ sync $25\%$}: it is the same as {\sl diag $+$ sync $50\%$} except that we terminate the CG procedure when $25\%$ of partitions have reached 
their local stopping conditions \eqref{cg-stopping-sample}.
\end{enumerate}
For each of the above methods, we consider the following implementation details.
\begin{enumerate}[1.]
\label{cg-stop-discuss-1}
\item
We set $C = l$ as the regularization parameter.
\item
We run experiments on G1 type instances on Microsoft Azure 
and let each instance use only one core. If instances
are not virtual machines on the same computer, our 
setting ensures that each variable partition corresponds to one machine.

\item
    To make the computational cost in each partition as balanced as possible, in our experiments we choose our partitions such that the maximum ratio between the numbers of variables ($|T_m| \times |T_{m-1}|$) among any two partitions is as low as possible. For example, in {\sf Pendigits}, the largest partition has $150 \times 150 = 22,500$ weight variables, and the smallest partition has $150 \times 10 = 1,500$ weight variables, with their ratio being $22500 / 1500 = 15$. For most data sets, the ratio is between 10 and 100 but not lower because the numbers of classes is relatively small, making the number of variables in the partitions involving the output layer smaller than those in other partitions.

\end{enumerate}
\setcounter{subfigure}{0}
\begin{center}
\begin{figure}[t]
	\vspace*{-1cm}
	\begin{center}
        \begin{subfigure}{\textwidth}
                \begin{tabular}{c c}
                \begin{subfigure}{0.5\textwidth}
                \includegraphics[width=\textwidth]{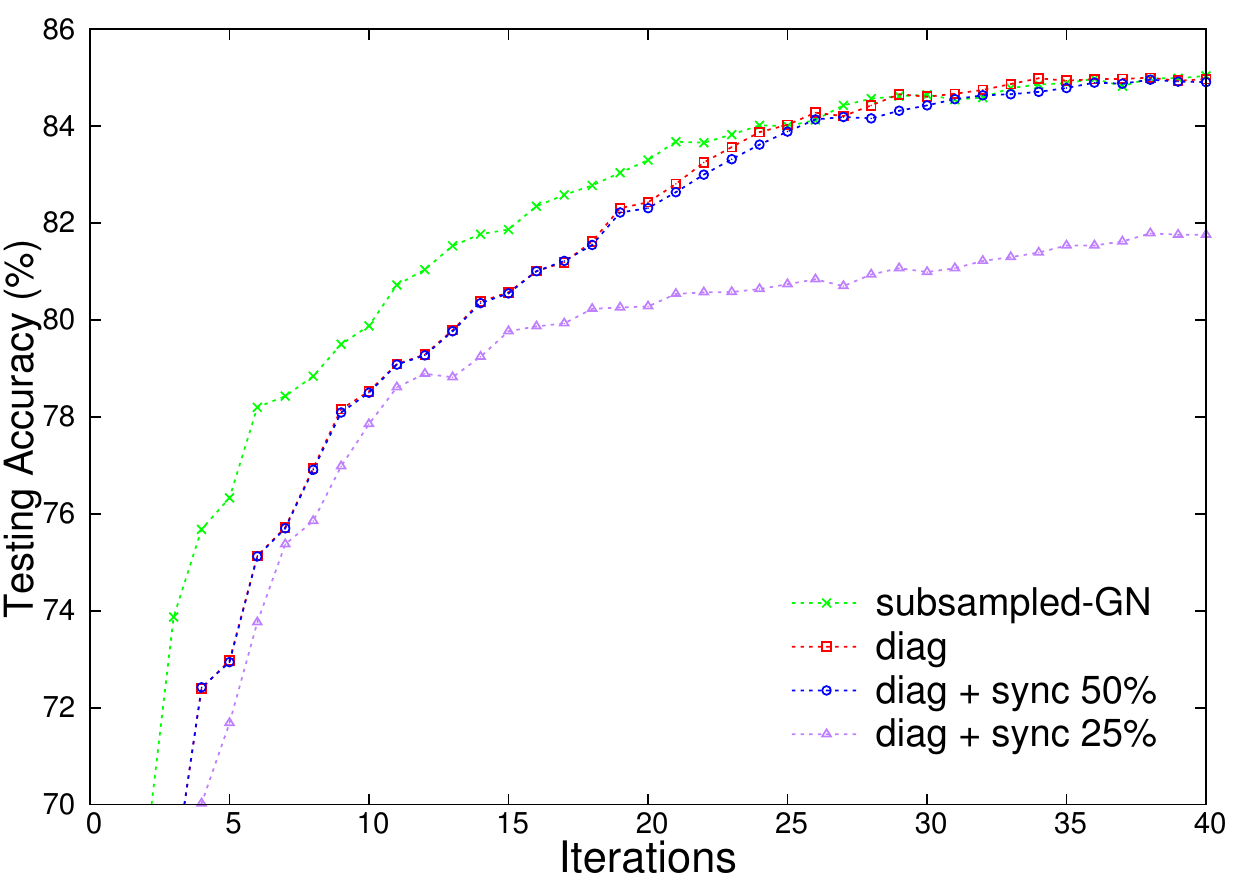}
                \end{subfigure}
				\begin{subfigure}{0.5\textwidth}
                \includegraphics[width=\textwidth]{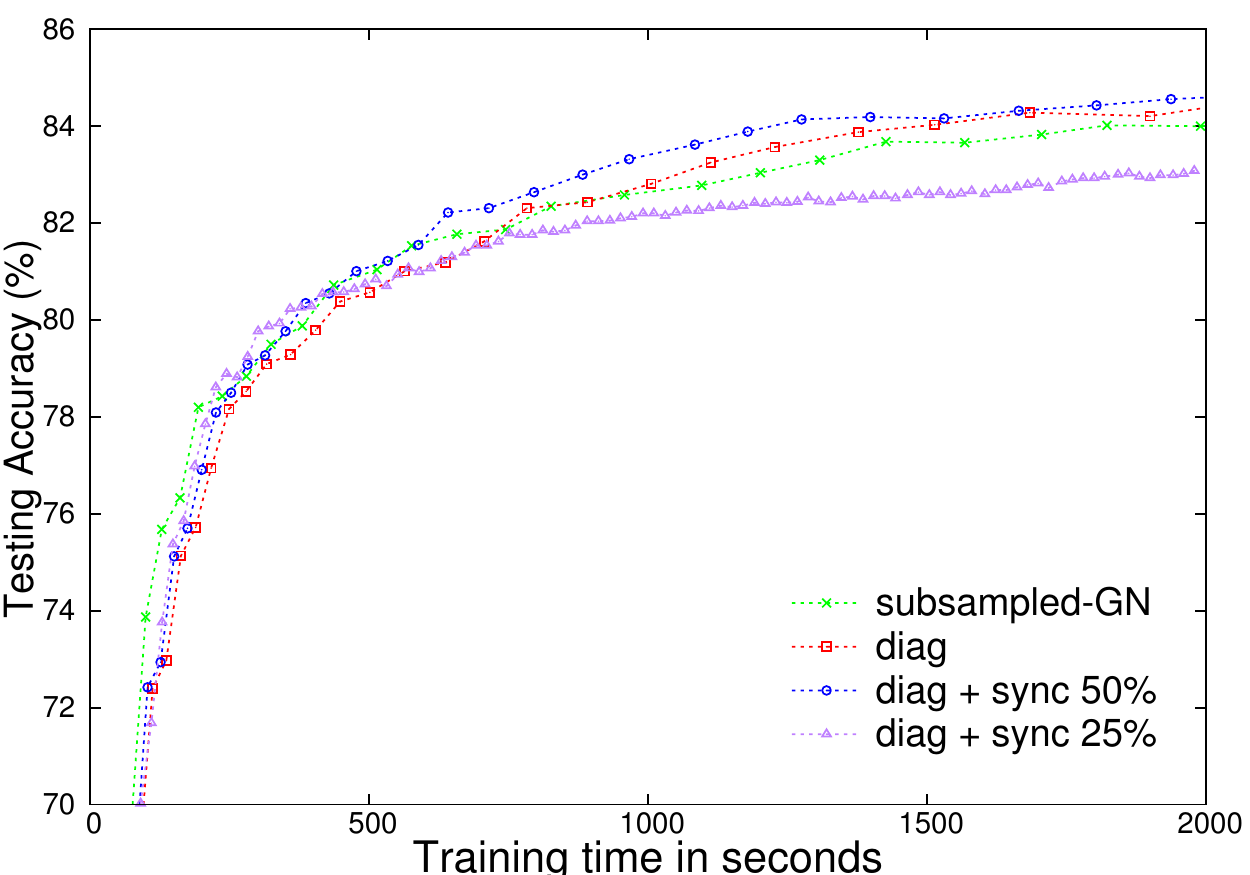}
                \end{subfigure}
                \end{tabular}
		\vspace*{-0.6cm}
        \caption{{\sf SensIT Vehicle}}
        \end{subfigure}
    \end{center}
	\vspace*{-0.55cm}
	\begin{center}
        \begin{subfigure}{\textwidth}
                \begin{tabular}{c c}
                \begin{subfigure}{0.5\textwidth}
                \includegraphics[width=\textwidth]{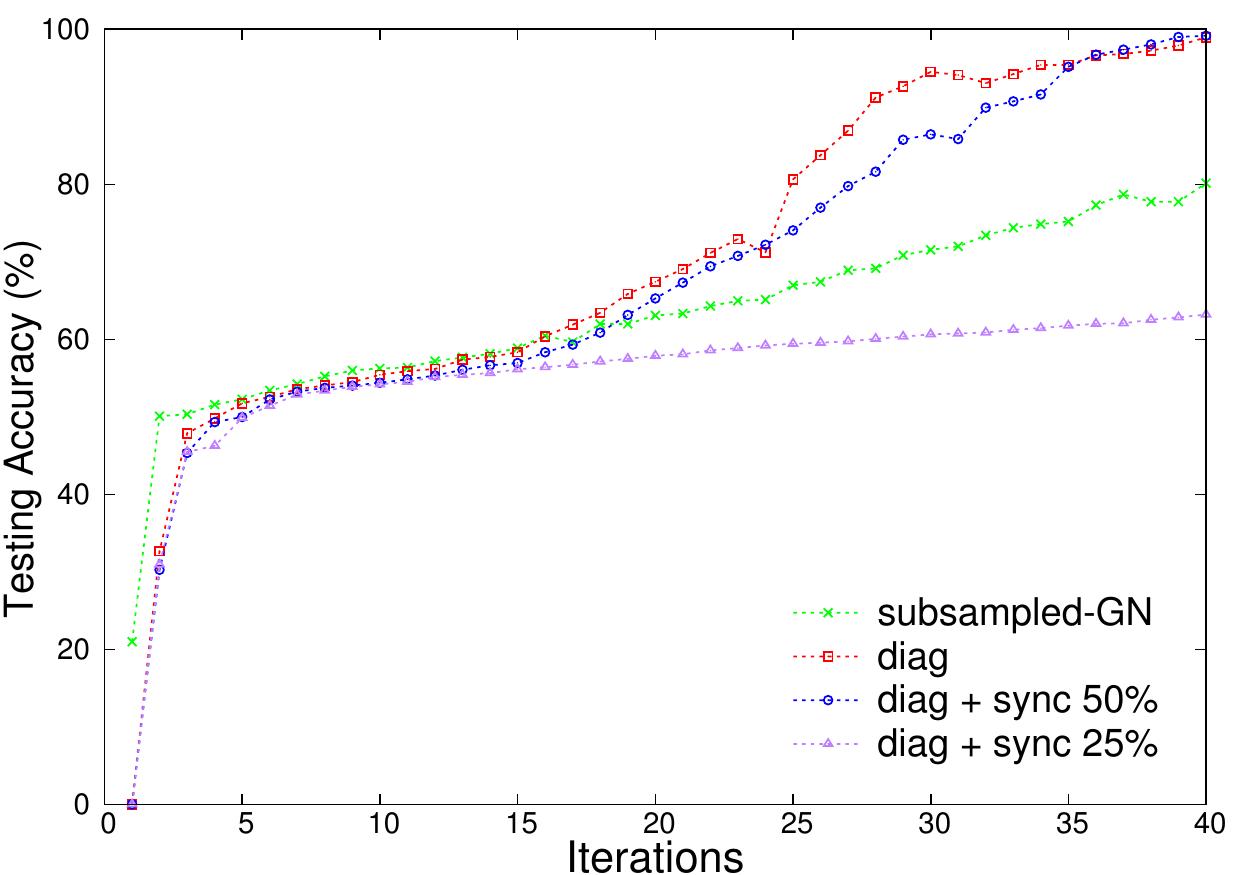}
                \end{subfigure}
				\begin{subfigure}{0.5\textwidth}
                \includegraphics[width=\textwidth]{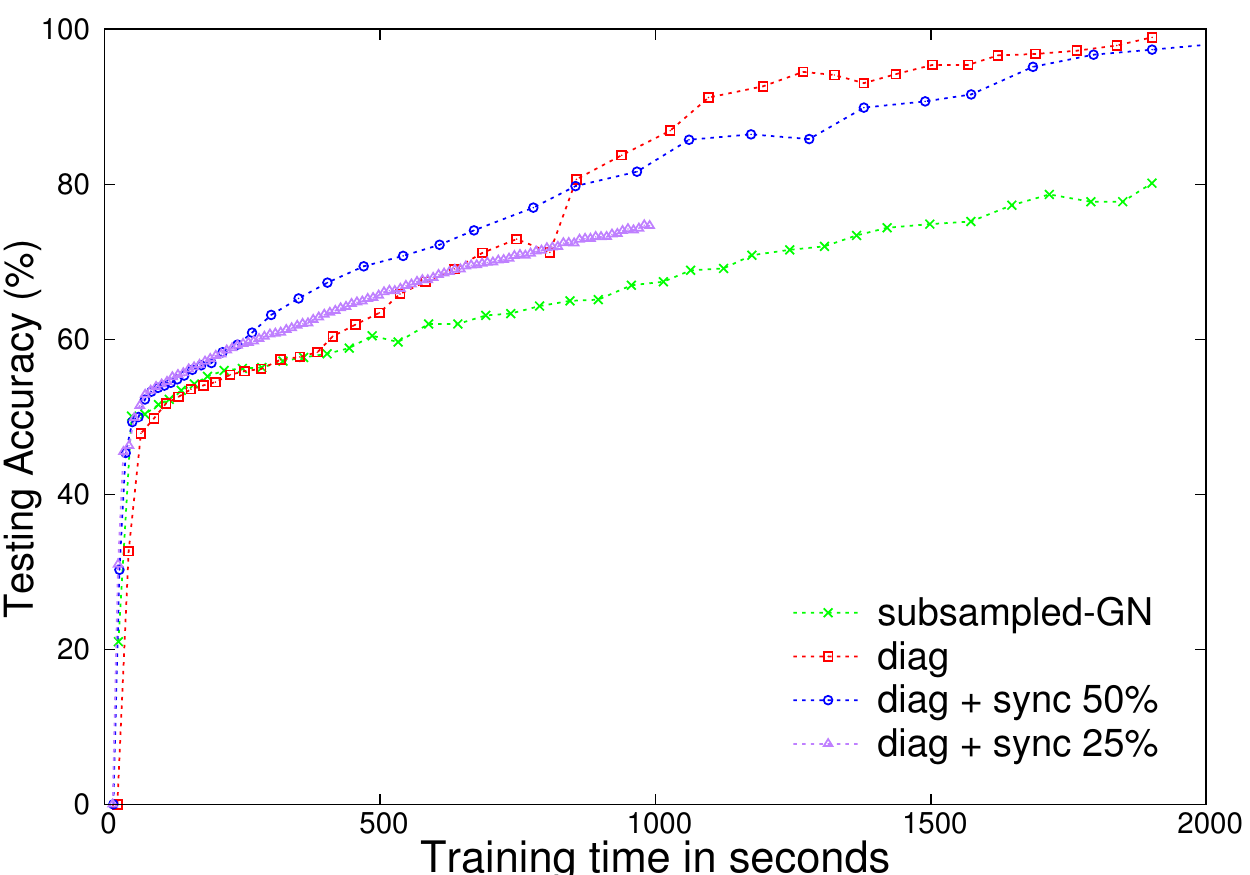}
                \end{subfigure}
                \end{tabular}
		\vspace*{-0.6cm}
        \caption{{\sf poker}}
        \end{subfigure}
    \end{center}
	\vspace*{-0.85cm}
	\begin{center}
        \begin{subfigure}{\textwidth}
                \begin{tabular}{c c}
                \begin{subfigure}{0.5\textwidth}
                \includegraphics[width=\textwidth]{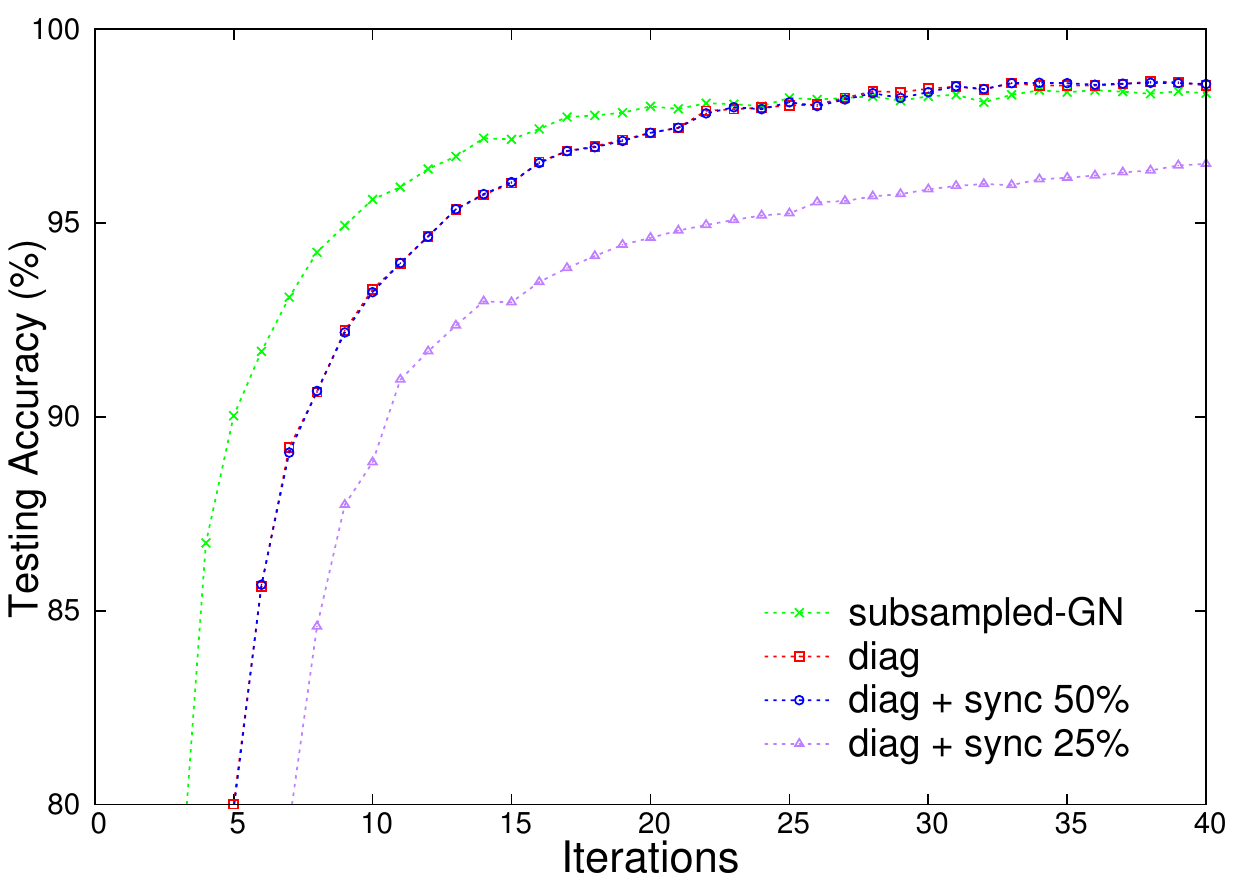}
                \end{subfigure}
				\begin{subfigure}{0.5\textwidth}
                \includegraphics[width=\textwidth]{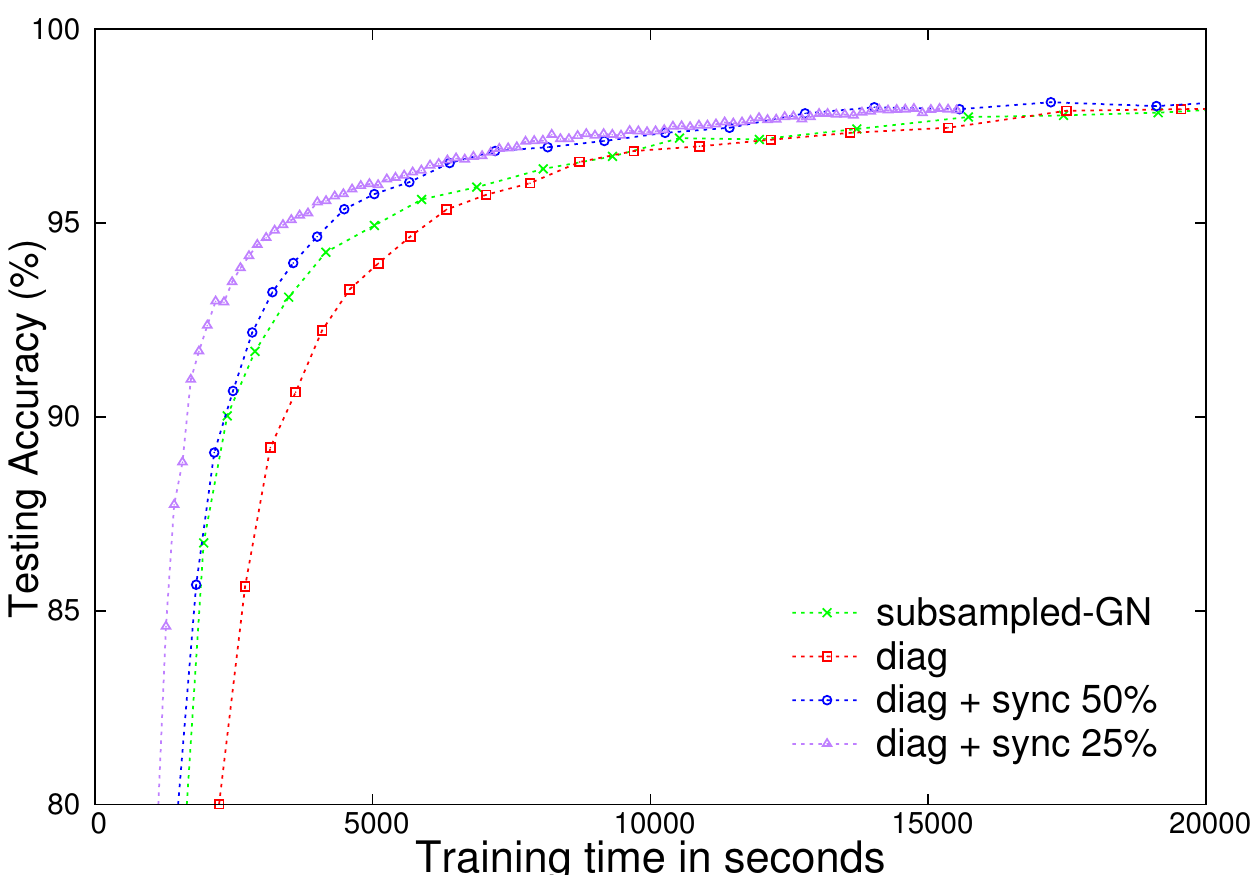}
                \end{subfigure}
                \end{tabular}
		\vspace*{-0.6cm}
        \caption{{\sf MNIST}}
        \end{subfigure}
    \end{center}
	\vspace*{-0.65cm}
	\begin{center}
        \begin{subfigure}{\textwidth}
                \begin{tabular}{c c}
                \begin{subfigure}{0.5\textwidth}
                \includegraphics[width=\textwidth]{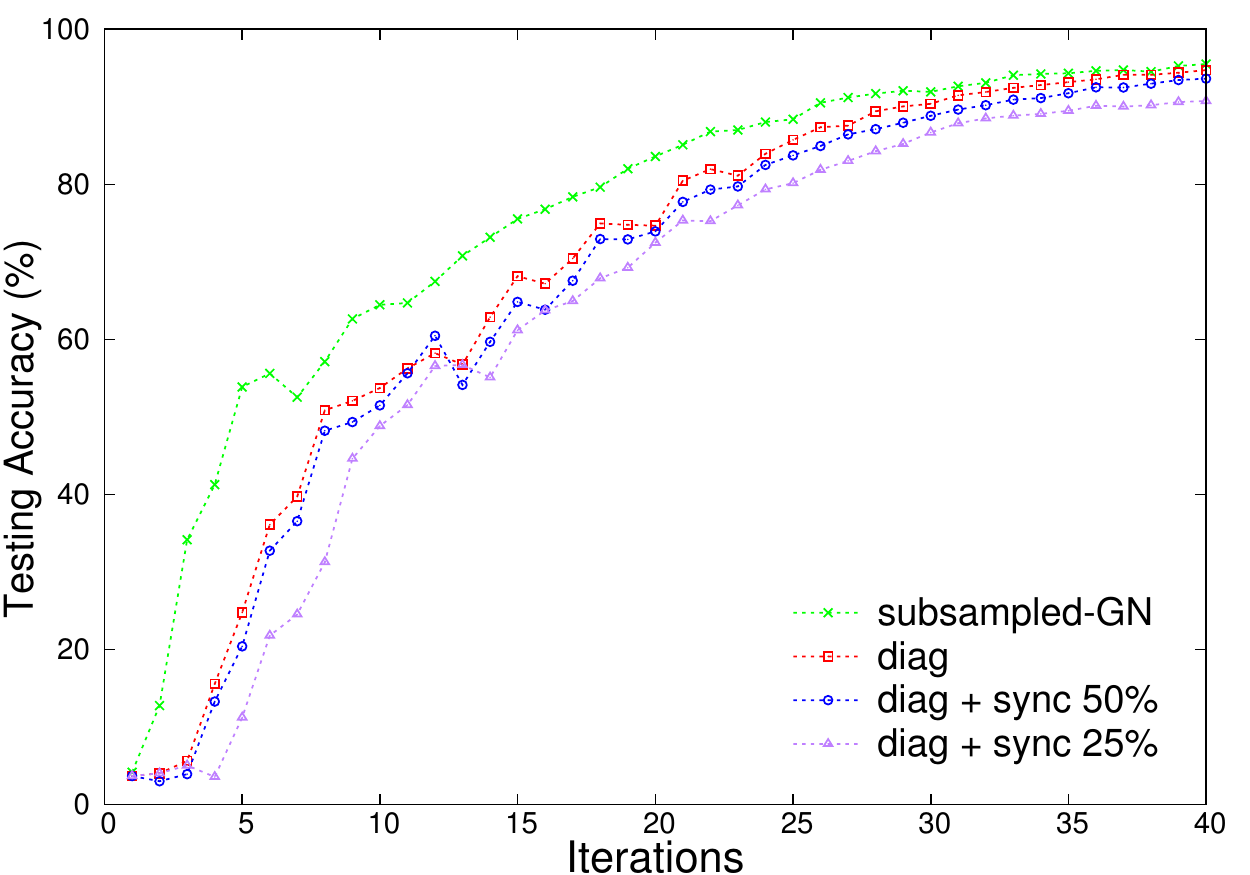}
                \end{subfigure}
				\begin{subfigure}{0.5\textwidth}
                \includegraphics[width=\textwidth]{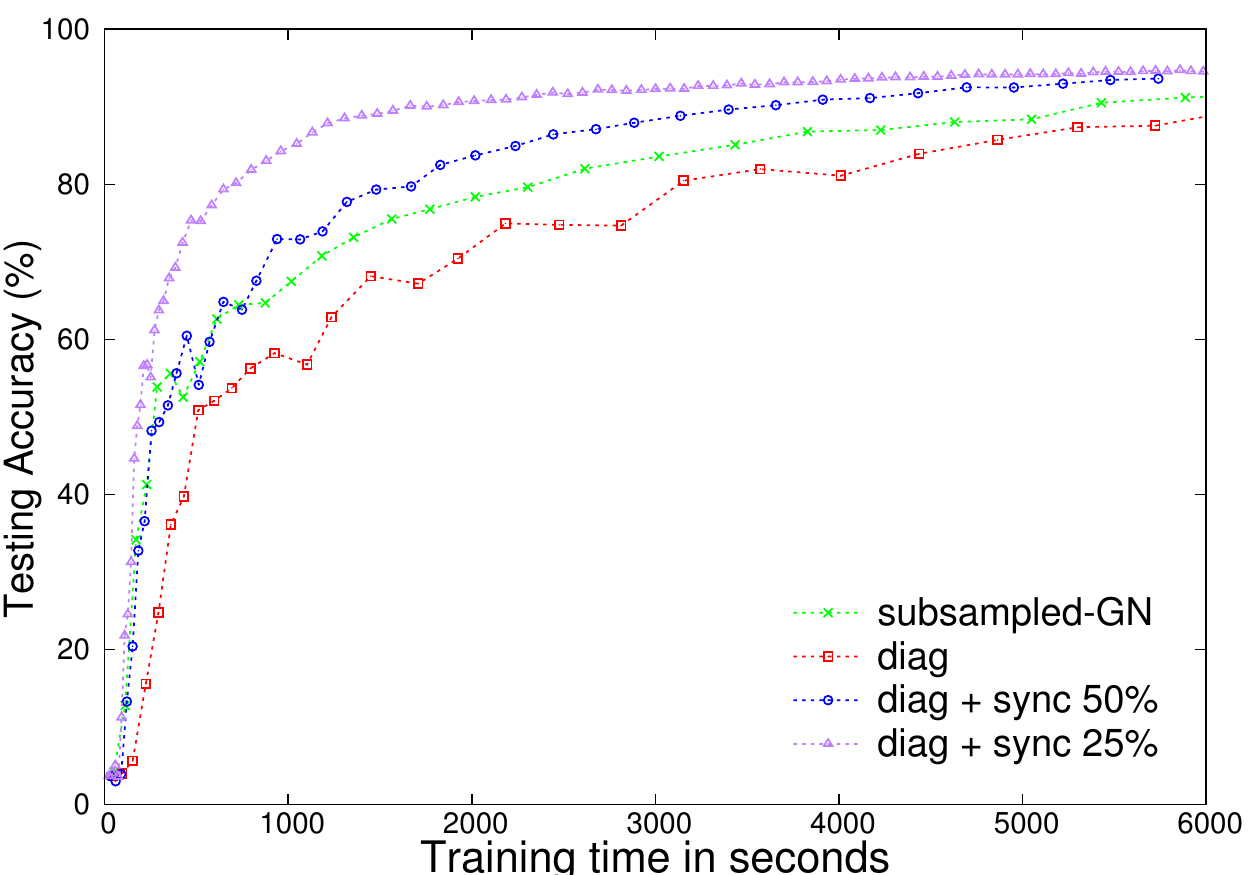}
                \end{subfigure}
                \end{tabular}
		\vspace*{-0.6cm}
        \caption{{\sf Letter}}
        \end{subfigure}
    \end{center}
\end{figure}
\end{center}

\begin{center}
\begin{figure}[t]
	\vspace*{-1cm}
	\begin{center}
        \begin{subfigure}{\textwidth}
                \addtocounter{subfigure}{4} 
                \begin{tabular}{c c}
                \begin{subfigure}{0.5\textwidth}
                \includegraphics[width=\textwidth]{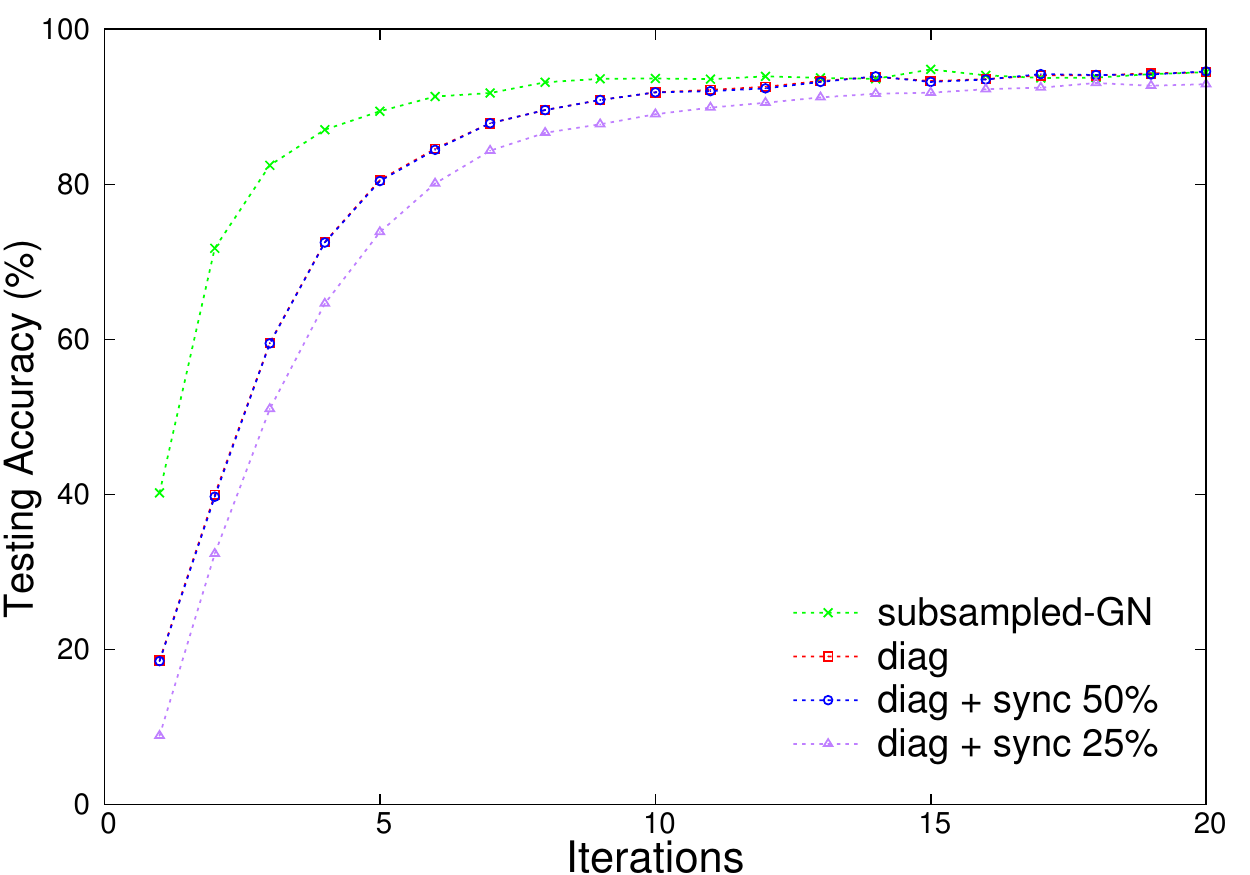}
                \end{subfigure}
				\begin{subfigure}{0.5\textwidth}
                \includegraphics[width=\textwidth]{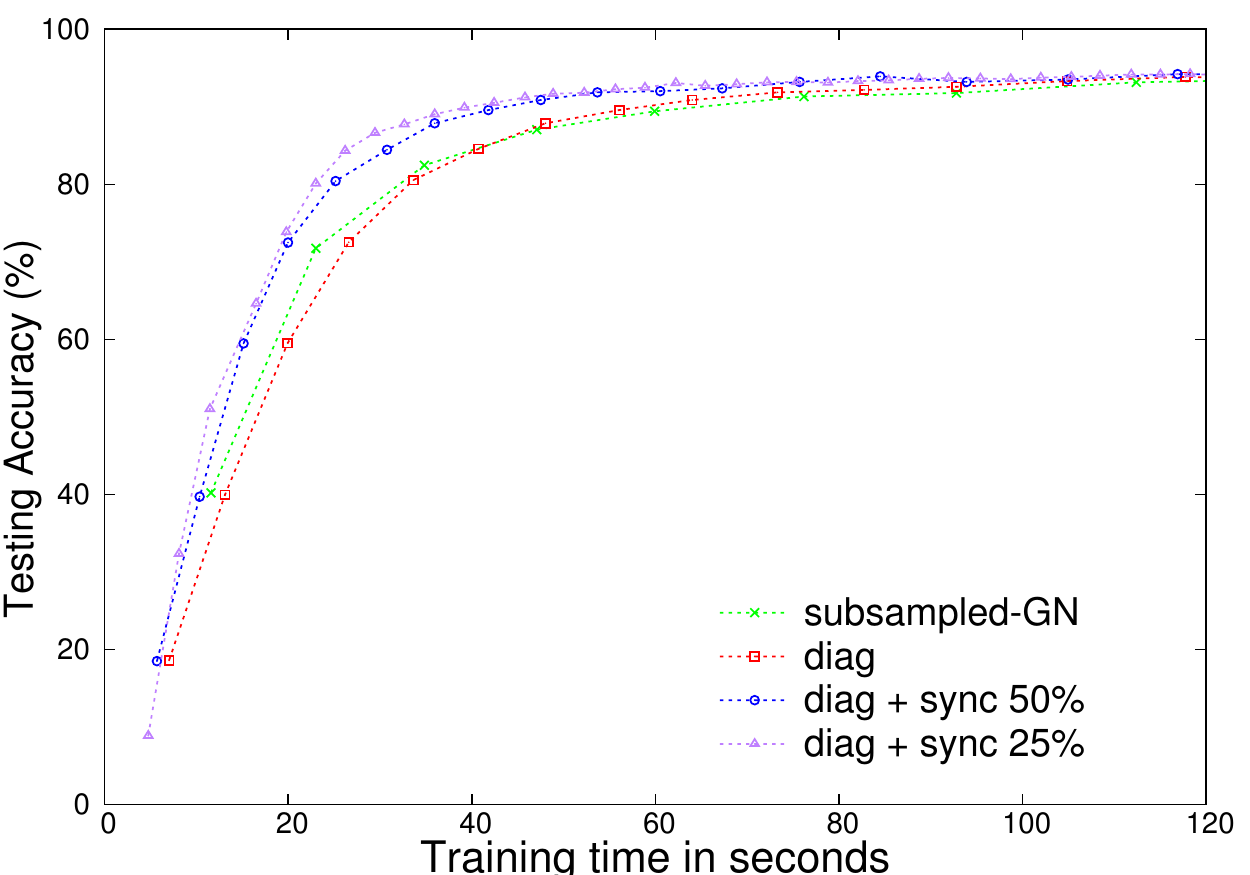}
                \end{subfigure}
                \end{tabular}
		\vspace*{-0.6cm}
        \caption{{\sf USPS}}
        \end{subfigure}
    \end{center}
	\vspace*{-0.55cm}
	\begin{center}
        \begin{subfigure}{\textwidth}
                \begin{tabular}{c c}
                \begin{subfigure}{0.5\textwidth}
                \includegraphics[width=\textwidth]{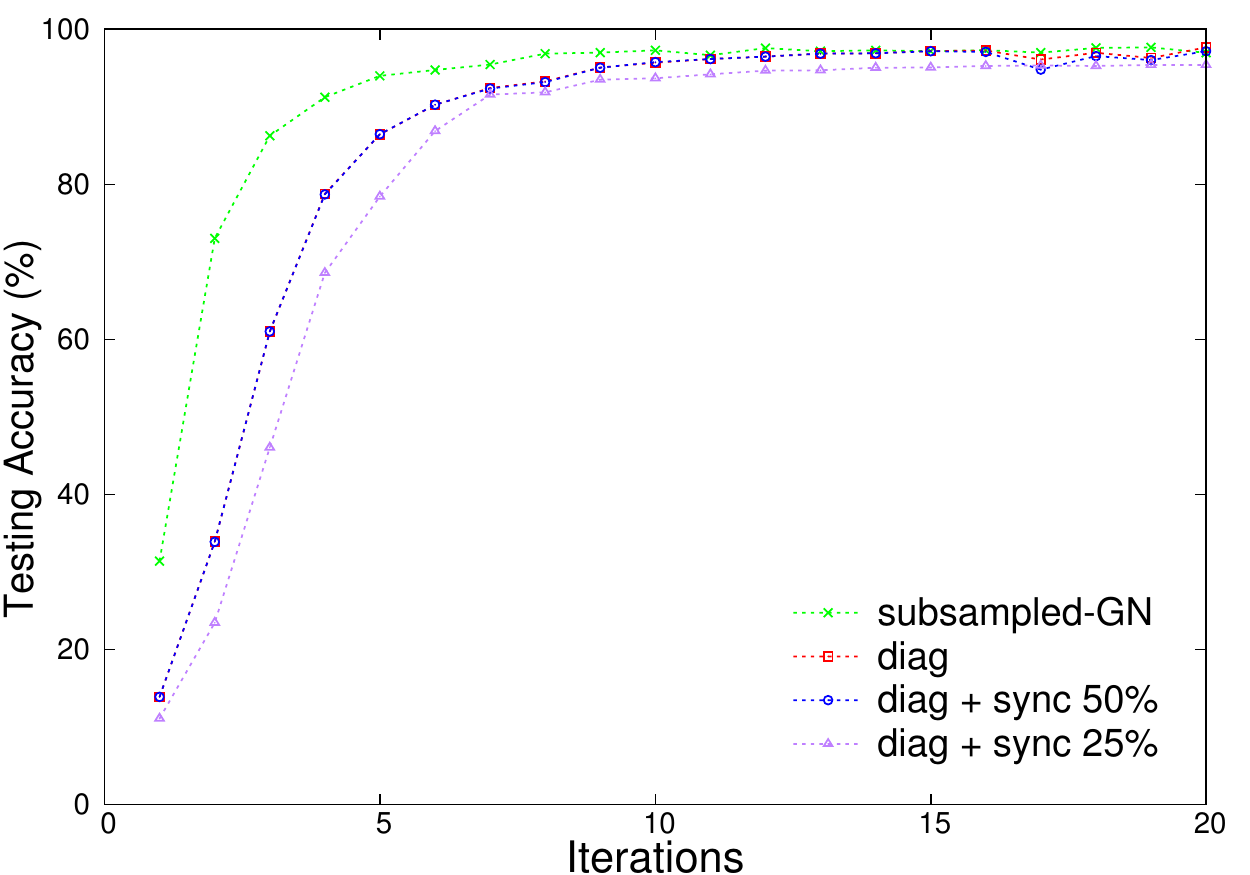}
                \end{subfigure}
				\begin{subfigure}{0.5\textwidth}
                \includegraphics[width=\textwidth]{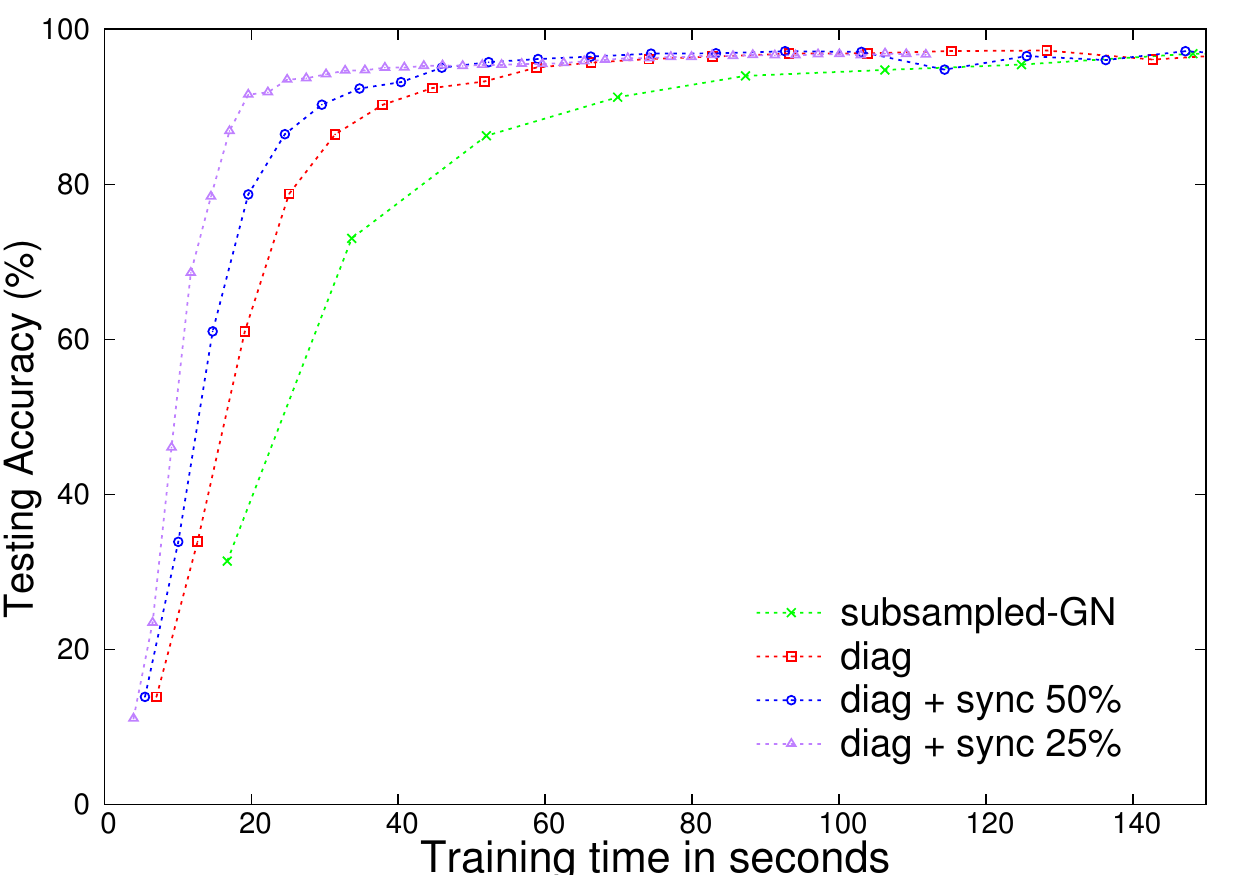}
                \end{subfigure}
                \end{tabular}
		\vspace*{-0.6cm}
        \caption{{\sf Pendigits}}
        \end{subfigure}
    \end{center}
	\vspace*{-0.85cm}
	\begin{center}
        \begin{subfigure}{\textwidth}
                \begin{tabular}{c c}
                \begin{subfigure}{0.5\textwidth}
                \includegraphics[width=\textwidth]{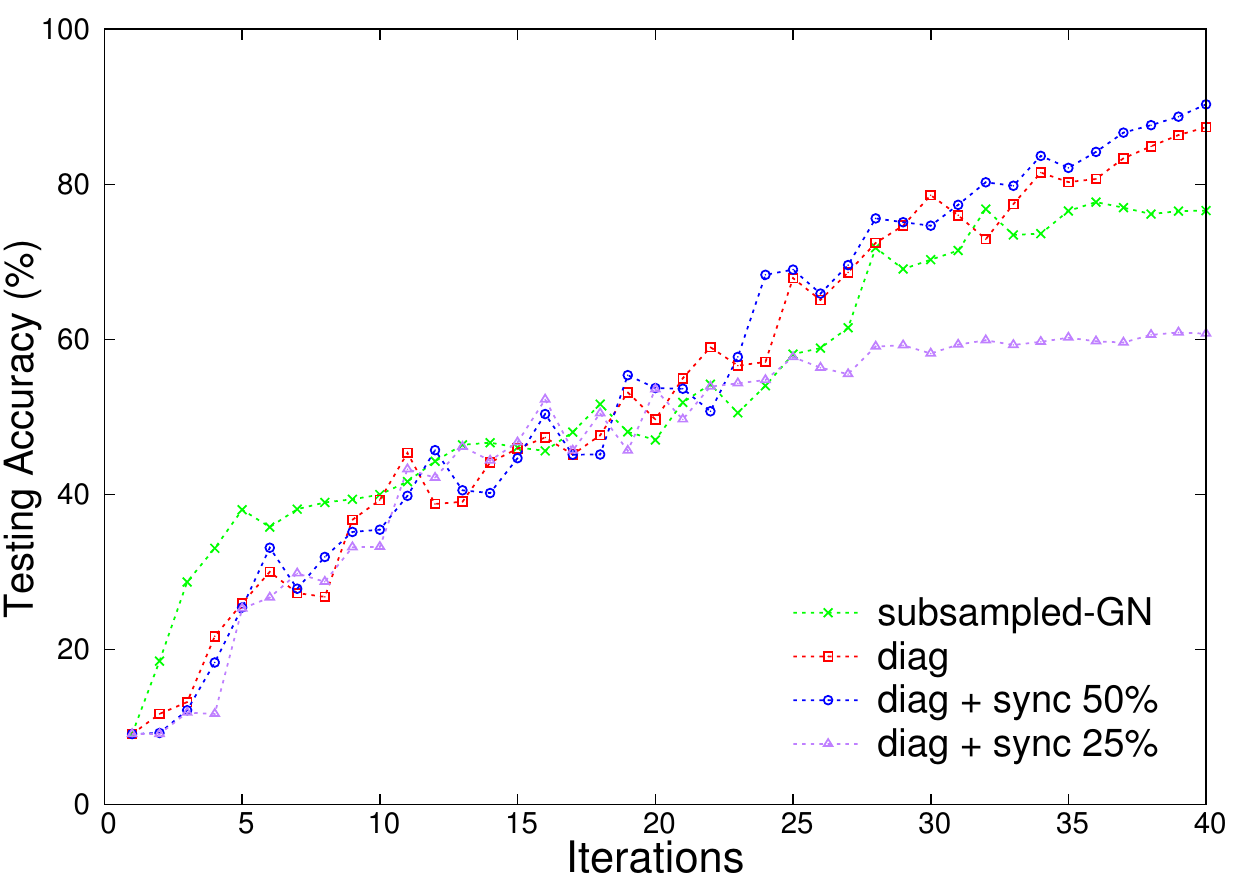}
                \end{subfigure}
				\begin{subfigure}{0.5\textwidth}
                \includegraphics[width=\textwidth]{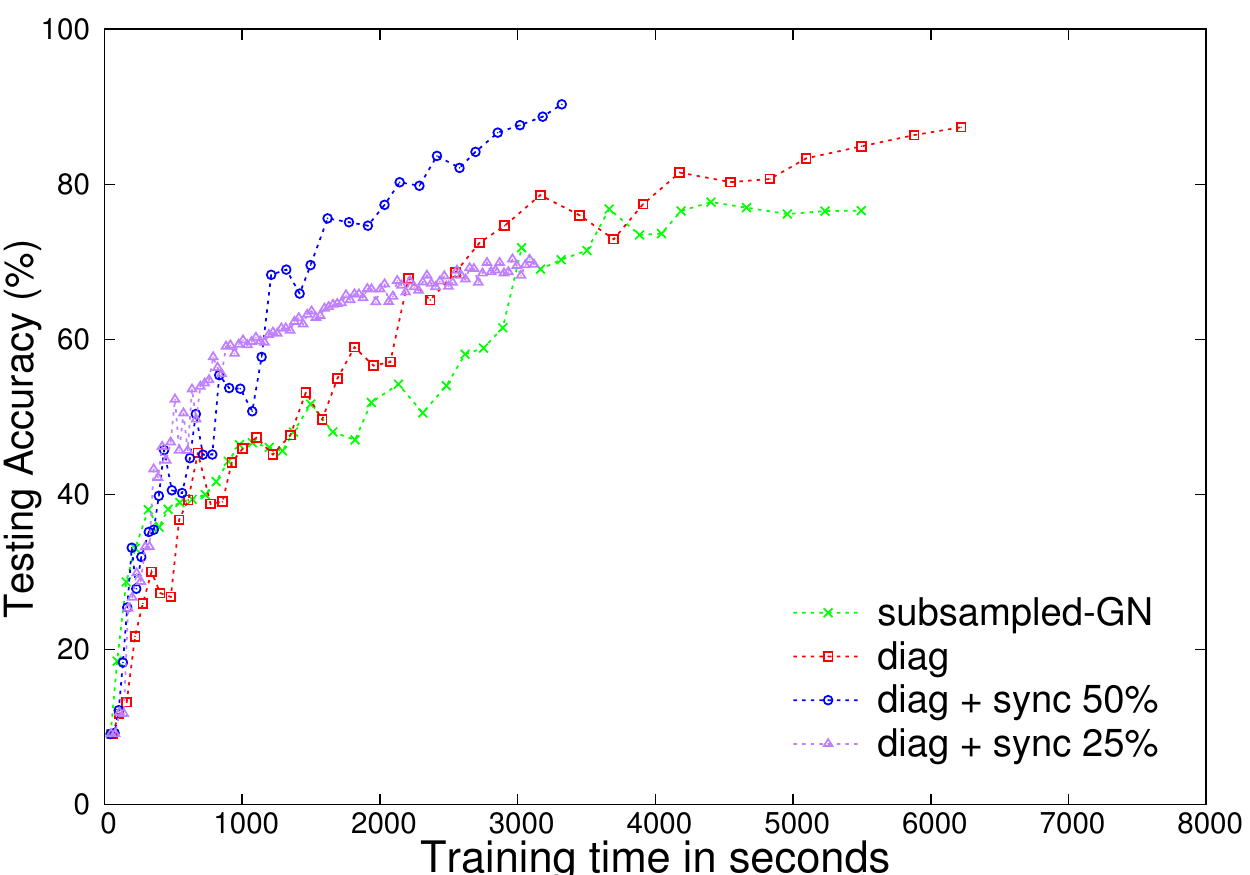}
                \end{subfigure}
                \end{tabular}
		\vspace*{-0.6cm}
        \caption{{\sf Sensorless}}
        \end{subfigure}
    \end{center}
	\vspace*{-0.65cm}
	\begin{center}
        \begin{subfigure}{\textwidth}
                \begin{tabular}{c c}
                \begin{subfigure}{0.5\textwidth}
                \includegraphics[width=\textwidth]{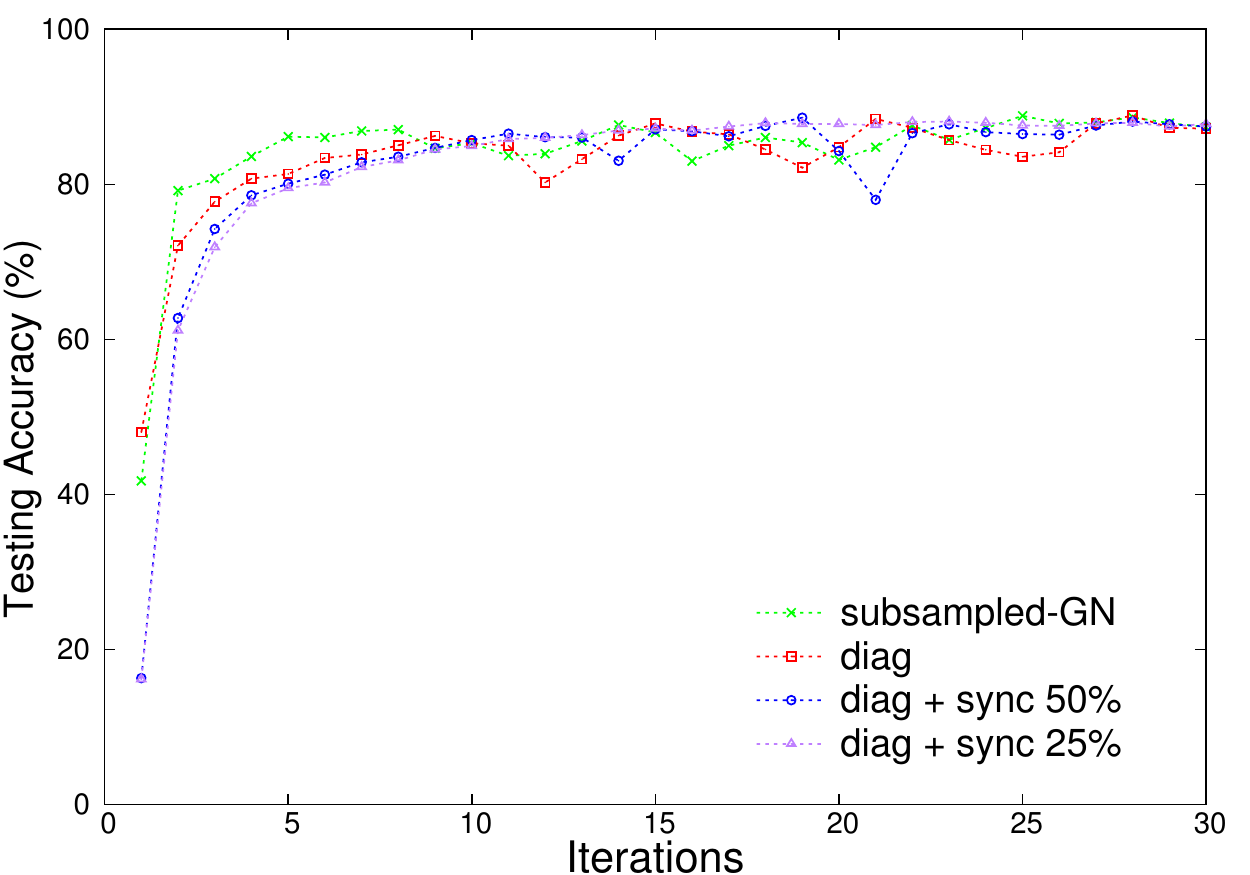}
                \end{subfigure}
				\begin{subfigure}{0.5\textwidth}
                \includegraphics[width=\textwidth]{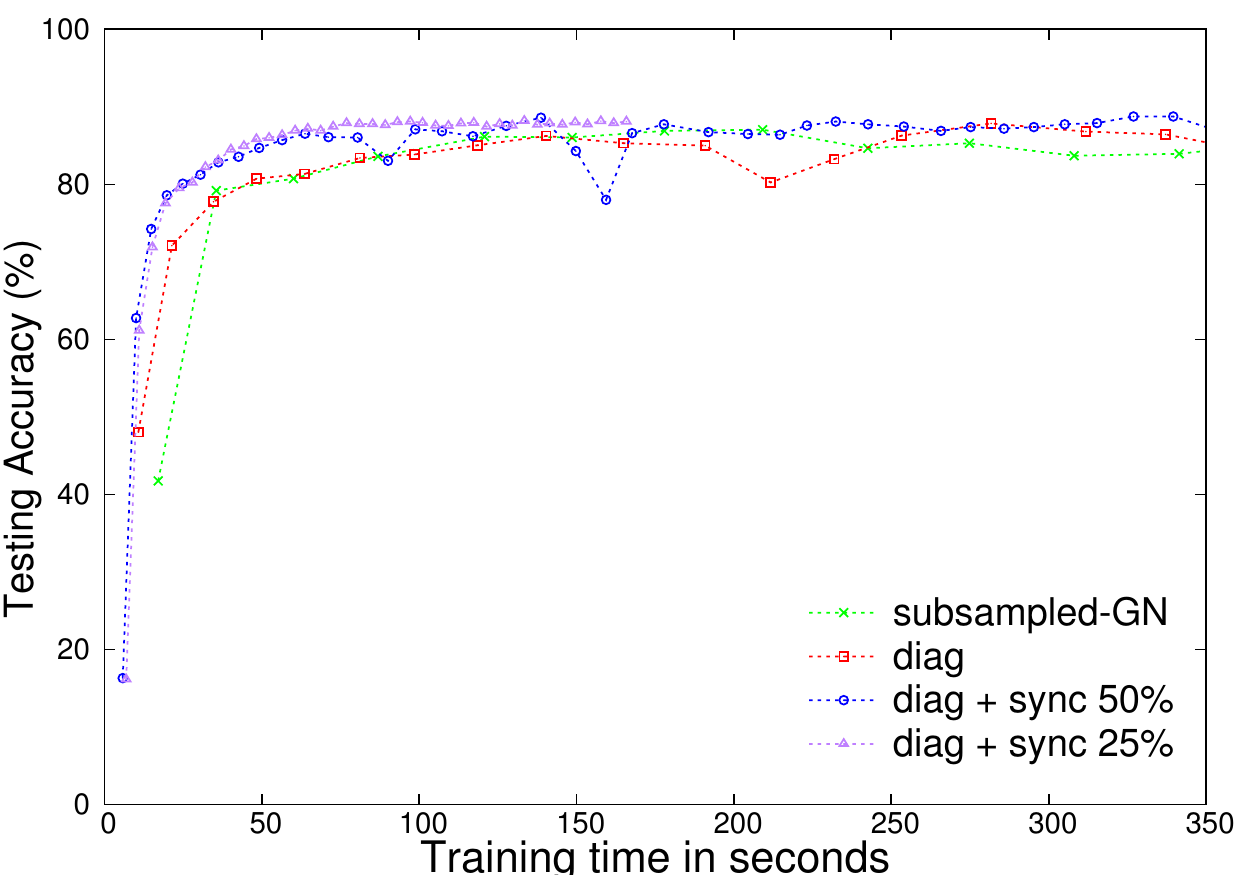}
                \end{subfigure}
                \end{tabular}
		\vspace*{-0.6cm}
        \caption{{\sf Satimage}}
        \end{subfigure}
    \end{center}
\vspace*{-1.2cm}
\caption{A comparison of different techniques to implement distributed Newton methods. Left: testing accuracy versus number of iterations.
Right: testing accuracy versus training time.}
\label{fig::MNIST-HIGGS}
\end{figure}
\end{center}

In Figure \ref{fig::MNIST-HIGGS}, we show the comparison results and have the following observations.
\begin{enumerate}[1.]
\item For test accuracy versus number of iterations, {\sl subsampled-GN} in general has the fastest convergence rate.
The reason should be that the direction in {\sl subsampled-GN} by solving the linear system \eqref{sample-gaussnewton} is closer to the full Newton direction than other methods, 
which consider further approximations of the Gauss-Newton matrix or the early termination of the CG procedure.
However, the cost per iteration is high, so for training time we see that {\sl subsampled-GN} may become worse than other approaches.
\item The early termination of the CG procedure can effectively reduce the cost per iteration. However, if we stop the CG procedure too early, the total training time may even increase. For example,
\begin{center} {\sl diag $+$ sync $25\%$} \end{center}
is generally the fastest in the beginning because of the least cost per iteration. It is still the fastest in the end for {\sf MNIST}, {\sf Letter}, {\sf USPS}, {\sf Satimage}, and {\sf Pendigits}. However, it has the slowest final convergence for {\sf SensIT Vechicle}, {\sf Poker}, and {\sf Sensorless}.
Take the data set {\sf Poker} as an example. As listed in Table \ref{table:exp1-data-setting}, the variables are split into four partitions, and the CG procedure stops if one partition (i.e., $25\%$ of the partitions) reaches its local stopping condition. This partition may have the lightest computational load or is the earliest one to start solving the local linear system.%
\footnote{Note that because of the backward process in Section \ref{subsec:Distributed-Jacobian}, the partitions corresponding to the last two layers begin their CG procedures earlier than the others.} %
Thus the other partitions may not have run enough CG iterations.
\par The approach \begin{center} {\sl diag + sync 50\%} \end{center} does not terminate the CG procedure that early. Overall we find that it is efficient and stable. Therefore, in subsequent comparisons with stochastic gradient methods, we use it as the setting of our Newton method.
\end{enumerate}
\label{cg-stop-discuss-2}
\par Because of the space consideration, we have evaluated only some techniques proposed in Section \ref{sec:Reduce-compu-communi}.
For the following two techniques we leave details in Sections VI and VII of the supplementary materials.
\begin{enumerate}[1.]
\item
In Section \ref{subsec:Subsampled}, we propose combining $\bd^k$ and $\bd^{k-1}$ as the update direction. We show that this technique is very effective.
\item
We mentioned in Section \ref{subsec:summary-procedure} that line search and the Levenberg-Marquardt (LM) method may not be both needed.
Our preliminary results show that the training speed is improved when both techniques are applied.
\end{enumerate}

\subsection{Comparison with Stochastic Gradient Methods and Support Vector Machines (SVM)}
\label{subsec:multiclass}
In this section, we compare our methods with SG methods and SVMs, which are popularly used for multi-class classification.
Settings of these methods are described as follows.
\begin{enumerate}[1.]
\item
{\sl Newton}: for our method  we use the setting {\sl diag $+$ sync $50$\%} considered in Section \ref{subsec:Analysis-newton} and let $C = l$. 
\item
{\sl SVM} \citep{BB92a}: We consider the RBF kernel. 
\begin{equation*}
K(\bx^i,\bx^j) = e^{-\gamma||\bx^i-\bx^j||^2},
\end{equation*}
where $\bx^i$ and $\bx^j$ are two data instances, and $\gamma$ is the kernel parameter chosen by users. Note that {\sl SVM} solves an optimization problem similar to \eqref{obj-function}, so the regularization parameter, $C$, must be decided as well. We conduct five-fold cross validation on the training set to select the best $C \in \{2^{-5}l,2^{-3}l,\ldots,2^{15}l\}$ and the best $\gamma \in \{2^{-15},2^{-13},\ldots,2^{3}\}$.%
\footnote{Here we consider an SVM formulation represented as \eqref{intro-obj}. In the form considered in {\sl LIBSVM}, the two terms $C$ and $1/l$ are combined together, so $C/l$ is the actual parameter to be selected. For {\sf SVHN} because of the lengthy time for parameter selection, we selected only $10,000$ instances by stratified sampling to conduct the five-fold cross validation.} %
We use the library {\sl LIBSVM} \citep{CC01a} for training and prediction.
\item
{\sl SG}: We use the code from \cite{PB14a}, which implements Algorithm \ref{alg:sgd-batch}. %
The objective function is the same as \eqref{obj-function}.%
\footnote{Following \cite{PB14a}, we regularized only the weights but not the biases. Through several experiments, we found that the performance is similar with/without the regularization of the biases.} %
The network structure for each data set is identical to the corresponding one used in {\sl Newton}, and we also set the regularization parameter $C=l$. The major modification we make is that we replace their activation functions with ours. In \cite{PB14a}, the authors use $\tanh$ as their activation functions in layers $1,\ldots,L-1$ and the sigmoid function in layer $L$, while in our experiments of Newton methods in Section 8.1, we use the sigmoid function in layers $1,\ldots,L-1$ and the linear function in layer $L$. The initial learning rate is selected from \{$0.05,$ $0.025,$ $0.01,$ $0.005,$ $0.002,$ $0.001$\} by the five-fold cross validation. After the initial learning rate has been selected, we conduct the training process to generate a model for the prediction on the test set.
\end{enumerate}

As regards the stopping condition for the training process, we terminate the {\sl Newton} method at the $100$th iteration. For {\sl SG}, it terminates after a minimal number of epochs have been conducted and the objective function value on the validation set does not improve much within the last $N$ epochs (see Algorithm \ref{alg:sgd-batch}). To implement the stopping condition, for {\sl SG} we split the input training set into 90\% for training and 10\% for validation.%
\footnote{Note that in the CV procedure we also need a stopping condition in training each sub-problem. We do an $80$-$20$ split of every four folds of data so that the $20\%$ of data are used to implement the stopping condition.} %
For {\sl SVM}, we use the default stopping condition of {\sl LIBSVM}.%
\footnote{
{\sl LIBSVM} terminates when the violation of the optimality condition calculated based on the gradient is smaller than a tolerance. 
}

\par Here we also investigate the effect of the initialization by considering the following two settings.
\begin{enumerate}[1.]
\item The sparse initialization discussed in Section \ref{subsec:Sparse-Init}.
\item The dense initialization discussed in \cite{PB14a}. The initial weights are drawn from the normal distribution $\mathcal{N}(0, 0.1^2)$ for the first layer, $\mathcal{N}(0, 0.001^2)$ for the output layer, and $\mathcal{N}(0, 0.05^2)$ for other hidden layers. The biases are initialized as zeros.
\end{enumerate}
To make a fair comparison, for each setting, {\sl Newton} and {\sl SG} are trained with the same initial weights and biases.

\par We present a comparison on test accuracy in Table \ref{table:results-poker-vehicle}, and make the following observations. 
\begin{enumerate}[1.]
\item For neural networks, the sparse initialization usually results in better accuracy than the dense initialization does. The difference can be huge in some cases, such as training using {\sl SG} on the data set {\sf Letter}. The low accuracy of the densely initialized {\sl SG} on {\sf Letter} may be because of the poor differentiation between neurons in dense initialization \citep{JM10a}. Other possible causes include the vanishing gradient problem \citep{YB94a}, or that the activations are trapped in the saturation regime of the sigmoid function \citep{XG10a}. Note that the impact of the initialization scheme on the {\sl Newton} method is much weaker.%
\footnote{We observe similar phenomena in the experiments with {\sf HIGGS} later in Section \ref{subsec:compare-sgd}. See Table \ref{table:higgs-result}.}







\item
    Between {\sl SG} and {\sl Newton}, if sparse initialization is used, we can see that {\sl Newton} generally gives higher accuracy.
\item 
If sparse initialization is used, our {\sl Newton} method for training neural networks gives similar or higher accuracy than {\sl SVM}.
In particular, the results are much better for {\sf Poker} and {\sf SVHN}.
\end{enumerate}

\par We compare our results on {\sf MNIST} with those reported in earlier works. \cite{LW13a} use a
fully connected neural network with two 800-neuron hidden layers to derive an error rate $1.36\%$,
under the setting of dense initialization,%
\footnote{In \cite{LW13a}, the initial weights are drawn from $\mathcal{N}(0,0.01)$, slightly different from the dense initialization we use.} %
sigmoid activations, and the dropout technique. By the same network structure and the same activation function, our error rate is $1.34\%$ at the $100$th iteration.

\par For {\sf SVHN}, we compare our results with \cite{BN15b}, in which the same network structure as ours is adopted, except that they use ReLU activations in the hidden layers. They choose the cross-entropy as their objective function, and utilize the dropout regularization. Under dense initialization,%
\footnote{In \cite{BN15b}, the initial weights $w^m_{t j}$ are drawn from $\mathcal{N}(0, 1 / n_{m-1})$, slightly different from the dense initialization we use.} %
they train their network with the Path-SGD method, which uses a proximal gradient method to solve the optimization problem. They report an accuracy slightly below $87\%$ (see their Figure 3), while the accuracy obtained by our {\sl Newton} method with sparse initialization is $83.12\%$.


\par For {\sf Poker}, we note that \cite{PL10c} uses {\sl abc-logitboost} to obtain a slightly higher accuracy, but his setting is different from ours. He expands the training set by including half of the test set, with the remaining half of the test set used for evaluation.

\par An issue found out in our experiments is that {\sl SG} is sensitive to the initial learning rate. 
In Table \ref{table:poker-lr}, we present the test accuracy of {\sl SG} under different initial rates for the {\sf Poker} problem.
Clearly an inappropriate initial learning rate can lead to much worse accuracy.

\begin{table}[t]
\begin{center}
\caption{Test accuracy of {\sl SVM}, {\sl Newton} and {\sl SG}. For {\sl SVM}, we also show parameters ($C$, $\gamma$) used. For {\sl SG}, we show (the initial learning rate, number of epochs to reach the stopping criterion). The bold-faced entries indicate the best accuracy obtained using the neural networks.}
\label{table:results-poker-vehicle}
\resizebox{\textwidth}{!}{
\begin{tabular}{llllll}
    & {\sl \quad\; SVM} & \multicolumn{4}{c}{Neural Networks} \\[-10pt]
  & & \multicolumn{2}{c}{Dense Initialization} & \multicolumn{2}{c}{Sparse Initialization} \\[-10pt]
	& & \multicolumn{1}{c} {\sl Newton} & \multicolumn{1}{c} {\sl SG} & \multicolumn{1}{c} {\sl Newton} & \multicolumn{1}{c} {\sl SG} \\
  \hline
  {\sf Letter} & $97.90\%\ (2^{7}l,2)$           & $90.26\%$  & $8.02\%\ (0.025,245)$  & {\bf 96.68\%} & $96.28\%\ (0.002,906)$ \\
  \hline                                                                                     
  {\sf MNIST} & $98.57\%\ (2^{3}l,2^{-5})$      & $98.52\%$ & $98.26\%\ (0.002,801)$ & {\bf 98.66\%} & $98.33\%\ (0.002,909)$ \\
  \hline                                                                                     
  {\sf Pendigits} & $98.06\%\ (2^7l,2^{-15})$   & $97.51\%$ & $97.71\%\ (0.001,513)$ & {\bf 97.83\%} & $97.71\%\ (0.002,1179)$ \\
  \hline
  {\sf Poker} & $58.78\%\ (2^{-1}l, 2^{-3})$    & $99.25\%$ & $99.24\%\ (0.005,316)$ & $99.25\%$ & {\bf 99.29\%}\ $(0.002,895)$ \\
  \hline                                                                                     
  {\sf Satimage} & $91.85\%\ (2l,2)$            & $89.35\%$ & $82.00\%\ (0.01,246)$  & {\bf 89.85\%} & $89.35\%\ (0.001,1402)$ \\
  \hline                                                                                     
  {\sf SensIT Vehicle} & $83.90\%\ (2l,2^{-1})$ & {\bf 85.16\%} & $83.34\%\ (0.01,311)$  & $84.60\%$ & $84.00\%\ (0.01,296)$ \\
  \hline                                                                                     
  {\sf Sensorless} & $99.83\%\ (2^{5}l,2^{3})$  & $97.19\%$ & $97.64\%\ (0.01,412)$  & {\bf 99.05\%} & $98.24\%\ (0.005,382)$ \\
  \hline                                                                                     
  {\sf SVHN} & $74.54\%\ (2^5l,2^{-7})$         & $80.96\%$      & $82.99\%\ (0.001,986)$  & {\bf 83.12\%} & $82.67\%\ (0.001,720)$\\
  \hline
  {\sf USPS} & $95.32\%\ (2^5l,2^{-5})$ & $95.17\%$&$94.97\%\ (0.025,395)$ & {\bf 95.27\%} & $95.07\%\ (0.001,1617)$
\end{tabular}
}
\end{center}
\end{table}

\begin{table}[t]
\begin{center}
\caption{Test accuracy on {\sf Poker} using {\sl SG} with different initial learning rates $\eta$. Dense initialization is used. 
Note that although $\eta = 0.005$ does not yield the highest test accuracy, it was selected for experiments in Table \ref{table:results-poker-vehicle} because of giving the highest CV accuracy.}
\label{table:poker-lr}
\begin{tabular}{l c c c c c c}
Initial learning rate $\eta$ & $0.05$ & $0.025$ & $0.01$ & $0.005$ & $0.002$ & $0.001$ \\
\hline
Test accuracy & $68.83\%$ & $98.81\%$ & $99.24\%$ & $99.24\%$ & $99.24\%$ & $99.25\%$ 
\end{tabular}
\end{center}
\end{table}

\subsection{Detailed Investigation on the {\sf HIGGS} Data}
\label{subsec:compare-sgd}
We compare AUC values obtained by our {\sl Newton} and {\sl SG} implementations with those reported in \cite{PB14a} on {\sf HIGGS}. In our method, the sampling rate for calculating the subsampled Gauss-Newton matrix is set to be 1\%. Following the setting in Section \ref{subsec:multiclass}, we consider two initializations (dense and sparse). Then for each type of initialization, both {\sl SG} and {\sl Newton} start with the same initial weights and biases. 
Note that our SG results are different from those in \cite{PB14a} because we use different activation functions and initial values for weights and biases.%
\footnote{Their initialization setting is the same as our dense initialization, but the values used by them are not available.} %
Because of resource constraints, we did not conduct a validation procedure to select SG's initial learning rate. Instead, we used the learning rate $0.05$ by following \cite{PB14a}. The results are shown in Table \ref{table:higgs-result} and we can see that the Newton method often gives the best AUC values.


\begin{table}[t]
\caption{A comparison between the AUC obtained by {\sl SG} and that by the distributed {\sl Newton} on the {\sf HIGGS} data set. We list the results in \cite{PB14a} as a reference, where ``NA'' means that the result is not reported. See explanation in Section \ref{subsec:compare-sgd} about the different results between our SG and Baldi et al.'s.}
\begin{center}
\resizebox{\textwidth}{!}{
\begin{tabular}{cccccccc}


\multirow{2}{50pt}{Network} & \multirow{2}{40pt}{Split} & \multicolumn{2}{c}{Dense Initialization} && \multicolumn{2}{c}{Sparse Initialization} & \multirow{2}{*}{\cite{PB14a}} \\[0pt]

\cline{3-4} \cline{6-7}

& & {\sl Newton} & {\sl SG} && {\sl Newton}& {\sl SG} & \\


\hline                                                                                           
$28$-$300$-$1$ & $2$-$2$-$1$                          & $0.843$ & $0.469$          && $0.843$     & $0.684$ & $0.816$\\ 
\hline                                                                                           
$28$-$600$-$1$ & $2$-$3$-$1$                          & $0.849$ & $0.501$          && $0.849$     & $0.759$ & NA\\
\hline                                                                                           
$28$-$1000$-$1$ & $2$-$4$-$1$                         & $0.851$ & $0.500$       && $0.853$     & $0.734$ & $0.841$\\ 
\hline                                                                                           
$28$-$2000$-$1$ & $2$-$8$-$1$                         & $0.853$ & $0.500$      && $0.855$ & $0.504$ & $0.842$\\
\hline                                                                                           
$28$-$300$-$300$-$1$ & $2$-$2$-$1$-$1$                & $0.851$ & $0.530$          && $0.860$     & $0.825$ & NA\\ 
\hline                                                                                           
$28$-$300$-$300$-$300$-$1$ & $2$-$2$-$2$-$1$-$1$      & $0.867$ & $0.482$          && $0.879$ & $0.849$ & $0.850$\\
\hline
$28$-$300$-$300$-$300$-$300$-$1$ & $2$-$2$-$2$-$2$-$1$-$1$      & $0.867$ & $0.504$          && $0.875$ & $0.848$ & $0.872$\\
\hline

\end{tabular}
}
\end{center}
\label{table:higgs-result}
\end{table}

\par
\begin{center}
\begin{figure}[t]
        \begin{subfigure}{\textwidth}
		\begin{tabular}{c c}
			\begin{subfigure}{0.5\textwidth}
			\includegraphics[scale=0.5]{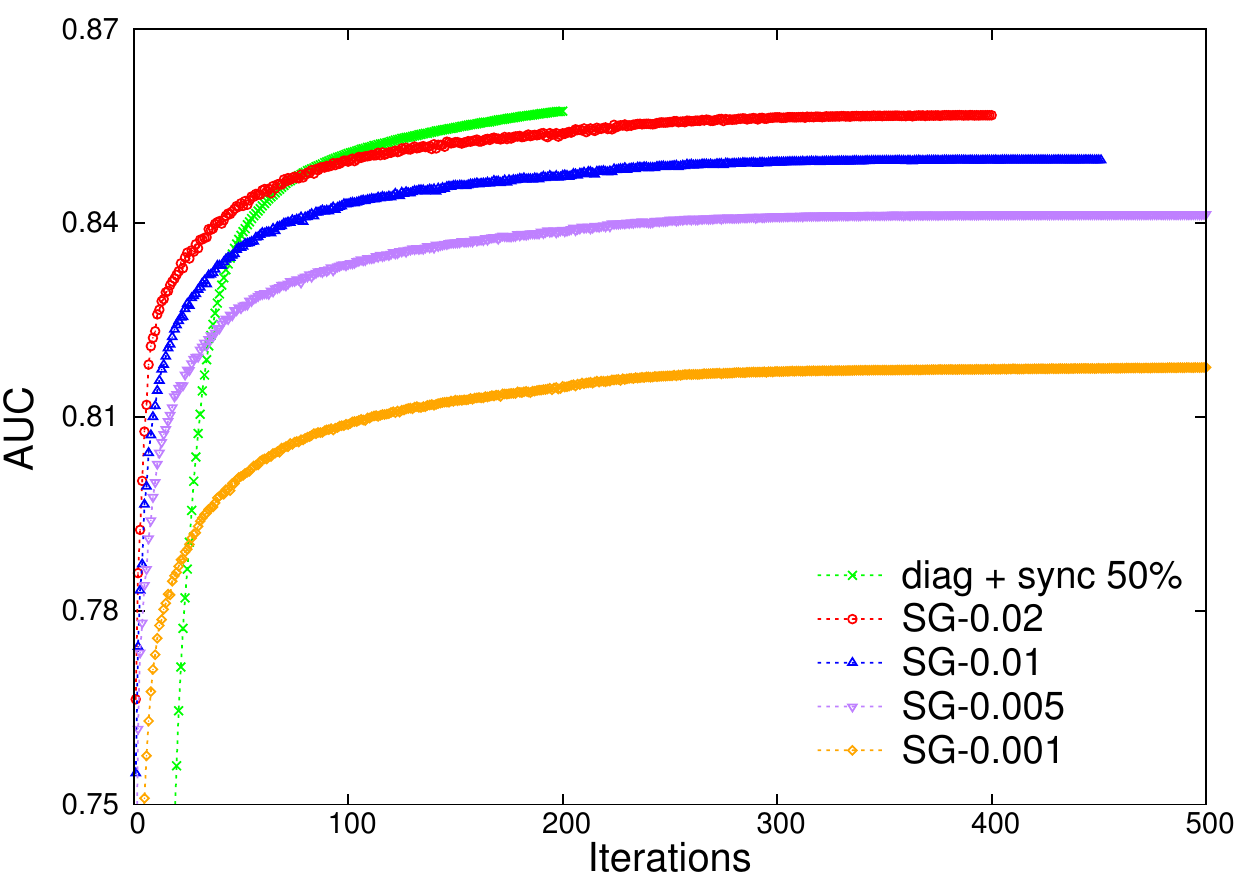}
			\end{subfigure}
			\begin{subfigure}{0.5\textwidth}
			\includegraphics[scale=0.5]{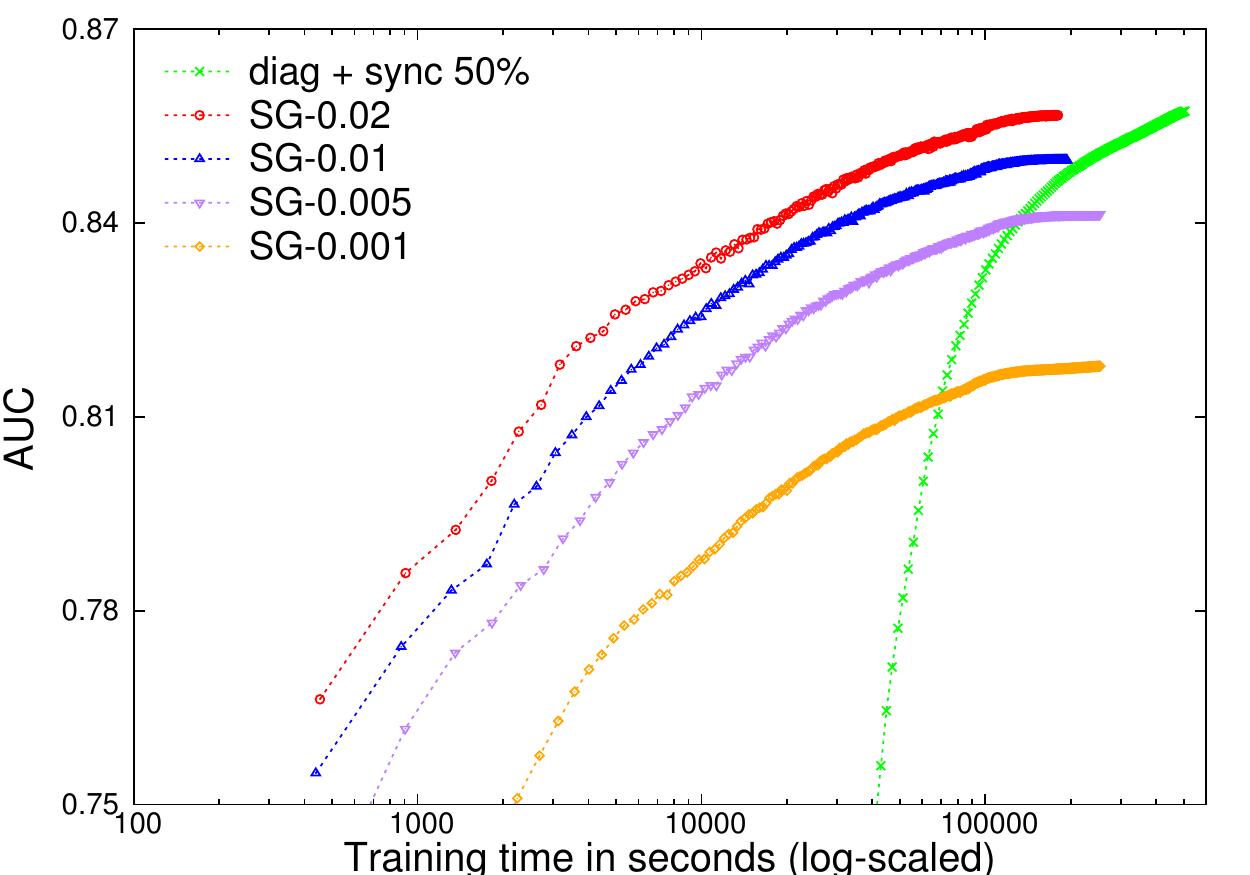}
			\end{subfigure}
        \end{tabular}
		\vspace*{-0.5cm}
        \caption{Dense initialization.}
        \end{subfigure}
		\begin{subfigure}{\textwidth}
		\begin{tabular}{c c}
			\begin{subfigure}{0.5\textwidth}
			\includegraphics[scale=0.5]{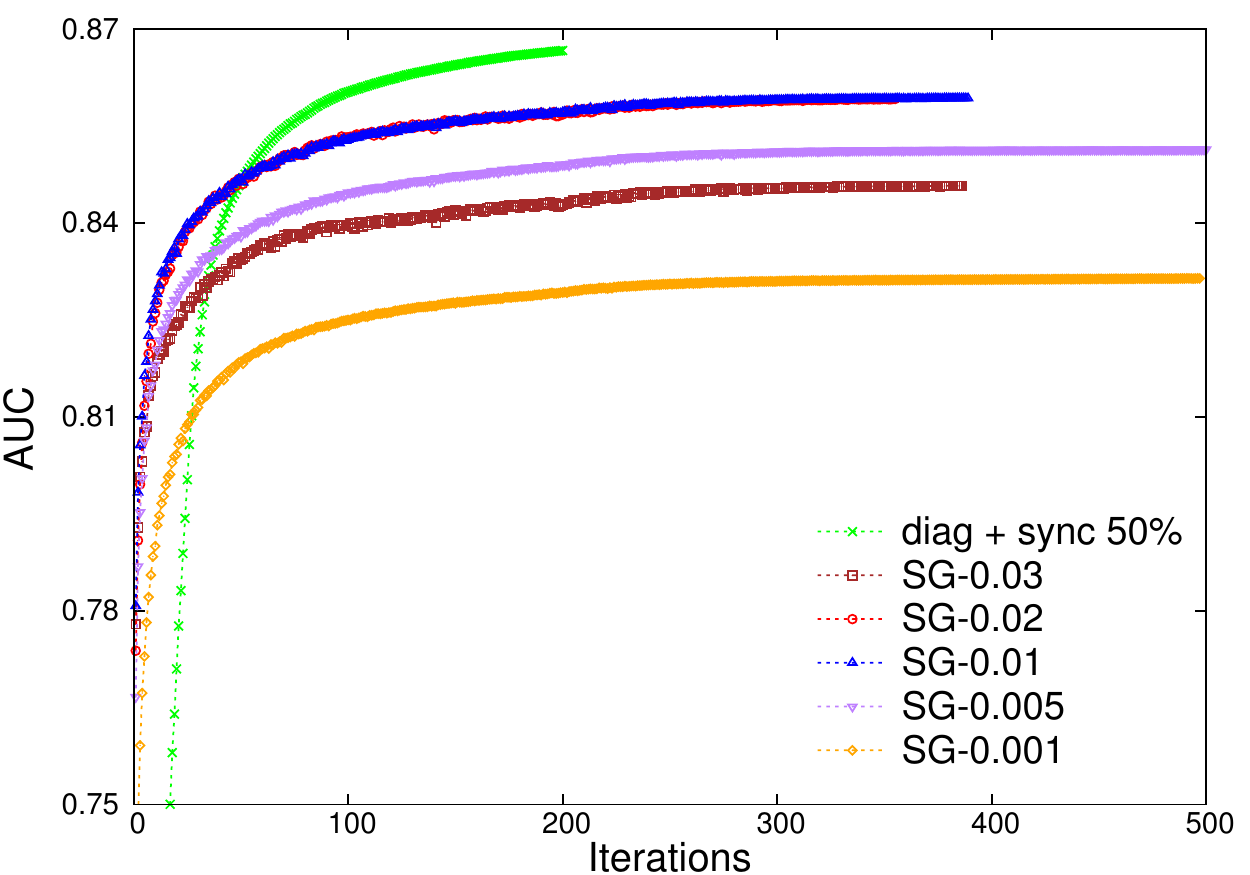}
			\end{subfigure}
			\begin{subfigure}{0.5\textwidth}
			\includegraphics[scale=0.5]{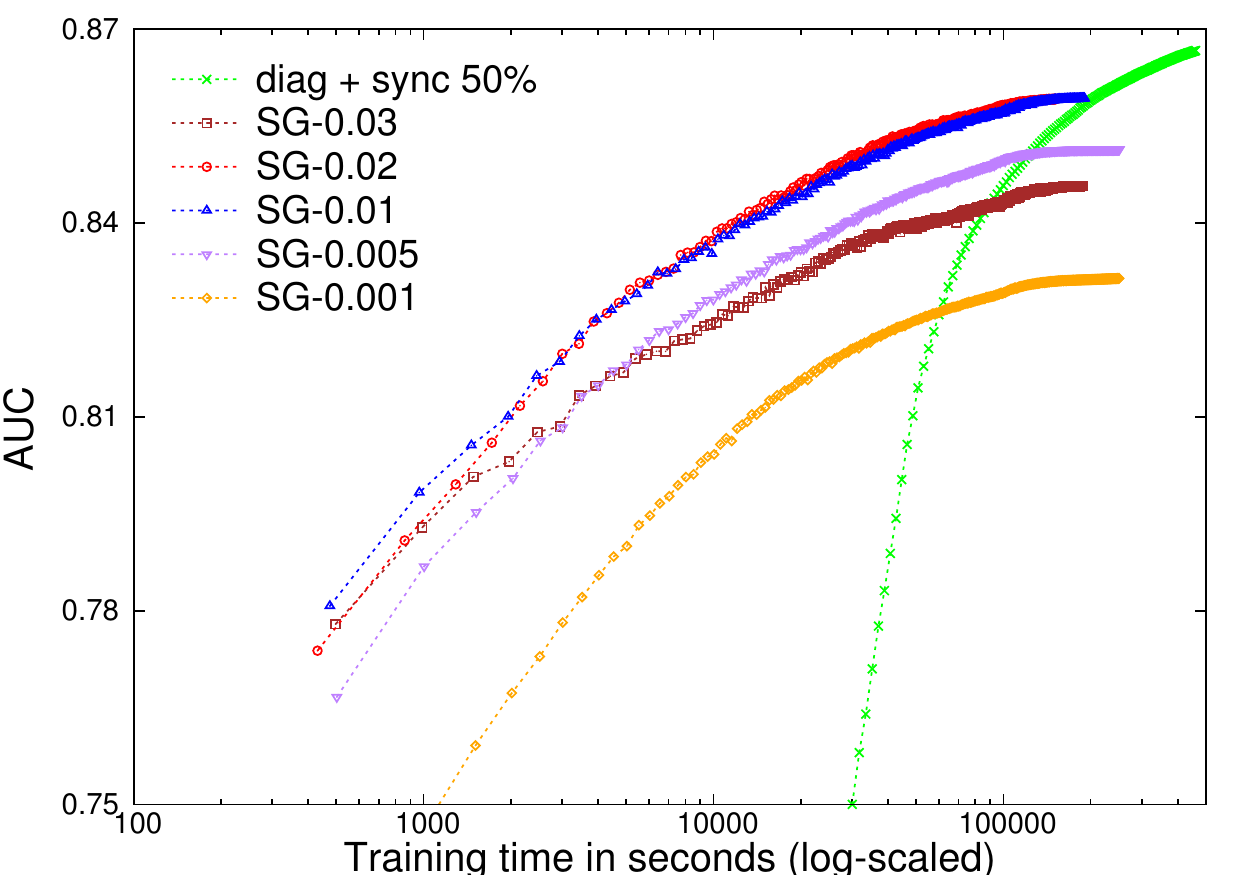}
			\end{subfigure}
        	\end{tabular}
		\vspace*{-0.5cm}
        \caption{Sparse initialization.}
        \end{subfigure}
\vspace*{-0.5cm}
\caption{A comparison between {\sl SG} and {\sl Newton}. A 28-300-300-1 network is applied to train HIGGS. {\sl SG}-$x$ means that the initial learning rate $x$ is used. For {\sl Newton},
each iteration means that we go through line $5$ to line $24$ in Algorithm \ref{alg:distri}, while for {\sl SG}, each iteration means that we go through the whole training data once.
The curve of {\sl SG-0.03} in the dense initialization is not presented because the AUC value never exceeds 0.5. Left: AUC versus number of iterations. Right: AUC versus training time in seconds (log-scaled).}
\label{fig:sgdslow}
\end{figure}
\end{center}

 In Section \ref{subsec:multiclass} we have mentioned that {\sl SG}'s performance may be sensitive to the initial learning rate. The poor results of {\sl SG} in Table \ref{table:higgs-result} might be because we did not conduct a selection procedure. Thus we decide to investigate the effect of the initial learning rate on the AUC value with the network structure 28-300-300-1 used in the earlier experiment in Table \ref{table:higgs-result}. To compare the running time, both {\sl SG} and {\sl Newton} run on the same G3 type machine with $8$ cores in Microsoft Azure. The results of the AUC values versus the number of iterations and the training time are shown in Figure \ref{fig:sgdslow}. We clearly see again that the performance of {\sl SG} depends significantly on the initial learning rate. Our experiments indicate that while {\sl SG} can yield good performances under suitable parameters, the parameter selection procedure is essential. In contrast, Newton methods are more robust because we do not need to fine tune their parameters.

\section{Discussion and Conclusions}
\label{sec:conclu}
 For the future works, we list the following directions.
\begin{enumerate}[1.]
\item
It is important to extend the proposed method for other types of neural networks. For example, convolutional neural networks (CNNs) are 
popular for computer vision applications \citep[e.g., ][]{AK12b, KS14a}.
Because CNNs generally have fewer weights per layer, our method has the potential to train deep networks for large-scale image classification.
\item
Instead of the Gauss-Newton matrix,
we may consider other ways to use or approximate the Hessian such as the recent works by \cite{XH16a}.
\item
For results in Tables \ref{table:results-poker-vehicle} and \ref{table:higgs-result}, we consider the model after running $100$ Newton iterations. An advantage of Newton over stochastic gradient
is that we can apply a gradient-based stopping condition. We plan to investigate its practical use.
\item
It is known that using suitable preconditioners can effectively reduce the number of CG steps in solving a linear system.
Studies of applying preconditioned CG methods in training neural networks include, for example, \cite{OC11b}. We plan to investigate how to apply
preconditioning in our distributed framework.
\end{enumerate}
\par In summary, in this paper we proposed novel techniques to implement distributed Newton methods for training large-scale neural networks, and achieved
both data and model parallelisms.
\section*{Acknowledgements}
This work was supported in part by MOST of Taiwan via the grant 105-2218-E-002-033
and Microsoft via Azure for Research programs.
\bibliographystyle{apalike}
\bibliography{sdp}

\end{document}